\documentclass{ieeetj}
\usepackage{cite}
\usepackage{amsmath,amssymb,amsfonts}
\usepackage{algorithmic}
\usepackage{graphicx,color}
\usepackage{textcomp}
\usepackage{xcolor}
\usepackage{hyperref}
\hypersetup{hidelinks=true}
\usepackage{algorithm,algorithmic}
\usepackage{caption}
\usepackage{setspace}
\usepackage{tabularx}

\usepackage{comment}
\usepackage{siunitx}
\usepackage[acronyms, shortcuts]{glossaries}
\usepackage{subfig}
\let\labelindent\relax
\usepackage{enumitem}
\usepackage[numbers]{natbib}
\usepackage{adjustbox}
\usepackage{balance}
\let\DeclareUSUnit\DeclareSIUnit

\DeclareUSUnit\feet{ft}

\newacronym[longplural={digital elevation models}]{dem}{DEM}{digital elevation models}
\newacronym[longplural={Permanently Shadowed Regions}]{psr}{PSR}{Permanently Shadowed Region}
\newacronym{lroc}{LROC}{Lunar Reconnaissance Orbiter Camera}
\newacronym{spa}{SPA}{South Pole-Aitken}
\newacronym{gitl}{GITL}{``ground-in-the-loop''}
\newacronym{conops}{ConOps}{concept-of-operations}

\newacronym{ins}{INS}{inertial navigation system}
\newacronym{ohv}{OHV}{off-highway vehicle}

\newacronym[longplural={convolutional neural networks}]{cnn}{CNN}{convolutional neural network}
\newacronym{sgbm}{SGBM}{Semi-Global Block Matching}
\newacronym{clahe}{CLAHE}{Contrast Limited Adaptive Histogram Equalization}
\newacronym{hed}{HED}{Holistically-Nested Edge Detection}
\newacronym{dsmn}{DSMNet}{Domain-Invariant Stereo Matching Network}

\newacronym{cboe}{CBOE}{coherent backscattering opposition effect}
\newacronym{shoe}{SHOE}{shadow hiding opposition effect}
\newacronym{osl}{OSL}{Open Shading Language}
\newacronym{brdf}{BRDF}{bidirectional reflectance distribution function}

\newacronym{rms}{RMS}{``root-mean-squared''}

\newcommand{\ignore}[1]{}  

\newcommand{\eg}{{e.g.}}

\newcommand\mydots{\makebox[1em][c]{.\hfil.\hfil.}}

\newcommand{\revised}[1]{{\leavevmode\color{black}#1}}
\def\BibTeX{{\rm B\kern-.05em{\sc i\kern-.025em b}\kern-.08em
    T\kern-.1667em\lower.7ex\hbox{E}\kern-.125emX}}
\AtBeginDocument{\definecolor{tmlcncolor}{cmyk}{0.93,0.59,0.15,0.02}\definecolor{NavyBlue}{RGB}{0,86,125}}

\def\OJlogo{\vspace{-4pt}}
\def\seclogo{\vspace{10pt}}

\def\authorrefmark#1{\ensuremath{^{\textbf{#1}}}}

\begin{document}
\receiveddate{XX Month, XXXX}
\reviseddate{XX Month, XXXX}
\accepteddate{XX Month, XXXX}
\publisheddate{XX Month, XXXX}
\currentdate{XX Month, XXXX}
\doiinfo{XXXX.2022.1234567}

\markboth{}{Atha {et al.}}

\title{{ShadowNav}: Autonomous Global Localization for Lunar Navigation in Darkness}

\author{Deegan Atha\authorrefmark{1}, R. Michael Swan\authorrefmark{1},
Abhishek Cauligi\authorrefmark{1}, Anne Bettens\authorrefmark{1,2},\\ Edwin Goh\authorrefmark{1},
Dima Kogan\authorrefmark{1}, Larry Matthies\authorrefmark{1}, Masahiro Ono\authorrefmark{1}}
\affil{Jet Propulsion Laboratory (JPL), California Institute of Technology (Caltech), Pasadena, CA, United States of America}
\affil{University of Sydney, Camperdown NSW 2050, Australia}
\corresp{Corresponding author: Deegan Atha (email: deegan.j.atha@jpl.nasa.gov).}
\authornote{The research was carried out at the Jet Propulsion Laboratory, California Institute of Technology, under a contract with the National Aeronautics and Space Administration (80NM0018D0004).}

\begin{abstract}
The ability to determine the \revised{position} of a rover in an inertial frame autonomously is a crucial capability necessary for the next generation of surface rover missions on other planetary bodies.
Currently, most on-going rover missions utilize ground-in-the-loop interventions to manually correct for drift in the \revised{position} estimate and this human supervision bottlenecks the distance over which rovers can operate autonomously and carry out scientific measurements.
In this paper, we present ShadowNav, an autonomous approach for global \revised{2D} localization on the Moon with an emphasis on driving in darkness and at nighttime.
Our approach uses the leading edge of Lunar craters as landmarks and a particle filtering approach to associate detected craters with known ones on an offboard map.
We discuss the key design decisions in developing the ShadowNav framework for use with a Lunar rover concept equipped with a stereo camera and an external illumination source.
Finally, we demonstrate the efficacy of our proposed approach in both a Lunar simulation environment and on data collected during a field test at Cinder Lakes, Arizona.
\end{abstract}

\begin{IEEEkeywords}
computer vision, extreme environments, GPS-denied operation, localization, perception
\end{IEEEkeywords}

\maketitle

\section{INTRODUCTION}
\IEEEPARstart{S}{pace} missions that require long-range autonomous driving on the surface of planetary bodies have gained significant interest lately especially for the Lunar surface.
For example, in the latest Decadal Survey \cite{NASEM2022}, the Endurance-A Lunar rover was recommended to be implemented as a strategic class mission as the highest priority of the Lunar Discovery and Exploration Program. 
This mission proposal calls for a \SI{2000}{\kilo\meter} traverse in the~\ac{spa} Basin to collect \SI{100}{\kilo\gram} of samples for delivery to Artemis astronauts.
This mission concept study~\cite{KeaneTikooEtAl2022} identifies several key capabilities required to complete this mission:
\begin{enumerate}
\item Endurance will need to drive 70\% of its total distance during the night to enable daytime hours dedicated to science and sampling.
\item The mission will require onboard autonomy for the majority of its operations, while the ground only handles contingencies. 
\item Global localization is necessary to maintain an error of $<$\SI{10}{\meter} relative to orbital maps. 
\end{enumerate}


\begin{figure}[t]
    \centering
    \subfloat[]{\includegraphics[width=0.48\columnwidth,trim={0cm 2cm 13cm 2cm},clip]{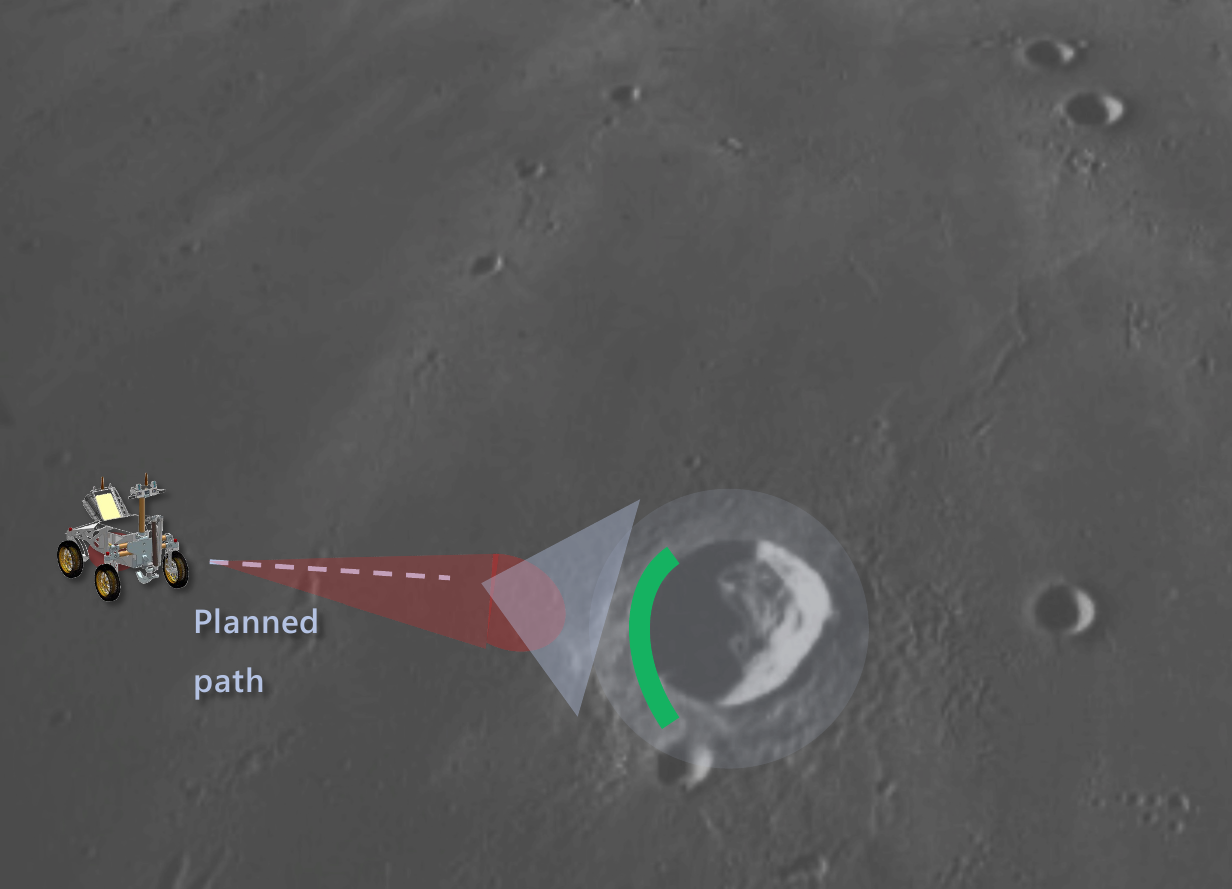}
    \label{fig:shadownav_concept_1}}
    \subfloat[]{\includegraphics[width=0.48\columnwidth,trim={0cm 2cm 13cm 2cm},clip]{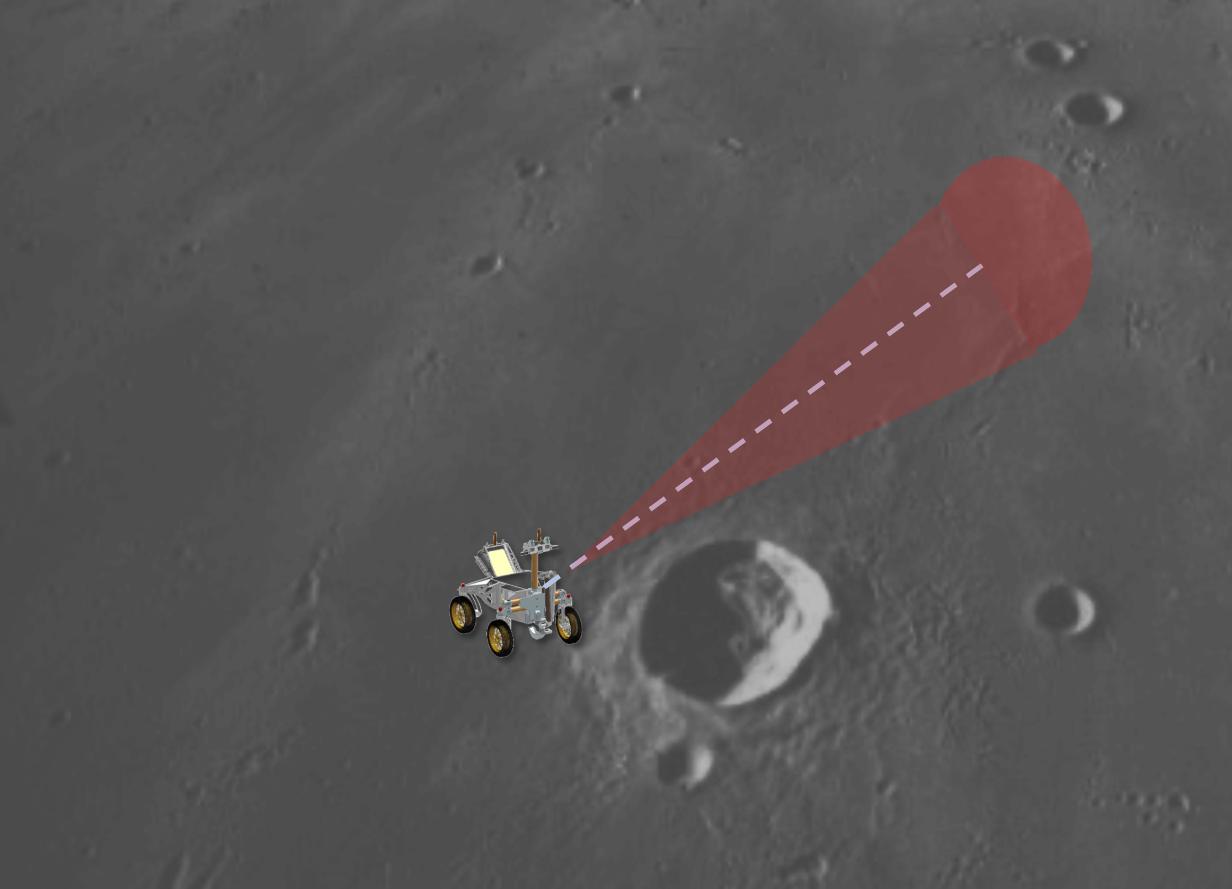}
    \label{fig:shadownav_concept_2}}
    \label{fig:shadownav_concept}
 \caption{The ShadowNav concept relies on using crater rims as landmarks. As the rover begins driving (a), its uncertainty in the global frame increases and must be corrected. By equipping the Lunar rover with a stereo camera and an external illumination source, the leading edges of detected craters (shown in green) are used to associate the crater with known ones from an offline map and decrease the pose uncertainty. As the rover continues to drive, (b) this process repeats intermittently to be able to continue mission operations.}
\end{figure}

This level of autonomy and drive distance would be an order of magnitude larger than any previous surface mission on a planetary body. 
For example, as the current state-of-the-art in beyond Earth surface autonomy, Mars 2020 has driven \SI{11.6}{\kilo\meter} autonomously over the course of a year and a half and the longest individual autonomous drive was approximately~\SI{700}{\meter}~\cite{VermaMaimoneEtAl2023}.
A key bottleneck that limits longer autonomous drives is the problem of absolute localization which is determining the position and orientation of the vehicle in an inertial frame.
Without a reliable means of accomplishing localization autonomously, the vehicle must stop and wait for ground-in-the-loop operators to manually perform absolute localization.
In the case of surface rovers used on planetary bodies, existing relatively localization methods experience approximately 2\% drift and this limits autonomous driving to approximately~\SI{500}{\meter} before a manual absolute localization is triggered once the error exceeds~\SI{10}{\meter}~~\cite{MaimoneChengEtAl2007,MaimonePatelEtAl2022}.

Currently however, the entirety of the Lunar surface does not have continuous communication with Earth.
Planning for a rover to wait until communications with the ground station is possible will severely slow down the traverse rate.
The lack of frequent absolute localization for the rover would lead to errors greater than the maximum \SI{10}{\meter} localization error of Endurance-A, which could present significant risks to the mission.
For proposed missions such as Endurance-A, the requirement to drive longer distances on the order of several kilometers a day necessitates the development of an autonomous and performant technique for absolute localization onboard the vehicle without ground-in-the-loop intervention.

In this work, we draw inspiration from recent techniques that have made use of craters as landmarks for performing absolute localization on the Moon during daytime driving~\cite{MatthiesDaftryEtAl2022,DaftryEtAl2023}.
For Lunar rover missions, craters have the potential to serve as landmarks as the average distance between craters of $\geq$\SI{10}{\meter} in diameter is~\SI{100}{\meter} on terrain with relatively fresh craters and~\SI{10}{\meter} on terrain with old craters~\cite{HiesingerVanDerBogertEtAl2012}.
Further, the~\ac{lroc} provides~\acp{dem} with a resolution between \SI{0.5}{\meter}-\SI{5}{\meter} per pixel \cite{RobinsonBrylowEtAl2010} and there are some~\acp{dem} within~\acp{psr}~\cite{CisnerosAwumahEtAl2017}.
However, unlike the daytime driving case, the lack of natural light available when driving within a~\ac{psr} or during the Lunar night limits what can be used as a landmark and the range at which the landmarks can be observed.

To address this challenge of degraded perception conditions, recent mission concepts have proposed equipping rover systems with a stereo camera and an illumination source for driving at night and \acp{psr}~\cite{KeaneTikooEtAl2022,RobinsonElliott2020, INSPIRE2022}.
In this work we propose an autonomous global \revised{2D} localization technique that uses a stereo camera with an external illumination source to detect crater rims in darkness for use in such mission architectures.
Global localization is accomplished by matching and scoring the detected crater rims against known craters from an orbital image using our novel \emph{Q-Score} metric.
A particle filter is used to estimate the absolute position amongst the uncertainty and nonlinearity of crater rim matches and corresponding Q-Scores.

{\bf Summary of Contributions: }
A preliminary version of our proposed approach was presented in~\cite{CauligiSwanEtAl2023}.
In this revised and extended version, we provide the following additional contributions:
\begin{enumerate}
\item Refinement of detection algorithm to utilize contour detection to remove false positives.
\item Use of intermittent localization wherein only specific landmark craters are used and other craters that may appear along a trajectory are ignored.
\item Expanded evaluation of perception within simulation including evaluating (a) impact of Gaussian noise and (b) hardware impacts such as camera height above ground.
\item Evaluation of optimal trajectories to view a landmark for absolution localization within a simulated Lunar environment.
\item Collection of data and evaluation of both perception and absolute localization performance on analogue Lunar data collected at Cinder Lakes Apollo Training Area over the course of three New Moon nights. A sample of this setup is in Figure~\ref{fig:cl-rig}. 
\end{enumerate}

{\em Organization: }
We begin in Section~\ref{sec:related_work} by reviewing prior work on global localization for surface rover missions and the underlying techniques necessary for detecting craters as landmarks.
In Section~\ref{sec:approach}, we present our proposed architecture on using stereo cameras to detect the leading edges of craters and the particle filter in which these craters are used as landmarks for global localization.
Section~\ref{sec:data_collection} presents the datasets used in our work, including (1) a photorealistic simulation environment used to generate synthetic stereo camera data and (2) nighttime datasets collected during field tests conducted in Cinder Lakes, Arizona.
Next, Section~\ref{sec:results} provides numerical results using these aforementioned datasets to validate the efficacy of our proposed approach.
Finally, we conclude in Section~\ref{sec:conclusion} and present directions of future work.

\section{RELATED WORK}\label{sec:related_work}

Existing Mars rover missions provide the benchmark for state-of-the-art in autonomous surface rover capabilities.
Visual odometry used on-board these Mars rovers for relative localization accrues approximately 2-3\% error~\cite{JohnsonGoldbergEtAl2008}.
\revised{Although the state-of-practice for Mars rover global localization entailed manual ground-in-the-loop interventions to correct this error, recent work has proposed an autonomous global localization framework for the Mars Perserverance rover~\cite{NashDwightEtAl2024,VermaNashEtAl2024}. However, this framework does not account for the statistical uncertainty associated with on-board camera measurements and is used only for daytime operations.}
For the proposed Endurance-A mission~\cite{KeaneTikooEtAl2022}, the mission~\ac{conops} requires that the rover maintain position knowledge in the inertial frame with less than~\SI{10}{\meter} error \revised{and be capable of performing during daytime and nightime driving cases.}
Under the assumption of~2\% relative localization error, the Endurance-A mission then expects to perform an absolute localization update approximately every~\SI{300}{\meter}.

In this work, we develop a framework that relies on craters as landmarks to determine \revised{absolute position.}
\revised{We assume that absolute orientation is obtained through the use of a star tracker, which is part of the Endurance sensor suite~\cite{KeaneTikooEtAl2022}, and prior work~\cite{EnrightBarfootEtAl2012} has demonstrated that heading can be determined within 0.02\textdegree \space accuracy using a star tracker.
However, a star tracker alone does not suffice for estimating absolute position as past work has reported up to~\SI{800}{\meter} of error~\cite{EnrightBarfootEtAl2012}.}
Prior work by~\cite{HiesingerVanDerBogertEtAl2012} has estimated the frequency of craters of size~\SI{10}{\meter} to be every~\SI{100}{\meter}, which allows for craters to be encountered frequently enough for a global localization update to be performed at the required intervals of~\SI{300}{\meter} for Endurance-A.
Moreover, we build upon prior work that has demonstrated the efficacy of using craters as landmarks for use in daytime Lunar driving~\cite{MatthiesDaftryEtAl2022}.
\revised{Other features for landmarks have been proposed for use on planetary surfaces.
These include skyline features such as mountains and the use of topographic maps~\cite{CozmanKrotkov1997,EbadiCobleEtAl2022,CozmanKrotkovEtAl2000} and matching surface maps created from lidar with orbital elevation maps ~\cite{CarleBarfoot2010}.
However, while skyline features are promising for use with daylight, detecting accurate skyline features in darkness is challenging.
Although matching surface maps to 3D orbital maps is a promising approach, prior works have relied on the use of lidar which is not used within this work.}

To detect craters as landmarks in this work, we utilize stereo cameras as the primary sensor.
Stereo cameras have been one of the primary perception technologies for rover navigation on planetary bodies, most notably the Mars rover missions, and are included as one of the primary perception sensors in proposed Lunar missions~\cite{KeaneTikooEtAl2022, RobinsonElliott2020}.
\revised{
Although lidar has been extensively studied and proposed for upcoming mission concepts~\cite{SunAbshireEtAl2013,LorenzTurtleEtAl2018}, this technology has not flown on a prior NASA mission and we limit our attention to the use of stereo cameras due to their flight heritage.}
In addition to classical methods for crater detection, recent approaches such as~\ac{sgbm}~\cite{Hirschmuller2007} and ~\acp{dsmn}~\cite{ZhangQiEtAl2020} have demonstrated promise in using deep learning for crater detection using stereo.

Crater detection is a crucial aspect of Lunar navigation and for using craters as landmarks for accurate absolute localization.
Various methodologies have been explored to achieve accurate crater detection in challenging Lunar environments.
Studies such as Lunarnet~\cite{LiounisSwensonEtAl2019} have used~\acp{cnn} to detect craters from orbital images captured by instruments on-board,~\eg{}, the~\ac{lroc}\cite{RobinsonBrylowEtAl2010} spacecraft.
Lunarnet aimed to determine spacecraft pose by analyzing crater patterns, although challenges such as false positives and sensitivity to sunlight were encountered. 
Other algorithmic approaches have also been investigated.
In a comparison of crater-detection algorithms,~\cite{WoickeMorenoEtAl2018} evaluated six different methods based on criteria such as detection rates, false detection, accuracy, robustness, and run-time.
These methods include edge detection, identifying bright or shaded portions of a crater, template matching, and supervised learning-based techniques.
Despite their merits, these algorithms often struggle with complex scenes, overlapping craters, and lighting conditions.
Additionally, works such as DeepMoon~\cite{SilburtAliDibEtAl2019} employ elevation maps generated from radar or lidar data for crater detection.
Template matching, as seen in PyCDA~\cite{Klear2018}, offers an open-source solution for automated crater detection using~\acp{cnn} and per-pixel likelihood estimation.
While these methods have contributed to advancements in crater detection, challenges remain in extending capabilities to nighttime or shadowed operations and improving accuracy in diverse Lunar terrains.

There have been several proposed methods to perform an absolute localization update.
In~\cite{HwangboDiEtAl2009}, the authors propose a localization procedure that matches an observed rover image with an orbital map.
However, this results in a deterministic estimate of the robot belief since it neglects the rover motion model.
Prior works have also proposed the use of Lunar satellites to provide inertial position updates for Lunar rovers~\cite{BhamidipatiMinaEtAl2023,CortinovisMinaEtAl2024,AudetMelmanEtAl2024}, but current orbiters do not provide this capability and a crater-based localization approach would still fill in temporal gaps between such orbiter updates.
\cite{WuPotterEtAl2019} presents a purely data driven model using a \ac{cnn} trained on synthetic data to match the rover observations with orbital imagery.
Similar to our approach, \cite{FranchiNtagiou2022} presents a particle filtering technique which uses a Siamese neural network to compare rover monocular camera imagery with orbital imagery to assign each particle a likelihood weight.
The authors in~\cite{DaftryEtAl2023} propose a similar approach for Lunar absolute localization known as LunarNav.
However, LunarNav focuses on the daytime localization problem and therefore considers different methods of crater matching that rely on greater knowledge of the surface geometry than available in the nighttime case.

\begin{figure*}[t]
    \centering
    \includegraphics[width=1.0\textwidth,trim={0cm 8cm 0cm 7cm},clip]{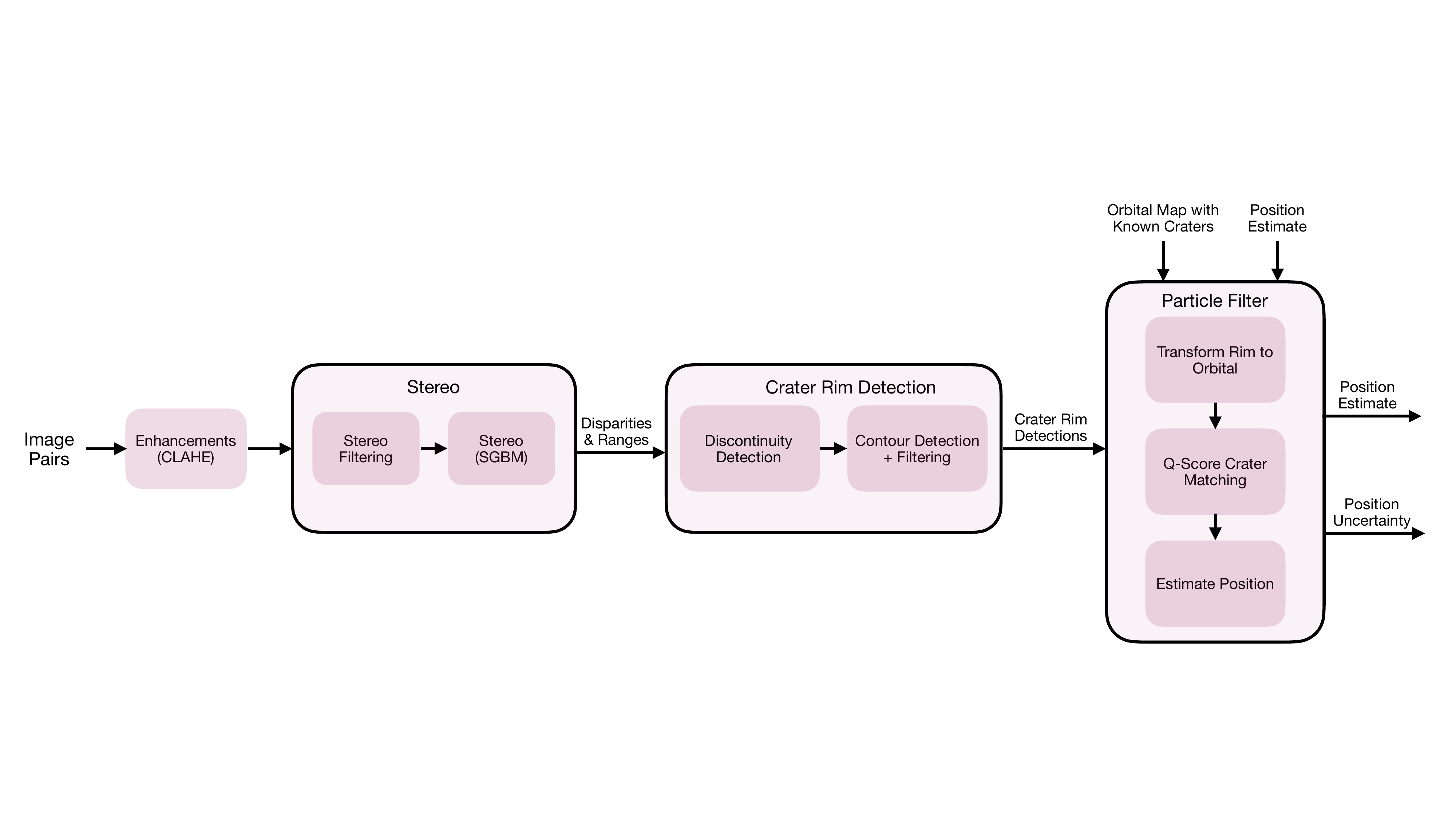}
    \caption{The four key components of the ShadowNav algorithmic pipeline are illustrated here. The system entails enhancing the image and then performing stereo and then a crater rim detection. This crater rim detection is then used as an input sample to a particle filter to perform absolute localization.
    }
    \label{fig:system}
\end{figure*}

A crucial component of localization in a Lunar environment is the problem of crater matching.
For instance, LunarNav~\cite{DaftryEtAl2023} introduces a novel approach that leverages crater matching for absolute localization for daytime Lunar driving using particle filtering and parametric matching using an on-board lidar.
Notably, the authors in~\cite{SilburtAliDibEtAl2019} and~\cite{SilvestriniPiccininEtAl2022} use neural networks to match craters, however these require a well labelled library to be known in advance for template matching.

\section{APPROACH}~\label{sec:approach}
In this section, we present our technical approach for performing absolute localization on the Lunar surface with the use of onboard illumination sources.
As shown in Figure~\ref{fig:system}, the four key components of the ShadowNav global localization pipeline include: (1) image enhancement, (2) stereo, (3) crater rim detection and refinement, and (4) the use of crater matching within a particle filter.

\subsection{Assumptions}
\revised{The methodology proposed in this work contains a few important assumptions primarily based on the Endurance concept \cite{KeaneTikooEtAl2022} and some consideration for other mission concepts for with driving in Lunar darkness~\cite{INSPIRE2022,RobinsonElliott2020}. 
As the Endurance concept baselines the use of a star tracker and past work~\cite{EnrightBarfootEtAl2012} has shown that star trackers provide a highly accurate estimate of orientation for planetary surface rovers, we focus our approach on 2D position localization within an orbital map.
We also assume that stereo cameras are the primary perception sensor and camera heights are assumed to fall in the range of proposed mast heights among the mission concepts \SI{1.5}-\SI{2.5}{\meter}.
As each mission concept focused on the nighttime driving case has called for, we baseline the use of an external illumination source provided by assuming the use of a headlight placed near the stereo cameras.
Lighting power is not specifically detailed, but the Endurance concept proposes the use of a ``high-intensity LED''.
Therefore, a high-intensity LED was used for experimental data collection.
The stereo camera specifications are based upon the NAVCAM cameras from the Perseverance rover~\cite{MakiGruelEtAl2020}.
Lidar could potentially replace or augment the stereo cameras, but is not the focus of this work as lidar is not a part of the mission concepts for which we are focused on and further lacks flight heritage.
With regards to available computational capabilities, we baseline the use of a Snapdragon processor due to its inclusion on missions such as the Ingenuity helicopter~\cite{BalaramCanhamEtAl2018} and the upcoming CADRE Lunar rover mission~\cite{DeLaCroixRossiEtAl2024}.
}

\subsection{Image Enhancement}
The first step in our system is to utilize~\ac{clahe}~\cite{YadavMaheshwariEtAl2014} to enhance the raw images.
In this work, we consider a single light source with consistent illumination provided across the field of light.
Therefore, there is a significant decrease in pixel intensity values captured by a camera as the range of the physical region being imaged increases.
Since~\ac{clahe} is adaptive, it equalizes the image within sub-regions, producing a more consistent image brightness to use as input to stereo algorithms.
A potential downside of~\ac{clahe} is that it can magnify noise within the image which is explored within this work. 

\subsection{Stereo}
The second step within our system is to compute stereo.
This is accomplished using~\texttt{mrcal}~\cite{mrcal} and~\ac{sgbm}~\cite{Hirschmuller2007}.
The disparity and range maps produced provide the input for crater detection algorithms.
An additional cleanup step is used after stereo is computed to handle noisy disparity measurements in the far range.

{\em Stereo Refinement: }
It is assumed that the as the range of terrain increases, the corresponding pixels will be closer to the top of an image.
Therefore, to filter the far range noisy stereo measurements, the average range for all image rows are first computed.
Next, these average row values are searched starting at ranges beyond a parameter (\SI{20}{\meter} in this work).
During this search, if the range decreases for a parameterized number of rows consistently, then all computed stereo values after this peak are removed.

\subsection{Crater Leading Edge Detection}
After stereo is computed, this output is used to perform crater leading edge detection in two steps.
The first is to detect discontinuities within the disparity and range images and the second is to connect these discontinuity locations into a single rim via a contour detection. 

{\em Discontinuity Detection: }
To perform the discontinuity detection, vertical columns within the disparity image are searched from bottom to top with holes removed from the search. 
Along the search if both a disparity discontinuity and range discontinuity are larger than set thresholds, then the pixel at the start of the discontinuity is marked within a mask.

{\em Contour Detection and Refinement: }
Once the discontinuity detection is complete, the detected mask image is dilated and then eroded to connect potentially noisy detections.
Next, a contour detection algorithm is run over the mask to connect marked pixels.
These contours are then filtered by both number of pixels and estimated length.
The length of the contour is estimated by taking the average range value of all pixels in a contour and computing the width via Equation~\eqref{eq:contour-width}:
\begin{equation}
\label{eq:contour-width}
W_{m} = \frac{W_{px} * R_{m}}{l_\textrm{focal}},
\end{equation}
where $W_m$ is the width of the crater rim in meters, $R_m$ the average range to each pixel in meters, and $l_\textrm{focal}$ if the focal length of the camera.

\subsection{Absolute Localization via Particle Filter}
\label{sec:particle-filter}

\begin{algorithm}[t]
\caption{Q-Score Computation}
\label{alg:Qscore}
\algsetup{indent=0.45em} 
\begin{algorithmic}[1]
{\small
    \REQUIRE Belief $b_i^t$, set of crater observations $\{ z_{0,\textrm{rover}}^t, \mydots{}, z_{m,\textrm{rover}}^t \}$, set of ground truth craters $\{ c_{0,\textrm{world}}^t, \mydots{}, c_{\ell,\textrm{world}}^t \}$, positive value $\varepsilon$~\label{line:Qscore_input}
    \STATE $\mathcal{Q}_\textrm{inc} \gets \varepsilon$~\label{line:Qinc_init}
    \FOR {$i=1,\ldots,m$}
        \STATE $z_{0,\textrm{world}}^t \gets \textrm{rover\_to\_world}(z_{0,\textrm{rover}}^t)$~\label{line:tr_meas}
        \STATE $d_\textrm{cr} \gets \min \| c_{j,\textrm{world}}  - z_{0,\textrm{world}}^t\|$~\label{line:cr_association}
        \STATE ${\mathcal{Q}}_\textrm{inc} \gets {\mathcal{Q}}_\textrm{inc} +  d_\textrm{cr} $~\label{line:Qinc_add}
    \ENDFOR
    \STATE $Q_\textrm{score}\gets \min \Big{(} 1, (\frac{1}{m}{\mathcal{Q}}_\textrm{inc})^{-1} \Big{)}$~\label{line:Qscore_comp}
    \RETURN $Q_\textrm{score}$
}
\end{algorithmic}
\end{algorithm}

\begin{algorithm}[h]
\caption{ShadowNav Particle Filtering Algorithm}
\label{alg:ShadowNavPF}
\algsetup{indent=0.45em} 
\begin{algorithmic}[1]
{\small
    \REQUIRE Initial belief distribution $(\mu_0, \Sigma_0)$, number of particles $N_s$, number of effective particles threshold $N_\textrm{eff,thresh}$~\label{line:pf_inputs}
    \STATE $\{ b_1^0, \mydots{}, b_{N_s}^0 \} \gets \textrm{sample\_beliefs}(\mu_0, \Sigma_0)$~\label{line:sample_init}
    \STATE $\{ w_1^0, \mydots{}, w_{N_s}^0 \} \gets \{ 1, \mydots{}, 1 \}$~\label{line:weight_init}
    \STATE $t \gets 1$
    \WHILE{ $\textrm{particle filter running}$}
        \STATE $\{ z_0^t, \mydots{}, z_m^t \} \gets \textrm{get\_observations()}$~\label{line:measurements}
        \STATE $\{ q_1^t, \mydots{}, q_{N_s}^t \} \gets \{ 0, \mydots{}, 0 \}$~\label{line:q_score_init}
        \FOR { $i = 1, \mydots{}, N_s$ }
            \STATE $b_i^{t} \gets \textrm{propagate\_sample}(b_i^{t-1})$~\label{line:belief_update}
            \STATE $q_i^{t} \gets \log\textrm{Q\_score}(b_i^{t}, \{ z_0^t, \mydots{}, z_m^t \})$~\label{line:q_score_comp}
        \ENDFOR
        \STATE $q_\textrm{min}^t \gets \min ( q_1^t, \mydots{}, q_{N_s}^t )$~\label{line:q_score_min}
        \FOR { $i = 1, \mydots{}, N_s$ }
            \STATE $w_i^{t} \gets w_i^{t-1} + q_i^t - q_\textrm{min}^t$~\label{line:weight_update}
        \ENDFOR
        \STATE $N_\textrm{eff} \gets \textrm{compute\_N}_\textrm{eff}( w_1^t, \mydots{}, w_{N_s}^t )$~\label{line:compute_Neff}
        \IF {$N_\textrm{eff} \leq N_\textrm{eff,thresh}$}
            \STATE $\{ b_1^t, \mydots{}, b_{N_s}^t \} \gets \textrm{resample\_beliefs}(\{ b_i^t \}_{i=1}^{N_s}, \{ w_i^t \}_{i=1}^{N_s} )$~\label{line:resample}
            \STATE $\{ w_1^t, \mydots{}, w_{N_s}^t \} \gets \{ 1, \mydots{}, 1 \}$~\label{line:weights_reassign}
        \ENDIF
        \STATE $t \gets t+1$
    \ENDWHILE
}
\end{algorithmic}
\end{algorithm}

\begin{algorithm}[t]
\caption{Systematic Resampling}
\label{alg:systematic_resampling}
\algsetup{indent=0.45em} 
\begin{algorithmic}[1]
{\small
    \REQUIRE Particles $\{ b_1^t, \mydots{}, b_{N_s}^t \}$ and associated weights $\{ w_1^t, \mydots{}, w_{N_s}^t \}$
    \STATE $n^t = \log \Big{(} \sum_{i=1}^{N_s} \exp(b_i^t) \Big{)}$~\label{line:weight_normalization_val}
    \STATE $\{ \tilde{w}_0^t, \mydots{}, \tilde{w}_{N_s}^t \} \gets \{0, \mydots{}, 0 \}$~\label{line:normalized_weights_init}
    \FOR {$i = 1, \mydots{}, N_s$}
        \STATE $\tilde{w}_i^t \gets \exp(w_i^t - n^t)$~\label{line:normalize_weights}
    \ENDFOR
    \STATE $\{ q_0, \mydots{}, q_{N_s} \} \gets \textrm{cum\_sum}(\{ \tilde{w}_0^t, \mydots{}, \tilde{w}_{N_s}^t \})$~\label{line:weights_cumsum}
    \STATE $n \gets 0$
    \STATE $m \gets 0$
    \STATE $u_0 \sim {\mathbb{U}} ( 0, \frac{1}{N_s})$~\label{line:sample_uniform_val}
    \WHILE {$n \leq N_s$}
        \STATE $u = u_0 + \frac{n}{N_s}$
        \WHILE {$q_m \leq u$}
            \STATE $m \gets m+1$
        \ENDWHILE
        \STATE $n \gets n+1$
        \STATE $b_n^t \gets b_m^t$
    \ENDWHILE
    \RETURN $\{ b_0^t, \mydots{}, b_{N_s}^t \}$
}
\end{algorithmic}
\end{algorithm}

Here, we provide an overview of the proposed ShadowNav particle filtering approach used to handle measurement uncertainties stemming from stereo and disparity computation errors.
First, we provide details on \revised{the motion model used for our system and the use orbital data as landmarks.}
Next, we describe our Q-Score metric used in the belief update step.
Finally, we present our approach for a filtering system using a particle filter.

\revised{
\subsubsection{Motion Model}~\label{label:rel-loc}
We estimate the belief $b$ associated with the 2D position of the rover,
\begin{equation}
    b = (x, y),
\end{equation}
where $x$ and $y$ are the two-dimensional position coordinates of the rover.
To simulate the estimated position of a Lunar rover, we assume a simplified motion model,
\begin{equation}
    b_{k+1} = \begin{pmatrix}
        x_k + s_{k+1,x} + w_{k,x} \\
        y_k + s_{k+1,y} + w_{k,y} \\
    \end{pmatrix},
    \label{eq:motion_model}
\end{equation}
where $s_{k+1}$ is the translation step computed from relative localization and $w_k$ is an additive noise term.
In this work we simulate the translation step estimate using
\begin{equation}
    s_{k+1} = \begin{pmatrix}
        s'_{k+1,x} + r_{k,x} \\
        s'_{k+1,y} + r_{k,x} \\
    \end{pmatrix},
\end{equation}
where $s'_{k+1}$ is the ground truth translation step and $r_k$ is a bias term.
The $r_k$ term is used with $s'_{k+1}$ to produce an estimate with error to simulate relative localization drift.
$r_k$ is a random vector sampled from a Gaussian distribution based on an error rate $p$ of $s_{k+1}$.
The mean of this distribution is,
\begin{equation}
    \mu_k = \begin{pmatrix}
        d_x p|s_{k+1}|\\
        d_y p|s_{k+1}|\\
    \end{pmatrix},
\end{equation}
where $\hat{d} = (d_x, d_y)$ is a unit direction vector that is randomly sampled from a uniform distribution of all possible directions. 
The covariance $\Sigma_k$ of the distribution is
\begin{equation}
    \Sigma_k = \textrm{diag}\begin{pmatrix}
        |\mu_{k,x}| \\
        |\mu_{k,y}| \\
    \end{pmatrix}.
\end{equation}%
The $\hat{d}$ unit direction vector is selected at the start of a replay of a trajectory.
Therefore, over the course of many Monte Carlo runs, the bias will be added in many different directions to test against drift in all directions.
The $w_k$ noise term models position uncertainty and should be based on the actual uncertainty of a relative localization approach.
In this work, $w_k$ is a randomly sampled vector about a Gaussian distribution with zero mean and a standard deviation based $p$.
}

\subsubsection{Orbital Crater Ground Truth}
In this work, we consider only specific ground truth craters which require a planned trajectory to drive towards.
To this end, only the specific landmark craters to be used for global localization are labeled. 
When the estimated particle filter position was not within \SI{20}{\meter} of the ground truth rim, the particle filter was not run and the position was propagated by a simulated noisy relative localization estimate update specified in Section \ref{label:rel-loc}.
Further, only the front half of a crater is used as ground truth. 
However, since the landmark crater can be approached from any angle, the entire crater rim is labeled and the front half of the rim is determined dynamically based on the heading of camera at a specific time.
Because using the entire crater rim for matching could lead to false positives, only the front edge of the crater is used as ground truth as our detection algorithm only attempts to detect the front half.

\subsubsection{Q-Score}\label{subsubsec:q_score}
In each iteration of the particle filter, we compute what is known as the {\em Q-Score}, the probabilistic weight that some position belief corresponds to the true rover position.
Algorithm~\ref{alg:Qscore} outlines the procedure for computing the Q-Score and the inputs necessary for the Q-Score computation are the belief $b_i^t$, a set of $m$ observed edges in rover frame, and a set of $\ell$ ground truth crater observations to associate these measurements with (Line~\ref{line:Qscore_input}).
The initial value $\mathcal{Q}_\textrm{inc}$ is set to some negligibly small, positive value $\varepsilon$ to later avoid divide-by-zero issues (Line~\ref{line:Qinc_init}) and is updated based on the distance between the observed edge and its associated ground truth observation (Line~\ref{line:cr_association}-\ref{line:Qinc_add}).
The Q-Score is computed as the reciprocal of $\mathcal{Q}_\textrm{inc}$ and a $\min$ operation is applied to normalize the value between 0 and 1 (Line~\ref{line:Qscore_comp}).
We note that the $\min$ operation is applied based on the orbital~\ac{dem} resolution in order to have the same Q-Score assigned for any observations and belief pairs that are less than or equal \SI{1}{\meter} away from ground truth.

\subsubsection{Filtering System}
\revised{
In this work, we use a particle filter as our localization back end to update the 2D position estimate.
The use of a particle filter was motivated by the need to be resilient to crater measurement errors.
\cite{FoxThrunEtAl2001} demonstrated that particle filters can be useful for a wide range of challenging localization scenarios and compared against a Kalman filter, can approximate a large range of probability distributions.
The ability for the localization to recover from poor solutions was a salient concern for the nighttime driving case for which the likelihood of \eg{} missed crater observations or incorrect associations between detected craters and craters in the landmark database is assumed to be higher. 
Moreover, the particle filter can accommodate the nonlinear motion model and process noise from~\eqref{eq:motion_model}.}
Algorithm~\ref{alg:ShadowNavPF} provides the full outline for the ShadowNav particle filtering algorithm.
The algorithm takes as input a Gaussian belief distribution $(\mu_0, \Sigma_0)$ assumed for the initial robot position, the number of particles $N_s$ to use in the particle filter, and a threshold for the effective number of beliefs $N_\textrm{eff,thresh}$ used to trigger resampling (Line~\ref{line:pf_inputs}).
The filter is initialized with $N_s$ particles drawn from the initial belief distribution and each assigned equal weight (Lines~\ref{line:sample_init}-\ref{line:weight_init}).
Given a new set of crater observations (Line~\ref{line:measurements}), a set of Q-Score measurements is initialized for computing for each individual particle (Line~\ref{line:q_score_init}).
After applying the motion model update to each particle (Line~\ref{line:belief_update}), the Q-Score for each updated particle is computed using the procedure from Alg.~\ref{alg:Qscore} by comparing against the current measurements (Line~\ref{line:q_score_comp}).
As is standard in many particle filtering implementations~\cite{ArulampalamMaskellEtAl2002}, we note that the particle weights are updated in the in $\log$-domain (Line~\ref{line:weight_update}) with a normalization step to ensure non-negative weights (Line~\ref{line:q_score_min}).
Next, the number of effective samples $N_\textrm{eff}$ at the current iteration is calculated (Line~\ref{line:compute_Neff}).
If $N_\textrm{eff}$ is below the threshold $N_\textrm{eff,thresh}$, then this is seen as an indication of particle filter ``degeneracy'', wherein the weights $\{ w_i^t \}$ collapse around a handful of particles.
In such a case, a new set of particles are resampled using the systematic resampling scheme (Line~\ref{line:resample}).

Algorithm~\ref{alg:systematic_resampling} provides an outline of the systematic resampling scheme used in this work.
Given a set of particles and their associated weights, the weights are first normalized to $(0, 1]$ from $\log$-domain (Lines~\ref{line:weight_normalization_val}-\ref{line:normalize_weights}) and the cumulative sum of these normalized weights $\tilde{w}_i^t$ computed (Line~\ref{line:weights_cumsum}).
The systematic resampling procedure then samples a random value $u_0$ from a uniform distribution inversely proportional to $N_s$ (Line~\ref{line:sample_uniform_val}) and this ensures that at least one particle is retained from each $\frac{1}{N_s}$ interval from the previous belief distribution.





\section{DATA COLLECTION}~\label{sec:data_collection}
In this section, we present details on the datasets generated to validate the efficacy of the proposed ShadowNav approach.
We first review the simulation environment used to generated photorealistic stereo data and then review the configuration used to collect nighttime field testing data from Cinder Lakes, Arizona.

\subsection{Simulation}\label{sec:sim-data}
We developed a simulation environment using the Blender software, which has become a popular software framework for generating photorealistic images in a Lunar environment\cite{CruesBielskiEtAl2023,CauligiSwanEtAl2023}.
Within this simulation, the Hapke lighting model \cite{Hapke2002,Hapke2012,SchmidtBourguignon2019} was implemented to approximate the Lunar surface reflectance.
This model will simulate the ``opposition effect'' which leads to a focused point of extreme saturation at a location within an image where the camera ray and light source are at zero phase angle.
We implemented the Hapke lighting model using ``old highland'' parameters of the Moon provided in~\cite{XuLiuEtAl2020}.
For our use case, the~\ac{cboe} was left out of our implementation and only the~\ac{shoe} was implemented as it dominates most or all lighting calculations.

To obtain a realistic 3D model of the surface geometry,~\acp{dem} produced from~\ac{lroc} are utilized.
~\ac{lroc} resolution is typically between \SI{2}{\meter}-\SI{5}{\meter} which allows for resolving craters of approximately \SI{10}{\meter}.
However, the~\ac{lroc} resolution is insufficient for generating smooth surface imagery within simulation.
Therefore, the~\acp{dem} from~\ac{lroc} are scaled down to be \SI{0.25}{\meter} resolution before being loaded into Blender.
Crater measurements for simulated data in future discussions are based on this scaled resolution.
This scaled~\ac{dem} is imported into Blender and a surface texture comprised of two scales of fractal Brownian motion is added.
This texture is a natural noise and simulates surface texture features for stereo algorithms to utilize.

\begin{figure}[t]
    \centering
    \includegraphics[width=1.0\columnwidth]{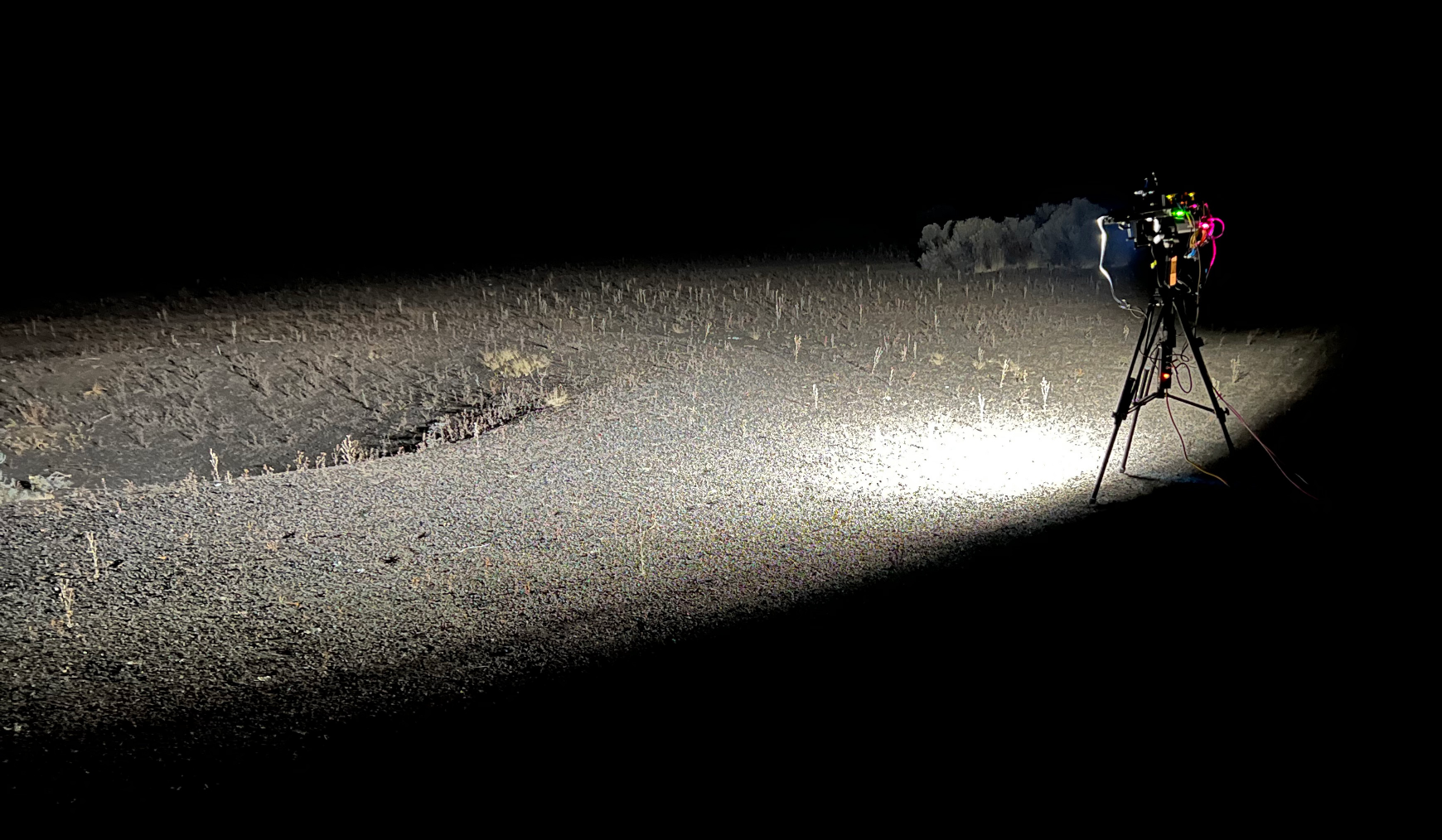}
    \caption{Image of the data collection rig capturing an image of a crater at Cinder Lakes Apollo Training Area.}
    \label{fig:cl-rig}
\end{figure}

\subsubsection{Simulated Dataset}
For simulation, we developed two datasets - one with the aim of testing the efficacy of our crater matching algorithm and a second dataset for testing the overall global localization pipeline.
The crater matching dataset was created by selecting four craters of diameters between \SI{5}{\meter} and \SI{20}{\meter}  from the orbital data.
The simulator was run to generate stereo pairs every \SI{1}{\meter} at four angles around the crater (0$^\circ$, 90$^\circ$, 180$^\circ$, and 270$^\circ$) and at ranges between \SI{5}{\meter}-\SI{25}{\meter}.
Additional datasets were created to test sensitivity to the parameters of the hardware configuration:
\begin{itemize}[noitemsep]
    \item Camera heights of \SI{1.5}{\meter}, \SI{2.5}{\meter}, and \SI{3}{\meter} off the ground
    \item Light offsets of \SI{0.2}{\meter} below the camera, inline with the camera and \SI{0.2}{\meter} above the camera
    \item Light power ranging from low, medium and high~\footnote{As Blender does not accurately model exposure times, the set of ``light power'' Blender parameters is used to change the brightness of rendered images.}
\end{itemize}

The second dataset generated was for localization across three different locations in the South Pole region of the Moon.
Additionally at each location, three different types of trajectories were generated:
\begin{itemize}[noitemsep]
    \item A straight line trajectory
    \item A half survey where 180 degrees of the ground truth crater is observed
    \item A full survey where 360 degrees of the ground truth crater is observed
\end{itemize}

\begin{figure}[t]
    \centering
    \includegraphics[width=1.0\columnwidth]{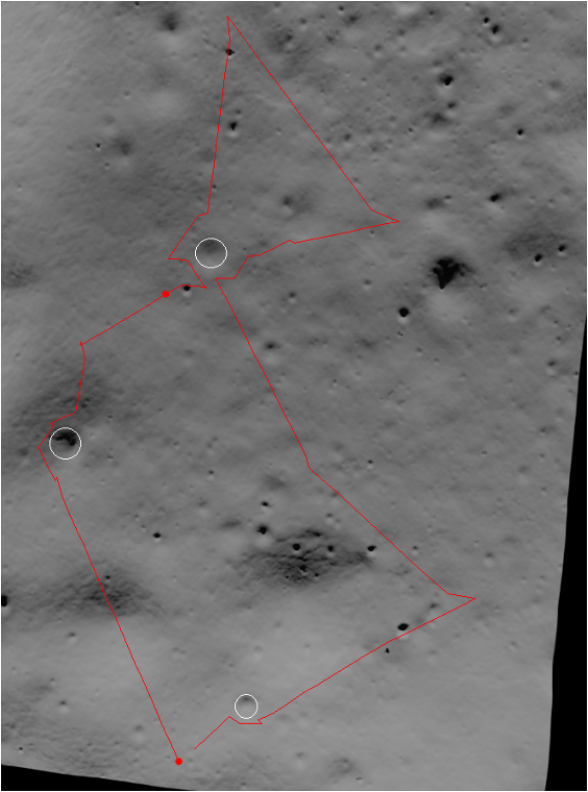}
    \caption{A \SI{1000}{\meter}+ simulated trajectory overlaid onto a simulated orbital map of the Lunar South Pole Region. The map contains ground truth landmark craters marked with white circles.}
    \label{fig:sim-gt-orbit}
\end{figure}

These trajectories ranged in length from \SI{536}{\meter} to \SI{840}{\meter} and contained two ground truth crater landmarks.
An additional trajectory was created in a different region where the trajectory path in one direction was \SI{1058}{\meter} and its reverse \SI{1211}{\meter}. The trajectory contained three ground truth crater landmarks and a top down view of this ground truth trajectory is in Figure \ref{fig:sim-gt-orbit}. 

\subsection{Cinder Lakes Apollo Training Area}\label{subsec:cl-data}
We collected datasets from Cinder Lakes near Flagstaff, Arizona.
This field test site was chosen as a Lunar analogue site as a realistic crater distribution of appropriate crater sizes was constructed during the Apollo era at two different sites:
\begin{enumerate}[noitemsep,topsep=0pt]
    \item The south site is~\SI{500}{\feet}-by-\SI{500}{\feet} in area and is well preserved as it is protected from motorized vehicles, but it does have areas of vegetation growth.
    \item The north site is 1200'-by-1200' and has almost no vegetation, but it is more heavily eroded due to being within an off-highway vehicle area.
\end{enumerate}

\begin{figure*}[t]
    \centering
    \subfloat[]{\includegraphics[width=0.29\textwidth]{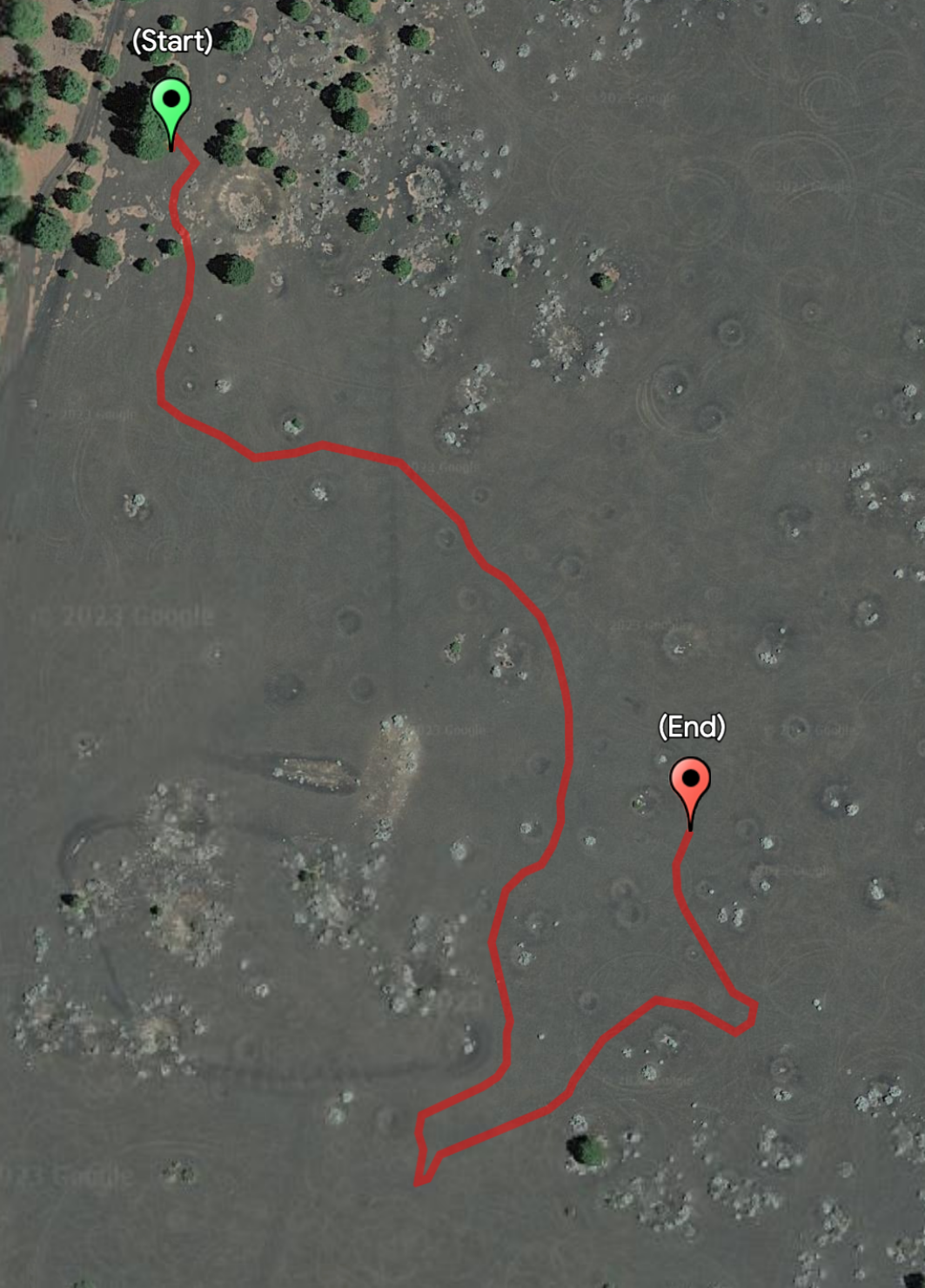}
    \label{fig:cl-south-vA}}
    \subfloat[]{\includegraphics[width=0.2878\textwidth]{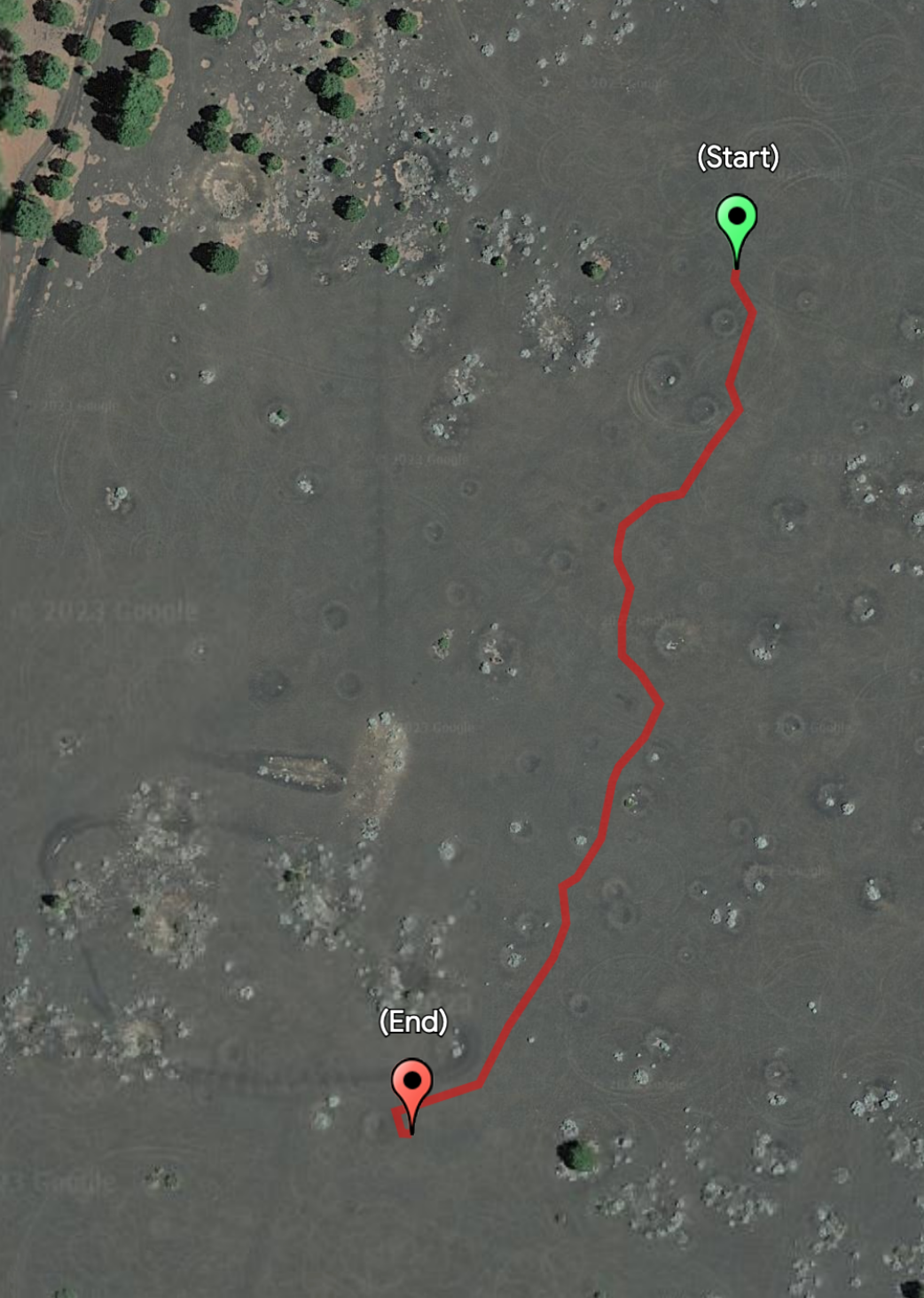}
    \label{fig:cl-south-NtoS}}
    \subfloat[]{\includegraphics[width=0.3875\textwidth]{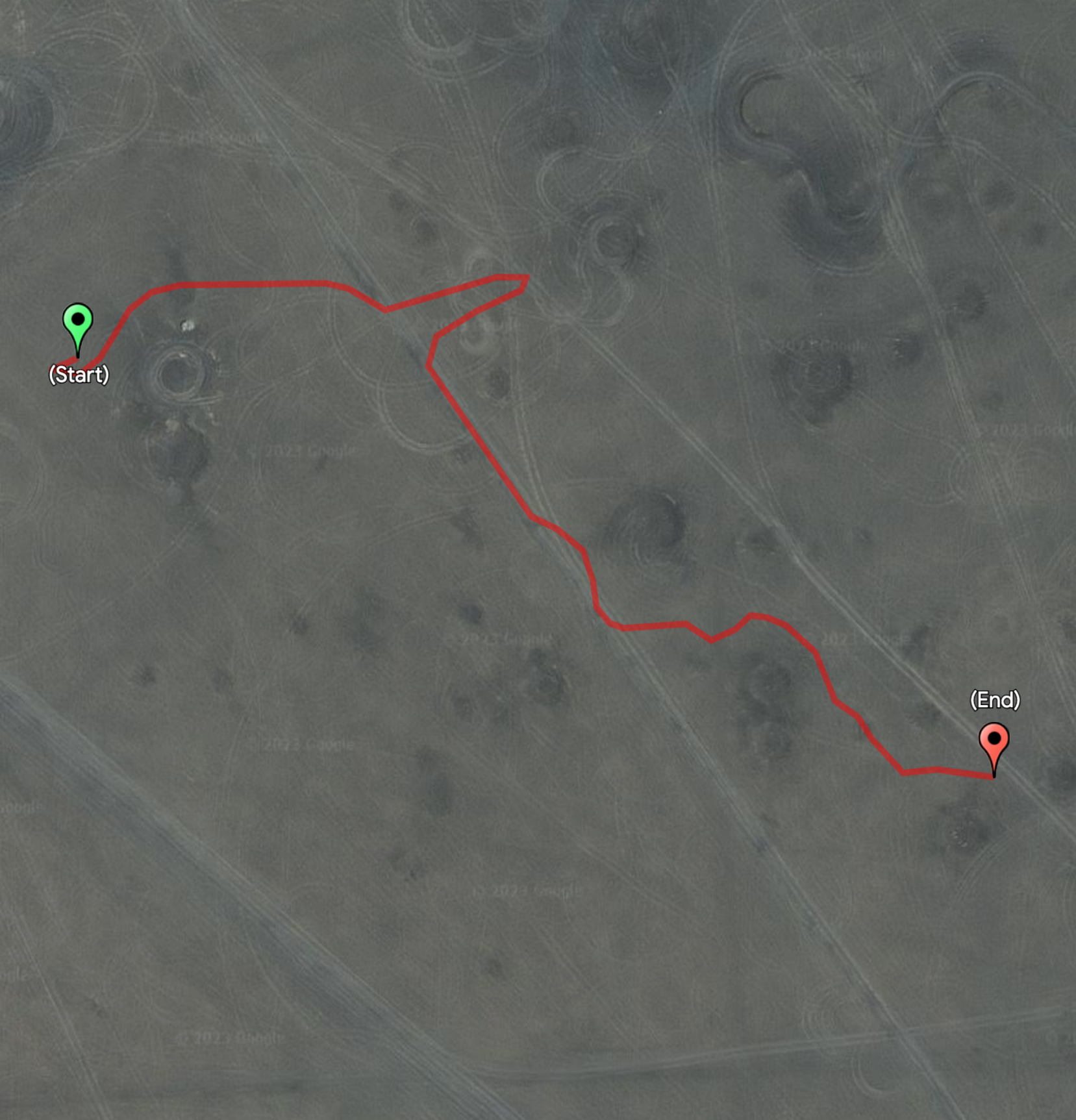}
    \label{fig:cl-north}}
    \caption{Three trajectories collected in the Cinder Lakes Apollo training area. (a) and (b) were collected in the better preserved South site and (c) was collected in the North site which is part of an \ac{ohv} area.}
    \label{fig:cl-trajectories}
\end{figure*}

Our datasets were collected during a New Moon, and given that Cinder Lakes is located outside of the dark sky town of Flagstaff, Arizona, the conditions allowed for an environment with minimal ambient light similar to what would be encountered in dark Lunar environments. 
We used a data collection rig consisting of a stereo camera with a \SI{0.26}{\meter} baseline, 5,120-by-3,840 pixel resolution, and 60$^\circ$ horizontal field-of-view as shown in Figure~\ref{fig:cl-rig}.
Additionally a \SI{50}{\watt}, \SI{5000}{\lumen} LED flood light was placed below the camera, with the light area ranging from \SI{0.1}{\meter}-\SI{0.2}{\meter} below the stereo cameras.
There was an \ac{ins} attached to the stereo cameras to provide ground truth position and orientation.

For the data collection, we performed two types of collections. 
The first type was geared toward perception tests.
This involved moving the data collection rig from approximately \SI{20}{\meter} away to \SI{5}{\meter} from the leading crater edge with stops for long exposures roughly every \SI{1}{\meter}. 
The second type was geared toward localization tests.
This entailed moving the data collection rig through the crater field with stops for long exposure approximately every \SI{5}{\meter}-\SI{10}{\meter}.
During these stops the rig was directed toward large craters if any were present. 
In total, in this work, 12 craters ranging from \SI{5}{\meter}-\SI{25}{\meter} in diameter and three trajectories were collected.
The three different trajectories are displayed in Figure \ref{fig:cl-trajectories} which shows the GPS path within Google Earth satellite imagery.
Of these, two trajectories were captured in the south site and one in the north site. 
In order to evaluate longer distances between craters, some experiments added drift error as if more distance had been traveled by artificially adding trajectory locations outside of the crater field at the start of a trajectory.

\section{RESULTS}~\label{sec:results}
In this section we detail the results of stereo, crater detection and matching, and absolute localization performance of our system. 

\subsection{Stereo}



Accurate stereo is required to be able to effectively detect and match crater rims within our approach. 
For detection, outliers in range values or stereo holes could lead to crater rim false positives.
Furthermore, inaccurate stereo ranges could lead to poor projections into the orbital map which could cause correctly identified crater rims in image-space to match poorly to orbital landmark craters.
To evaluate stereo performance, we created a simulated dataset of craters with ground truth ranges which was detailed in Section \ref{sec:sim-data}.

To evaluate stereo performance, three metrics based on~\cite{ScharsteinSzeliskiEtAl2001} binned across ground truth range values are used. 
These metrics are:
\begin{enumerate}[noitemsep,topsep=0pt]
\item The percent of pixels with a stereo hole where the corresponding ground truth pixel has a value within a specified range.
\item The percent detected pixels where a corresponding ground truth pixel has a value within a specified range that is not a hole but has a detected range whose error relative to ground truth value is greater than a threshold.
\item The \acp{rms} range error for all detected pixels that are not a hole and have a corresponding ground truth pixel within a specified range.
\end{enumerate}
We note that a crucial difference between the metrics used in~\cite{ScharsteinSzeliskiEtAl2001} is that range is used rather than disparity as range is what ultimately has the impact on crater matching. 

To validate the impact of hardware system parameters such as light location, camera height, and light power, these parameters were swept and stereo performance evaluated.
Light power is used as a proxy for exposure time and light brightness as these are not modelled well in Blender.
High light power corresponds to the near range of the image being fully exposed (pixel value of 255). 
A zero-mean 1.2 pixel intensity sigma Gaussian noise was added to all images to provide additional realism for nighttime imagery.
The results of this analysis are in Table~\ref{tab:sim-stereo}.
In these results, we observe that the optimal configuration for stereo is with a \SI{2.5}{\meter} camera off ground, light offset \SI{0.2}{\meter} below the camera, and high light power. 
Medium light power at a camera height of \SI{2.5}{\meter}, wherein the near range is bright but few pixels are fully exposed, demonstrates improved performance within \SI{30}{\meter} but at the slight cost of performance beyond \SI{30}{\meter}.
Moreover, although this configuration has robustness to camera height ranges, we observe that a camera height of~\SI{1.5}{\meter} performs slightly worse than at \SI{2.5}{\meter} and this is attributed to cameras closer to the ground losing vertical resolution with range.
Finally, a key takeaway is that placing the camera inline with the camera significantly degrades performance and placing the light above the camera renders extremely poor performance . 
We found that placing the light below the camera minimizes the worst impacts of the opposition effect and provides some shadowing on the surface for more visible features.

\begin{table*}[t]
\caption{Stereo performance evaluation using 4 different simulated crater samples across different parameters. ``Height'' is camera height above ground and ``Offset'' is the light offset relative to cameras. All metrics utilize range output from stereo. \%hole is percentage of pixels that are holes where there is ground truth. \%err is the percentage that have a range estimate greater than a threshold. RMS is the root-mean-square error in meters of reported range values. \revised{Using high or medium power light with a \SI{1.5}{\meter}-\SI{2.5}{\meter} camera height from ground is capable of computing stereo at ranges to \SI{40}{\meter} as long as the light source is offset below the camera.}}
\label{tab:sim-stereo}
\centering
\begin{adjustbox}{width=1\textwidth}
\begin{tabular}{|c|c|c|c|c|c|c|c|c|c|c|c|c|c|c|c|}
\hline
\multicolumn{4}{|c|}{} & \multicolumn{3}{|c|}{0-10m Range} & \multicolumn{3}{|c|}{10-20m Range} & \multicolumn{3}{|c|}{20-30m Range} & \multicolumn{3}{|c|}{30-40m Range} \\
\hline
 & Height & Offset & Light & & \%err & RMS & & \%err & RMS & & \%err & RMS & & \%err & RMS \\
Algo. & (m) & (m) & Power & \%hole & $>$0.5m & (m) & \%hole & $>$1m & (m) & \%hole & $>$2m & (m) & \%hole & $>$3m & (m) \\
\hline
SGBM & 2.5 & -0.2 & high & 6.82 & 1.27 & 0.28 & 11.94 & 18.06 & 1.02 & 34.67 & 20.85 & 1.47 & 82.84 & 8.90 & 1.47 \\
SGBM & 2.5 & -0.2 & med & 3.68 & 1.07 & 0.28 & 11.76 & 15.34 & 0.77 & 38.64 & 20.30 & 1.37 & 85.22 & 14.93 & 1.74 \\
SGBM & 2.5 & -0.2 & low & 4.04 & 0.96 & 0.28 & 23.99 & 16.79 & 0.64 & 73.98 & 25.77 & 1.57 & 98.05 & 45.26 & 3.75 \\
SGBM & 2.5 & 0.0 & med & 11.38 & 2.54 & 0.36 & 31.60 & 18.60 & 0.98 & 55.52 & 18.08 & 1.29 & 89.84 & 11.48 & 1.51 \\
SGBM & 2.5 & 0.2 & med & 74.13 & 16.50 & 1.34 & 60.97 & 25.64 & 1.45 & 69.53 & 17.81 & 1.28 & 93.26 & 11.96 & 1.53 \\
SGBM & 1.5 & -0.2 & med & 6.71 & 1.58 & 0.22 & 27.53 & 24.94 & 1.15 & 50.30 & 21.78 & 1.48 & 88.38 & 21.51 & 2.34 \\
\hline
\end{tabular}
\end{adjustbox}
\end{table*}

\subsection{Crater Detection and Matching}
To evaluate crater detection and matching, we used the real data collected at Cinder Lakes detailed in Section~\ref{subsec:cl-data}.
Additional simulated data was collected to evaluate stereo and positional errors as detailed in Section~\ref{sec:sim-data}.
We use the percent of the ground truth front rim detected as a metric to evaluate detection performance:
\begin{equation}
\label{eq:rim-percent}
\%_{\textrm{rim}_\textrm{det}} = \frac{px_{\textrm{gt}_\textrm{matched}}}{px_\textrm{gt}} * 100
\end{equation}
,where $px_{\textrm{gt}_\textrm{matched}}$ is the number of ground truth crater points in world frame that have a detection that is closest to that ground truth point and $px_\textrm{gt}$ is the total number of all ground truth points.
This detection does not consider false positives.
To evaluate crater matching performance, we use the Q-Score metric detailed in Section \ref{subsubsec:q_score} to evaluate how closely a perception sample aligns to orbital ground truth considering both true detections and false positives.

\begin{figure}[t]
    \centering
    \subfloat[]{\includegraphics[width=0.48\columnwidth]{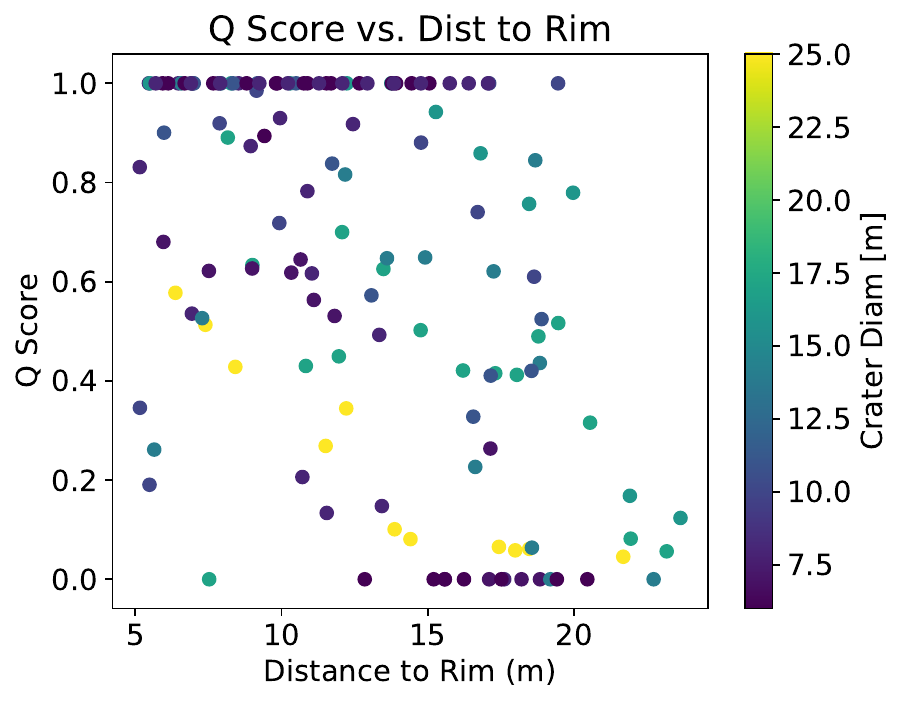}
    \label{fig:cl-q-sgbm}}
    \subfloat[]{\includegraphics[width=0.48\columnwidth]{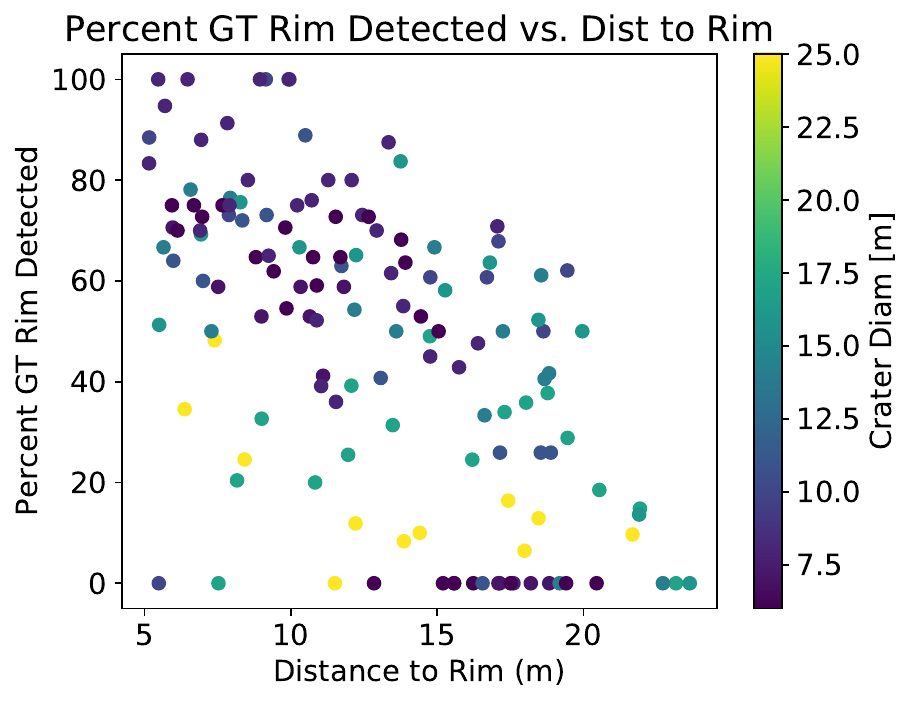}
    \label{fig:cl-percent-sgbm}}
    \caption{Cinder Lakes crater detection and matching results using SGBM and discontinuity detection. (a) Q-Scores. (b) Percent front arc of ground truth crater detected. \revised{The majority of Q-Scores for all but the smallest of $<$\SI{7.5}{\meter} (purple) and the largest $>$\SI{20}{\meter} (yellow) have Q-Scores above a target of 0.4 within 20 meters. Nearly all of the crater rims are detected at some percentage within \SI{15}{\meter} and to \SI{20}{\meter} for craters that are larger than \SI{7.5}{\meter} and smaller than \SI{20}{\meter} in diameter. Overall craters of diameters between \SI{7.5}{\meter} and \SI{20}{\meter} can be detected reliably with potential for lower accuracy at increased range.}}
    \label{fig:cl-q-percent}
\end{figure}

\ac{sgbm} stereo and discontinuity detection were run across the Cinder Lakes crater dataset to evaluate crater detection and matching performance.
Figure~\ref{fig:cl-q-percent} shows the results with respect to percent of front rim detected and Q-Scores versus range and crater diameter.
From these results, we observe that nearly all craters have some level of detection between \SI{5}{\meter}-\SI{15}{\meter}. 
However, at \SI{15}{\meter} the \SI{7}{\meter}-crater is no longer detected.
Furthermore, the \SI{25}{\meter} has worse results across most of the set of camera distances from the crater rim.
The reduction in performance for smaller and larger craters is expected with our current approach.
A larger crater diameter requires stereo to obtain matches at farther ranges to observe the leading and trailing edge of a discontinuity. 
A smaller crater diameter limits the size of a discontinuity, which in turns limits the ability to detect it with few false positives. 
Overall, we observe robust filter performance, with most samples containing a Q-Score of 0.4 or better which means that the average pixel detection is within \SI{2}{\meter} of the crater rim and indicates low false positives and high accuracy.

\begin{figure}[ht]
    \centering
    \subfloat[]{\includegraphics[width=0.48\columnwidth]{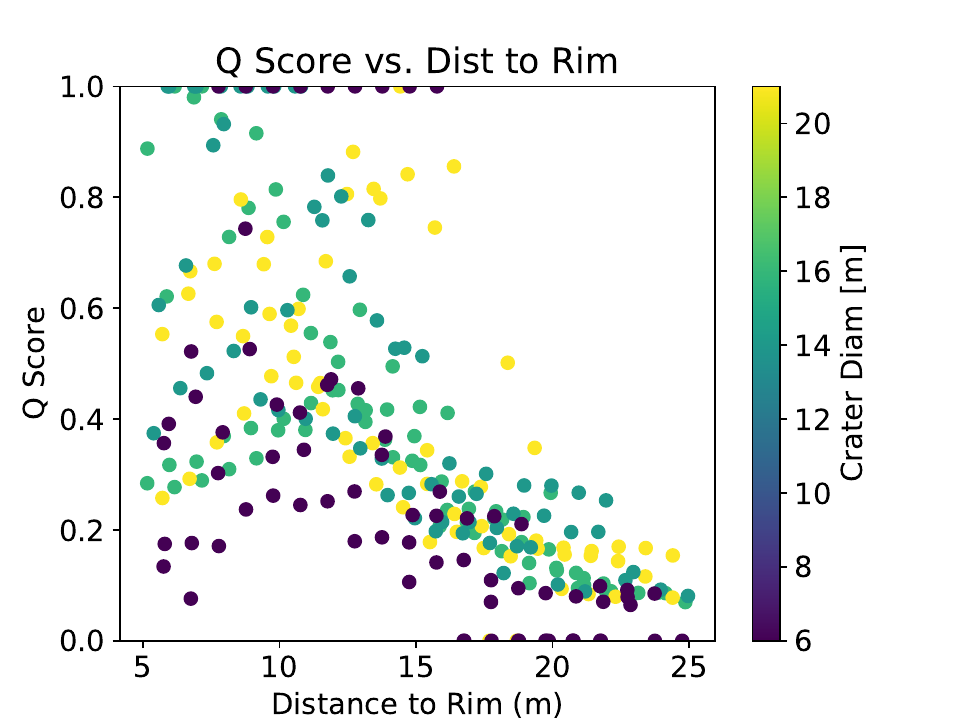}
    \label{fig:sim-q-nonoise}}
    \subfloat[]{\includegraphics[width=0.48\columnwidth]{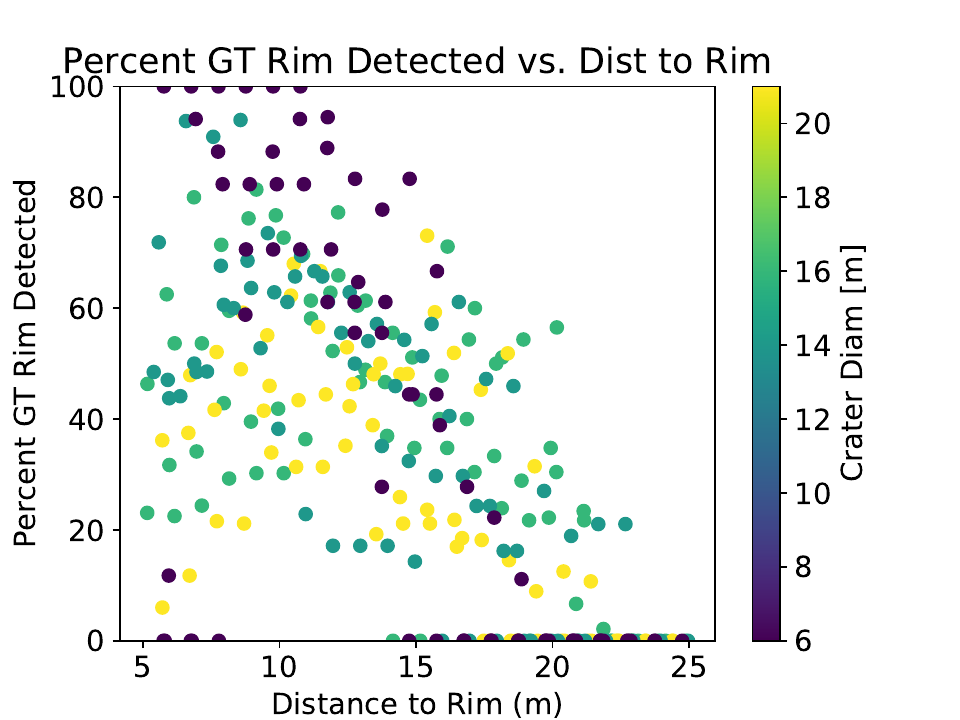}
    \label{fig:sim-percent-nonoise}} \\
    \subfloat[]{\includegraphics[width=0.48\columnwidth]{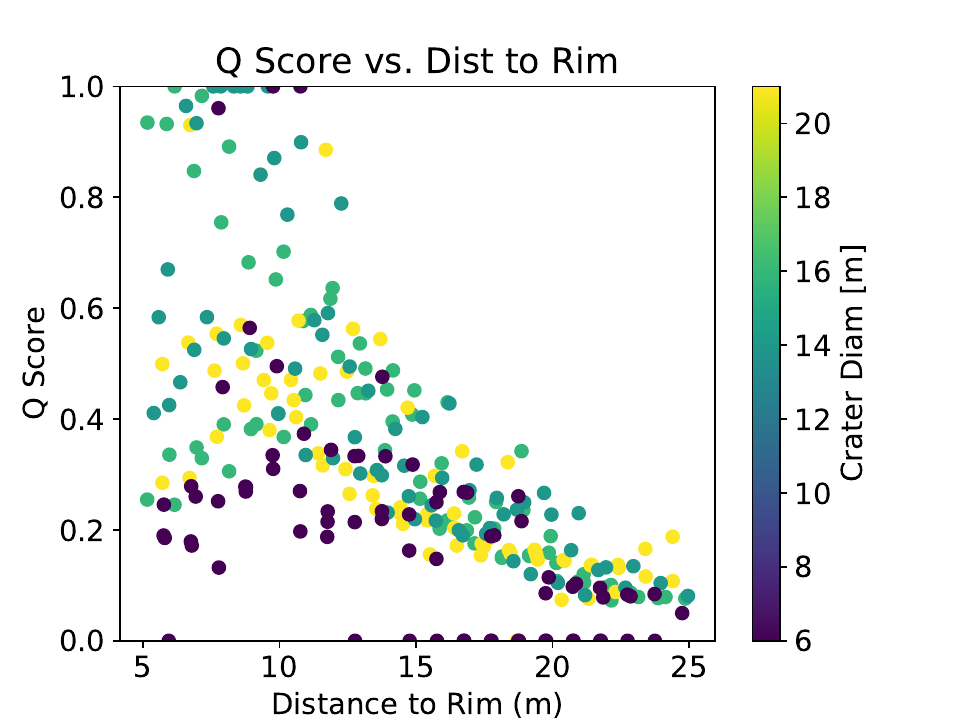}
    \label{fig:sim-q-noise}}
    \subfloat[]{\includegraphics[width=0.48\columnwidth]{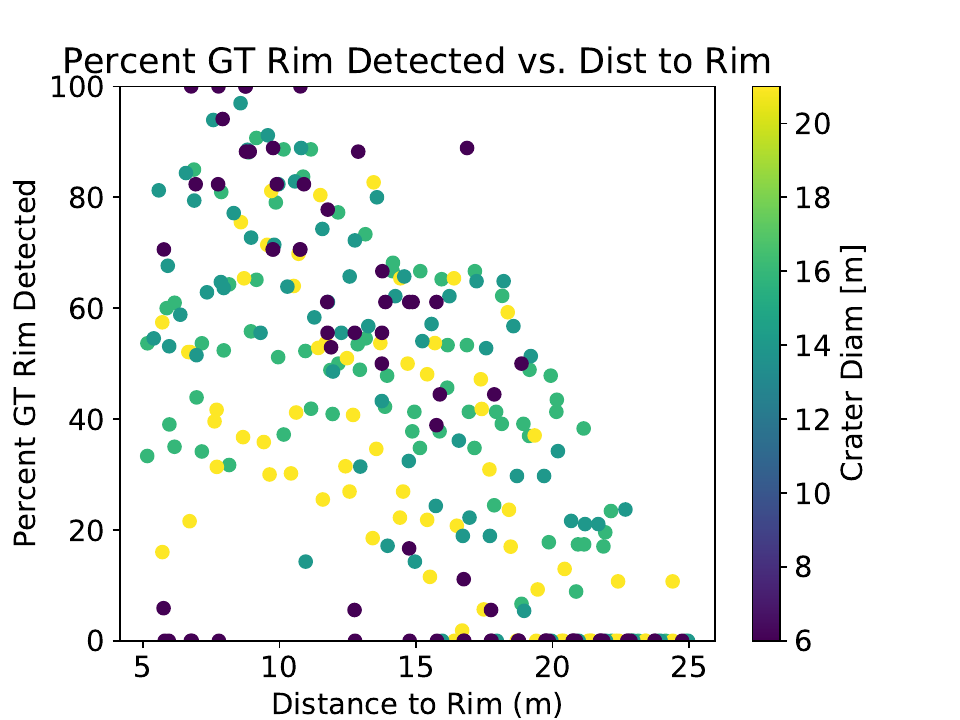}
    \label{fig:sim-percent-noise}}
    \caption{Crater detection and matching results with an without Gaussian noise added to simulated images on data with a camera height of 2.5m a light offset of 20cm below camera and high exposure. For images with noise added, 0 mean and 1.2 pixel intensity sigma were used. (a) Q-Score no noise (b) Percent detected no noise (c) Q-Score 1.2 sigma noise (d) Percent detected 1.2 sigma noise. \revised{The Q-Score and percentages of crater rim detected results share similar patterns with and without noise added to the simulated images highlighting the robustness of the detection approach to potential noise present in real imagery.}}
    \label{fig:sim-noise-eval}
\end{figure}

\begin{figure*}[!h]
    \centering
    \subfloat[]{\includegraphics[width=0.24\textwidth]{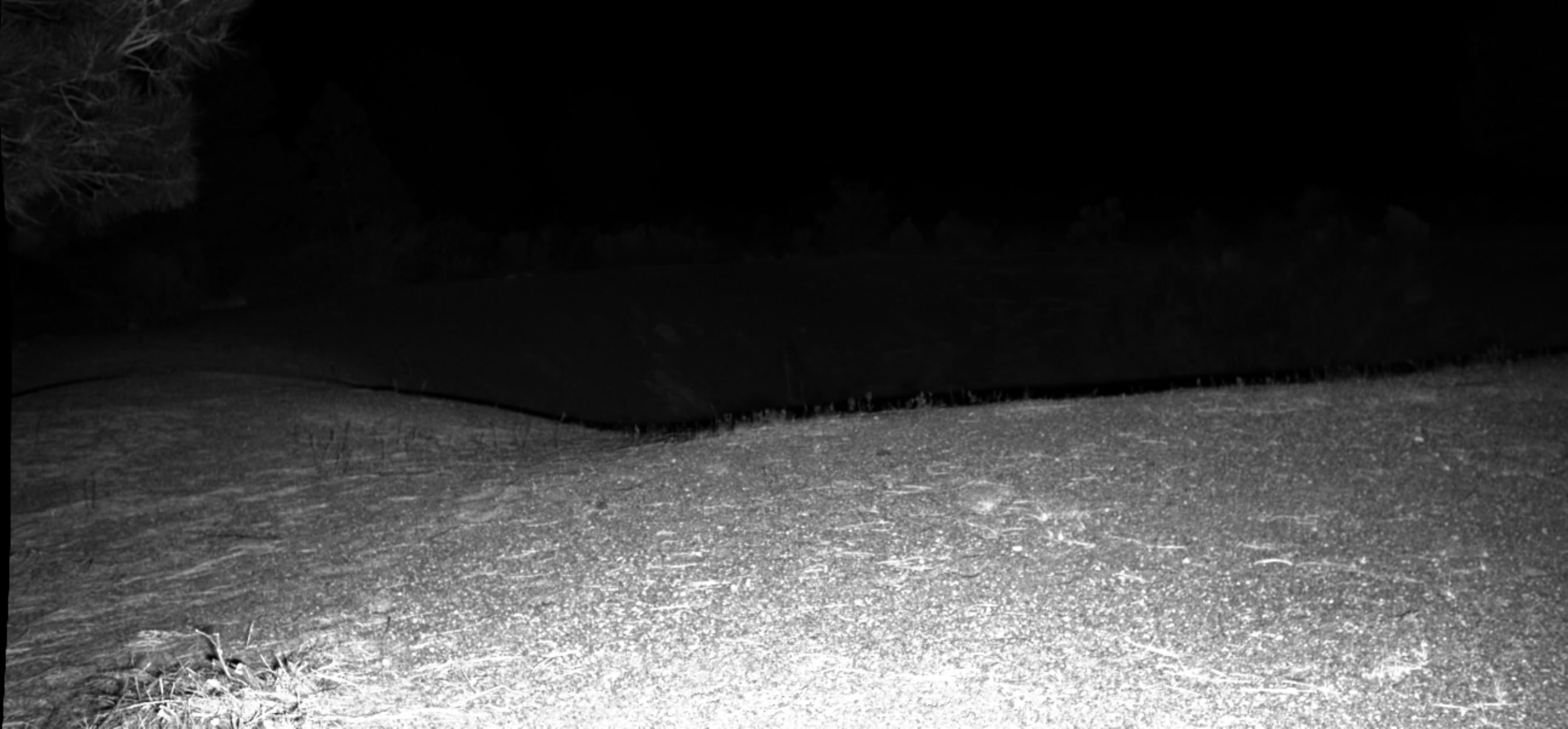}}
    \subfloat[]{\includegraphics[width=0.24\textwidth]{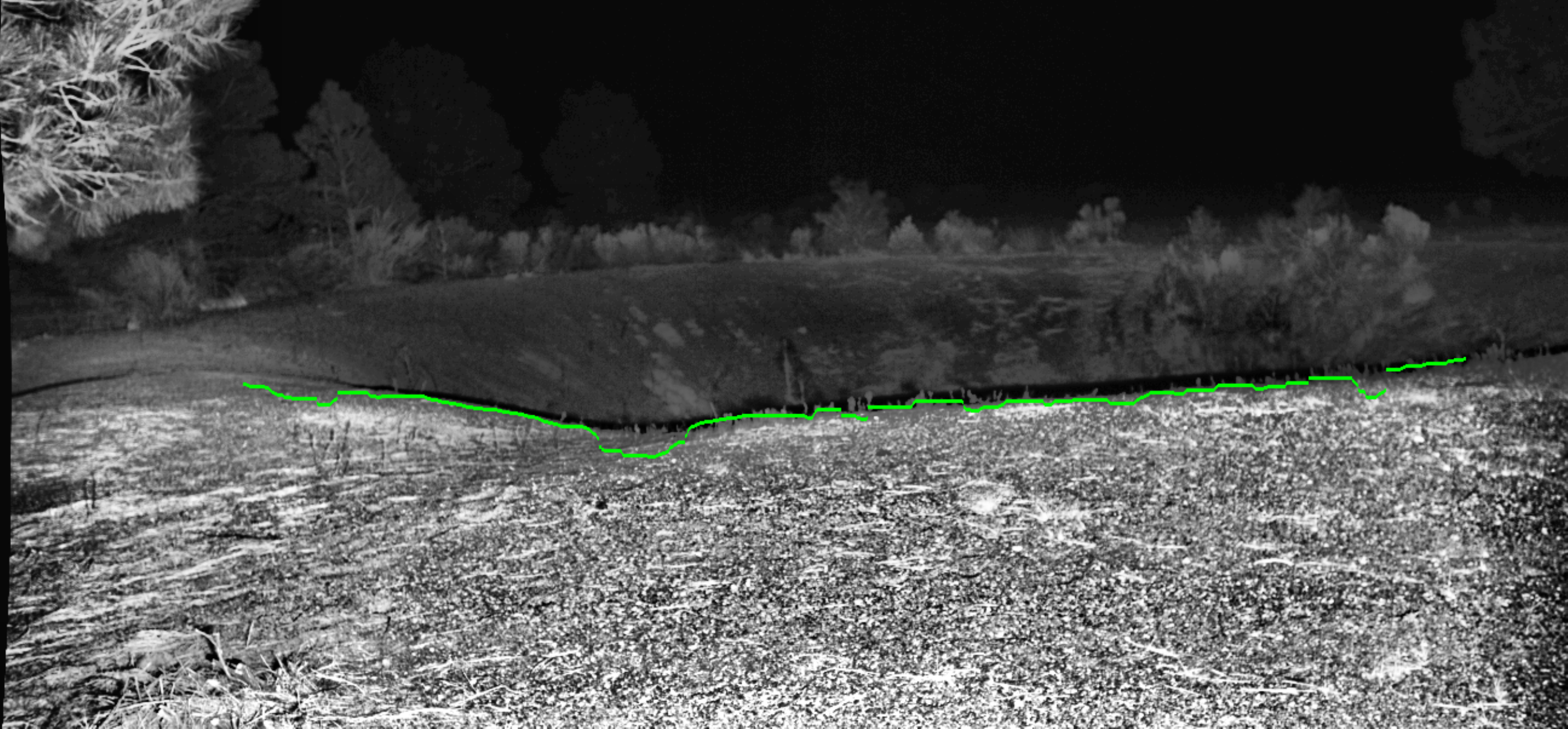}}
    \subfloat[]{\includegraphics[width=0.24\textwidth]{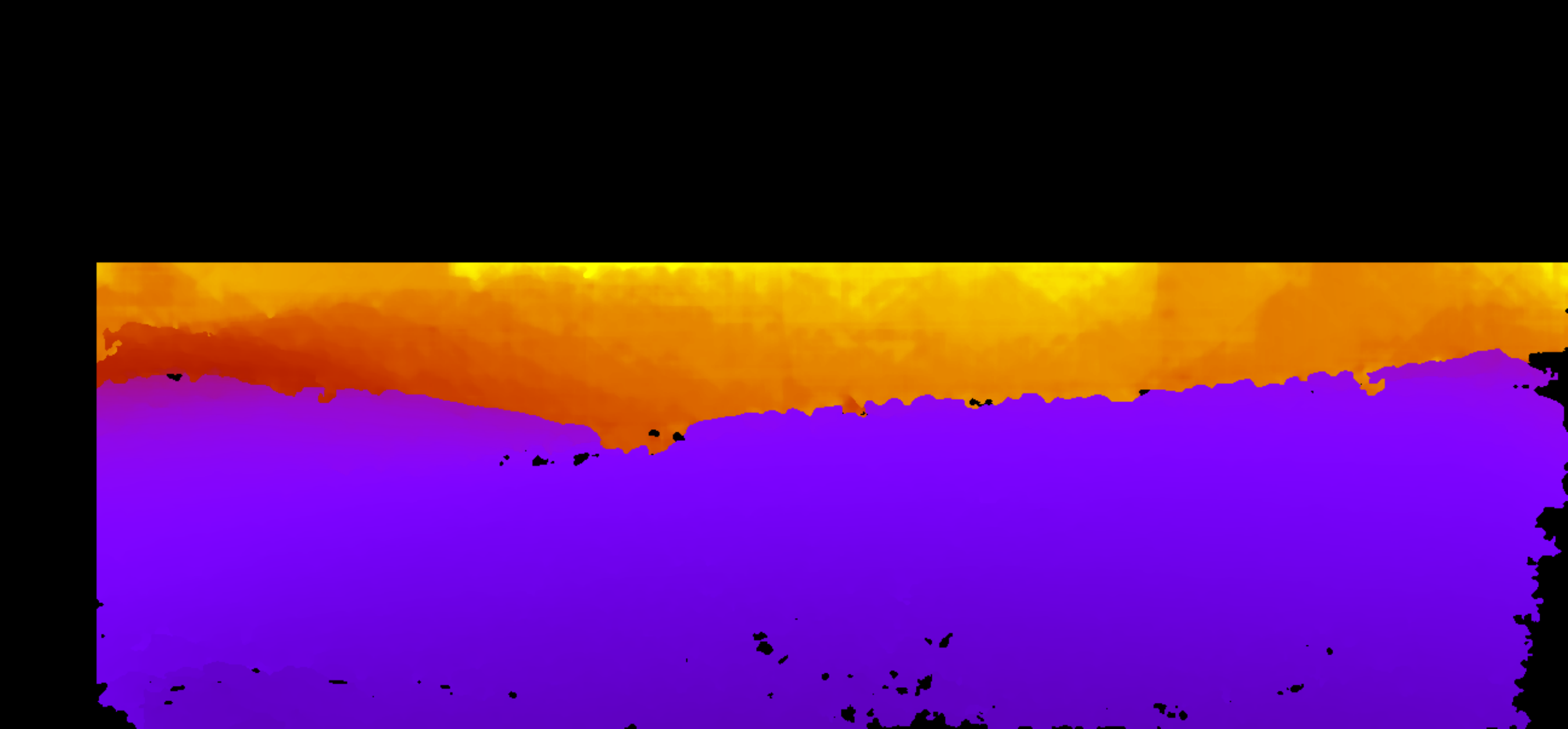}}
    \subfloat[]{\includegraphics[width=0.24\textwidth]{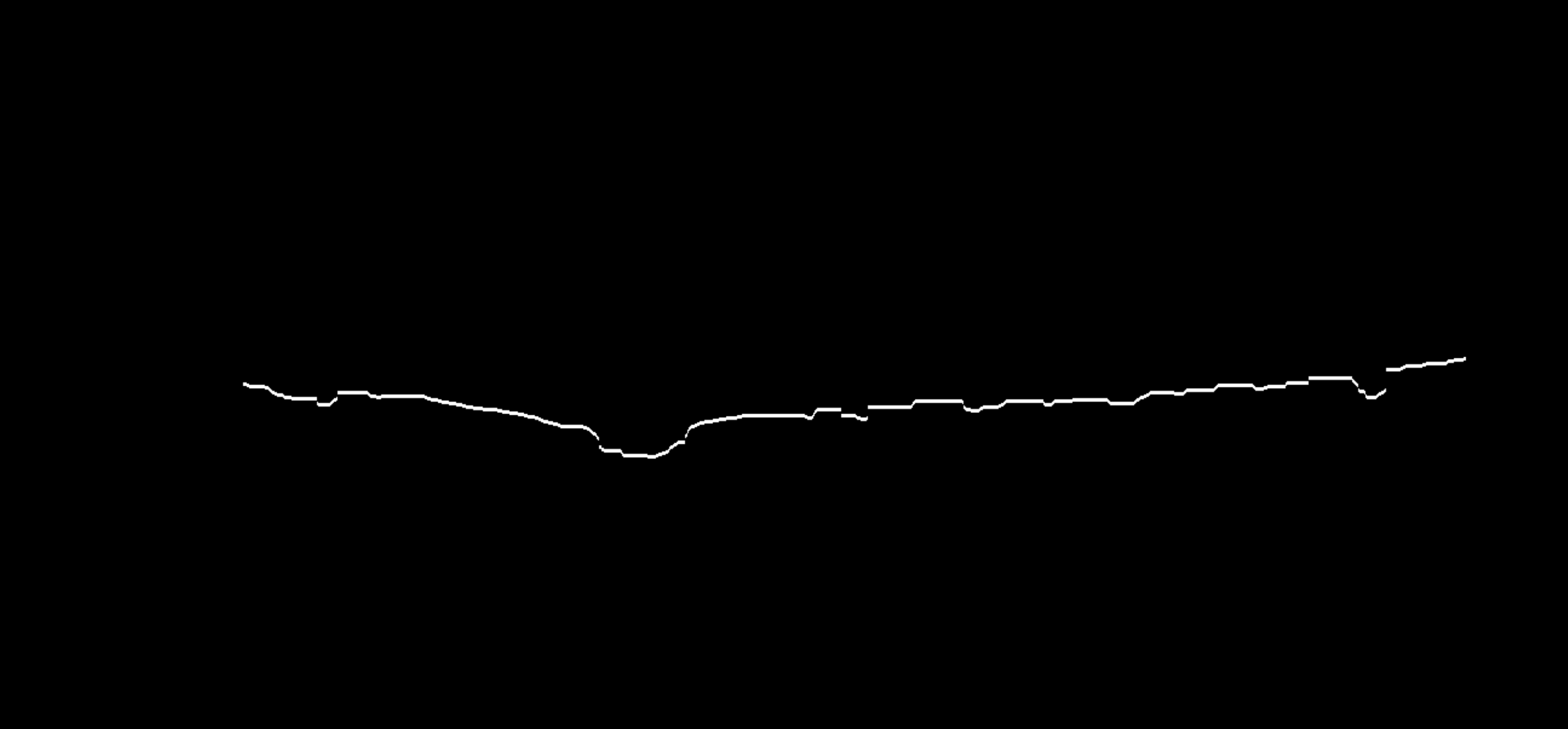}}

    \subfloat[]{\includegraphics[width=0.24\textwidth]{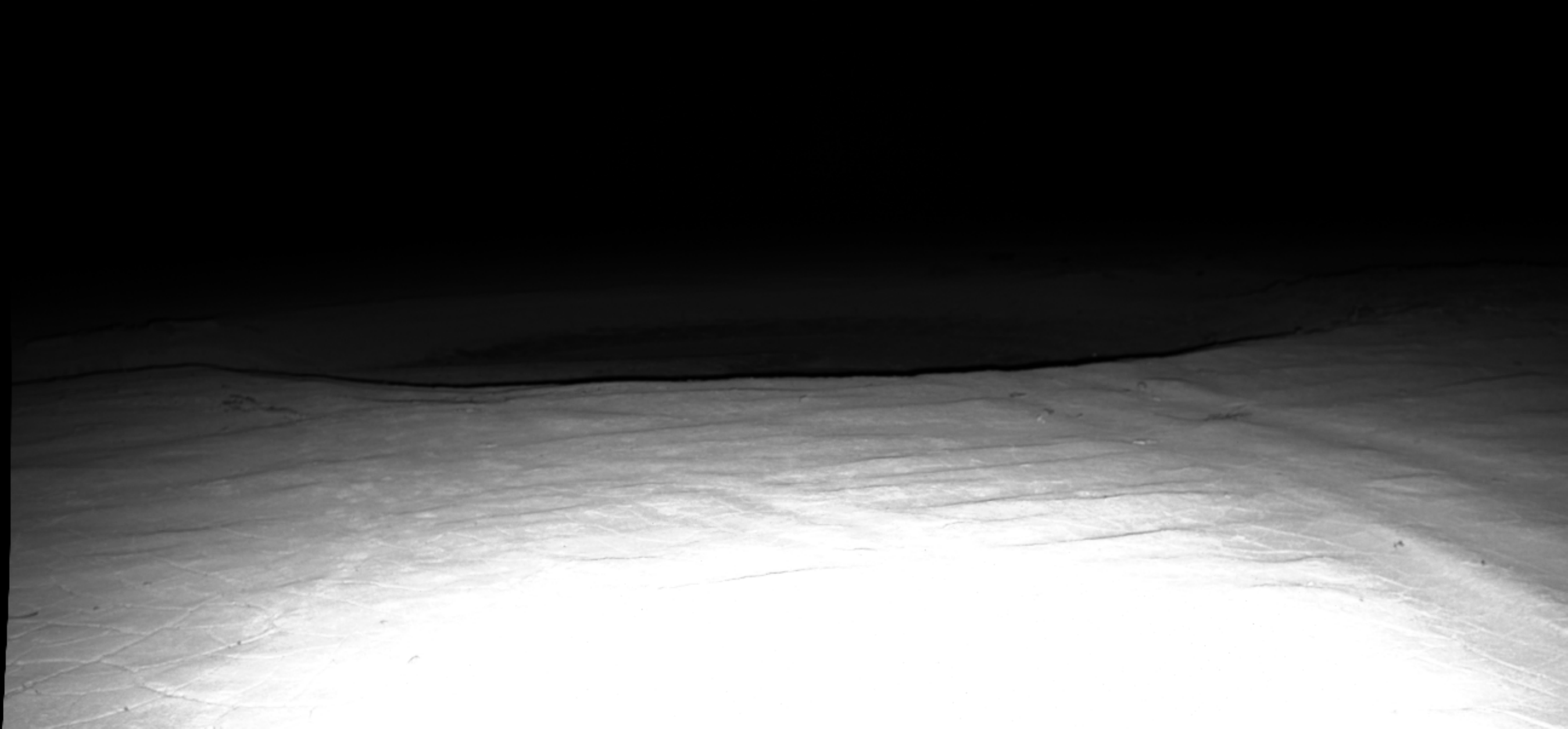}}
    \subfloat[]{\includegraphics[width=0.24\textwidth]{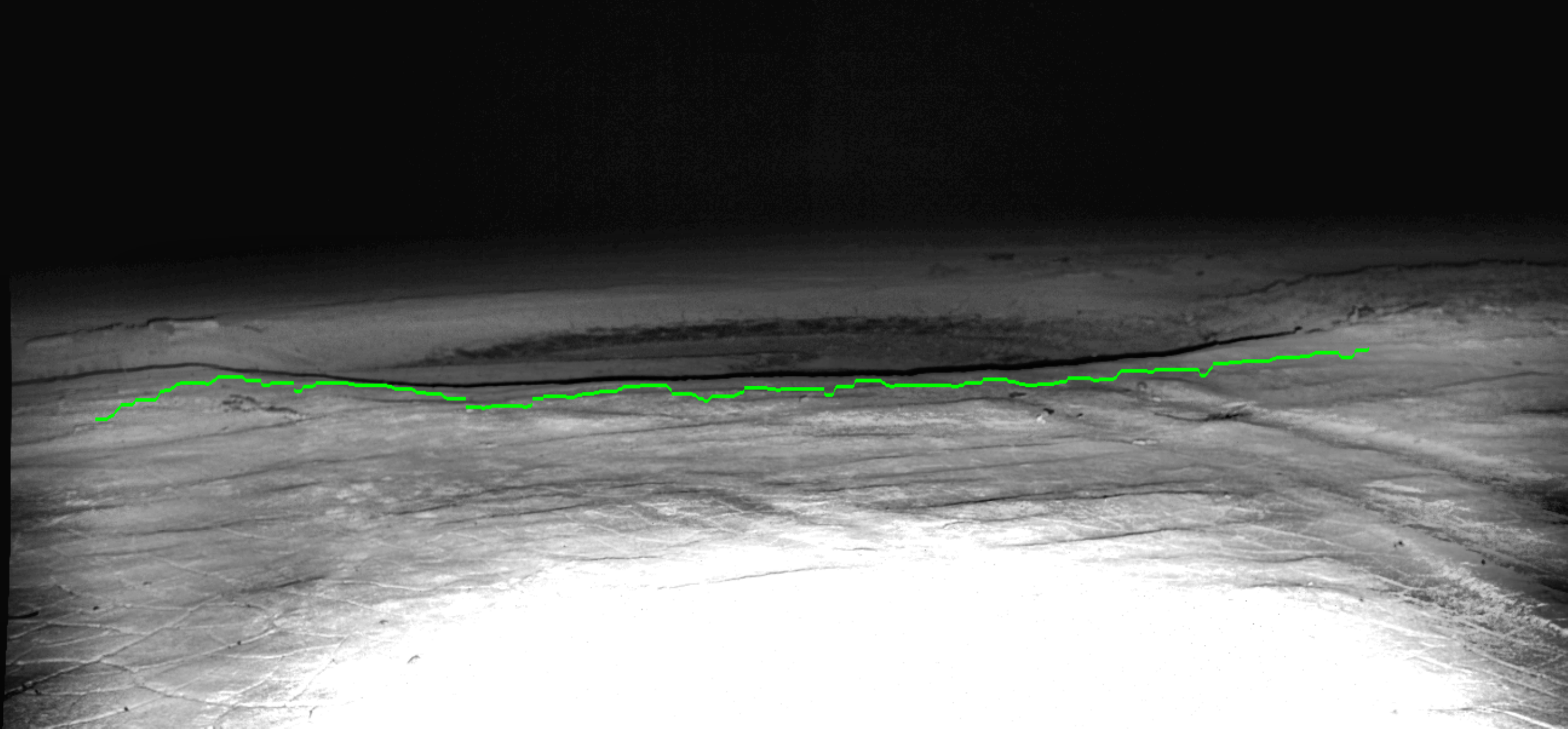}}
    \subfloat[]{\includegraphics[width=0.24\textwidth]{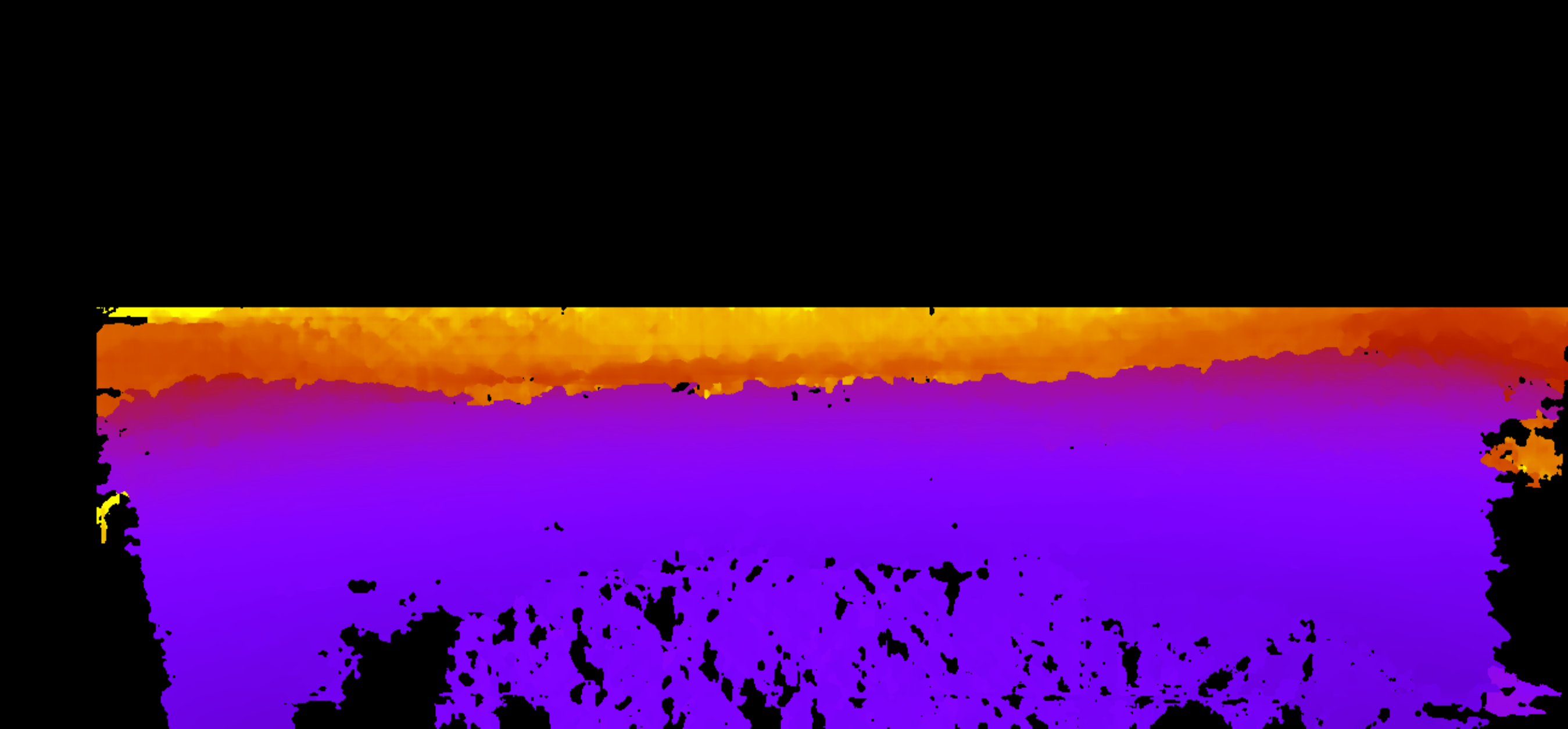}}
    \subfloat[]{\includegraphics[width=0.24\textwidth]{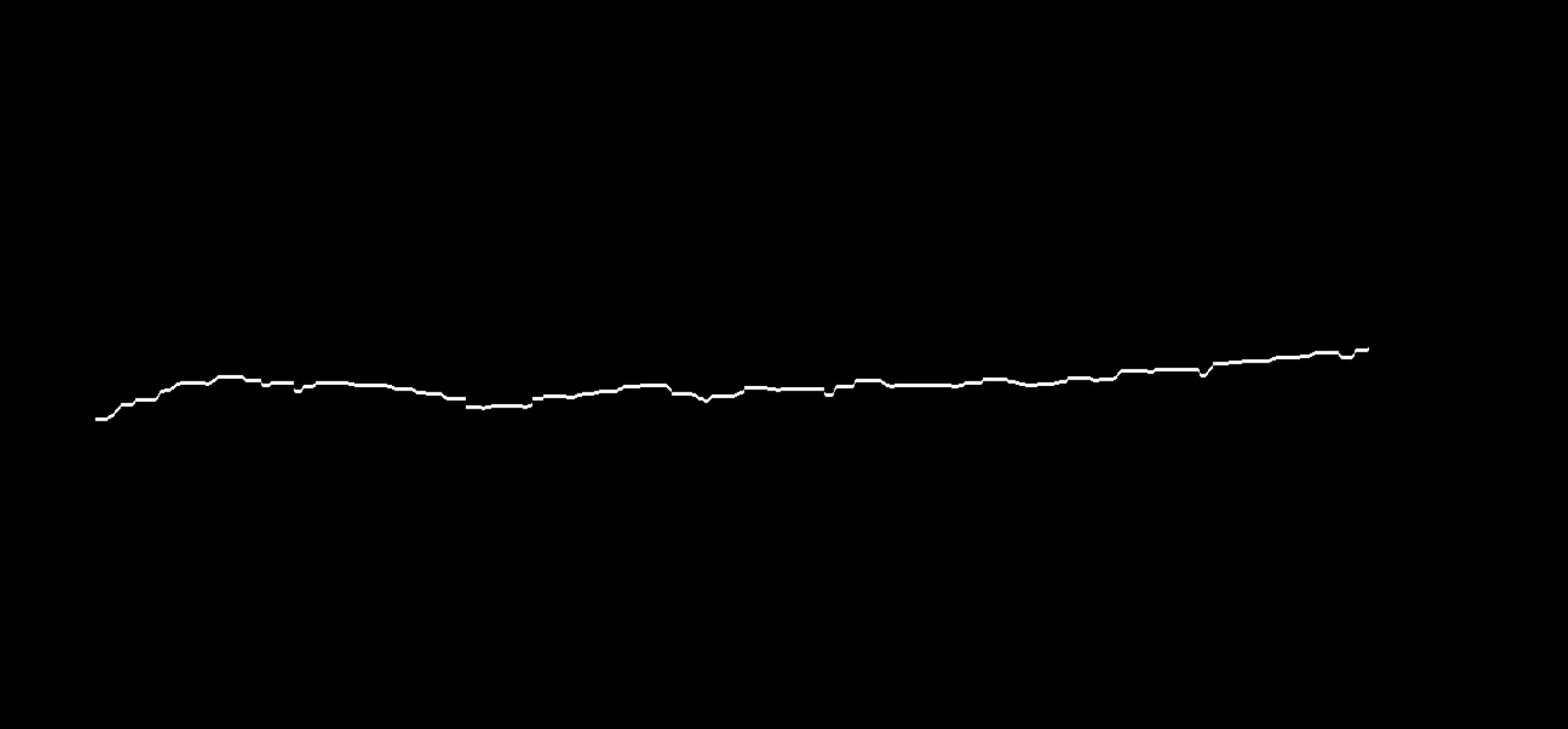}}

    \subfloat[]{\includegraphics[width=0.24\textwidth]{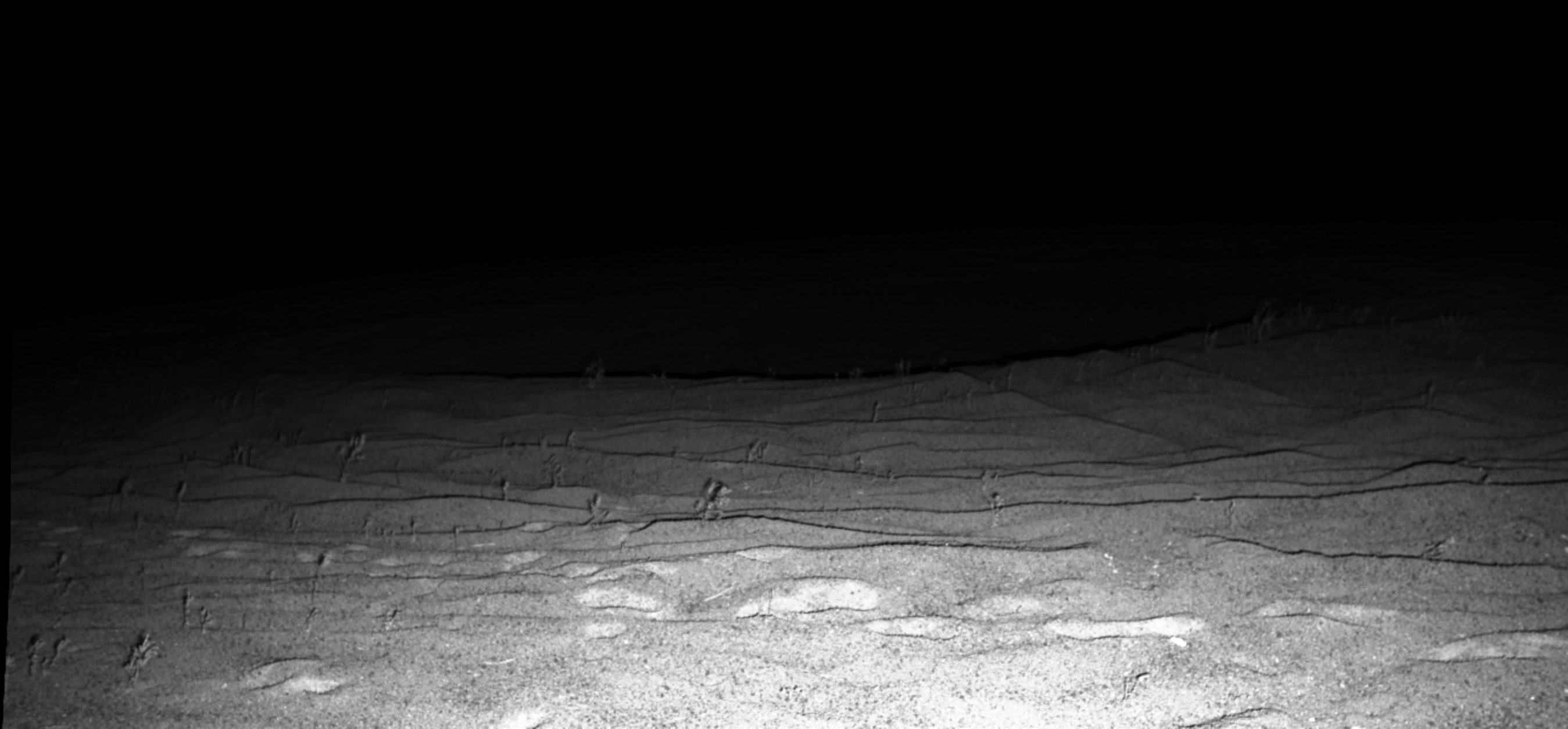}}
    \subfloat[]{\includegraphics[width=0.24\textwidth]{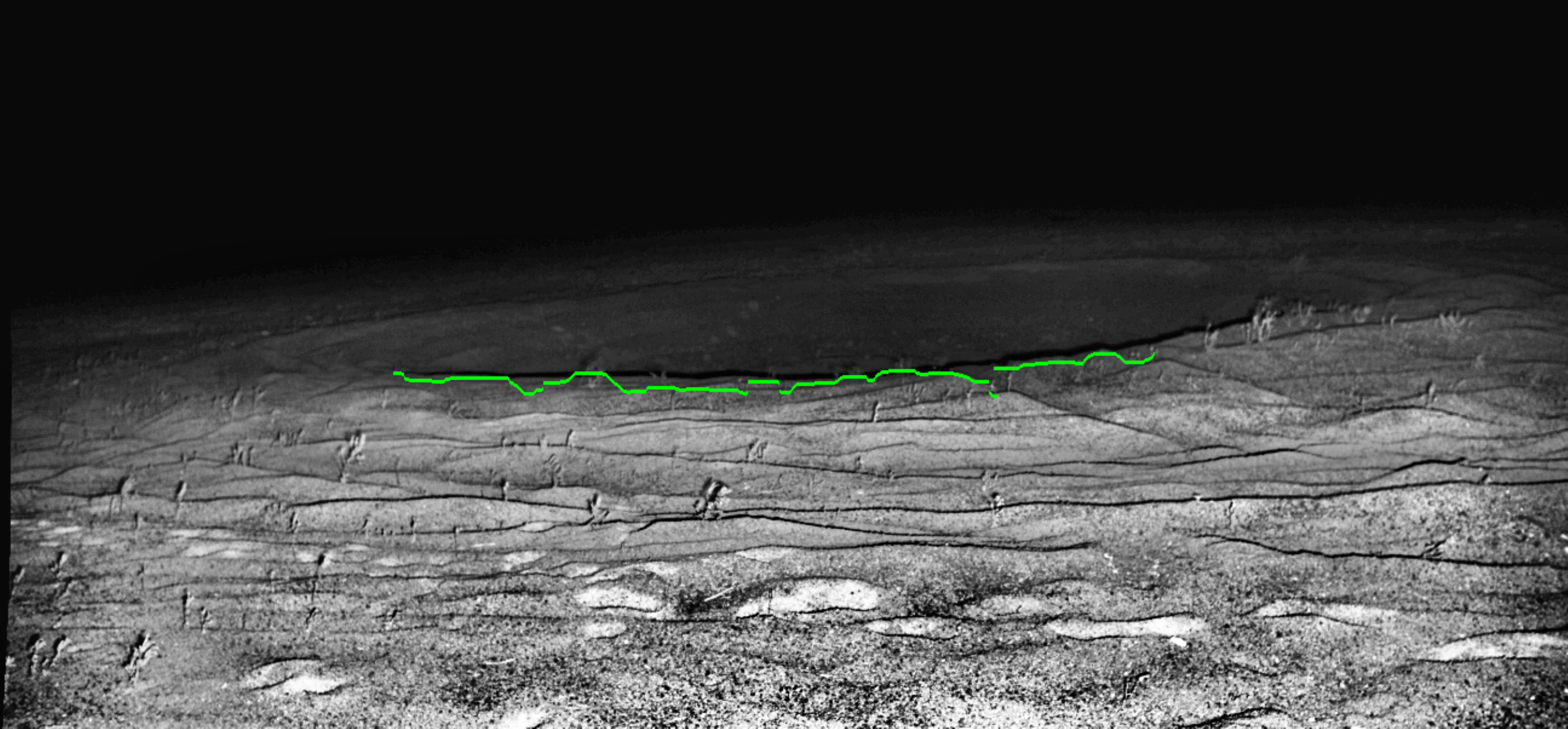}}
    \subfloat[]{\includegraphics[width=0.24\textwidth]{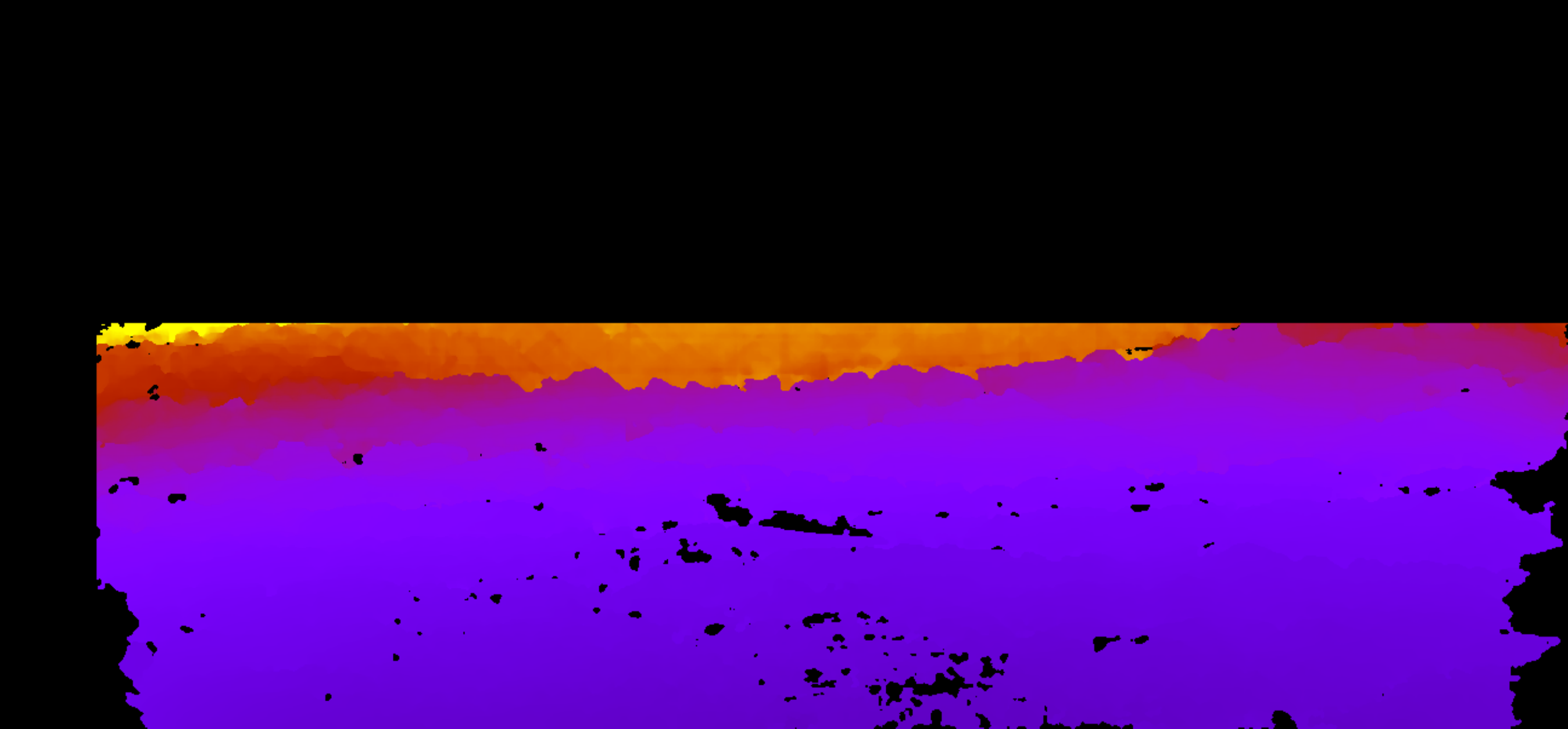}}
    \subfloat[]{\includegraphics[width=0.24\textwidth]{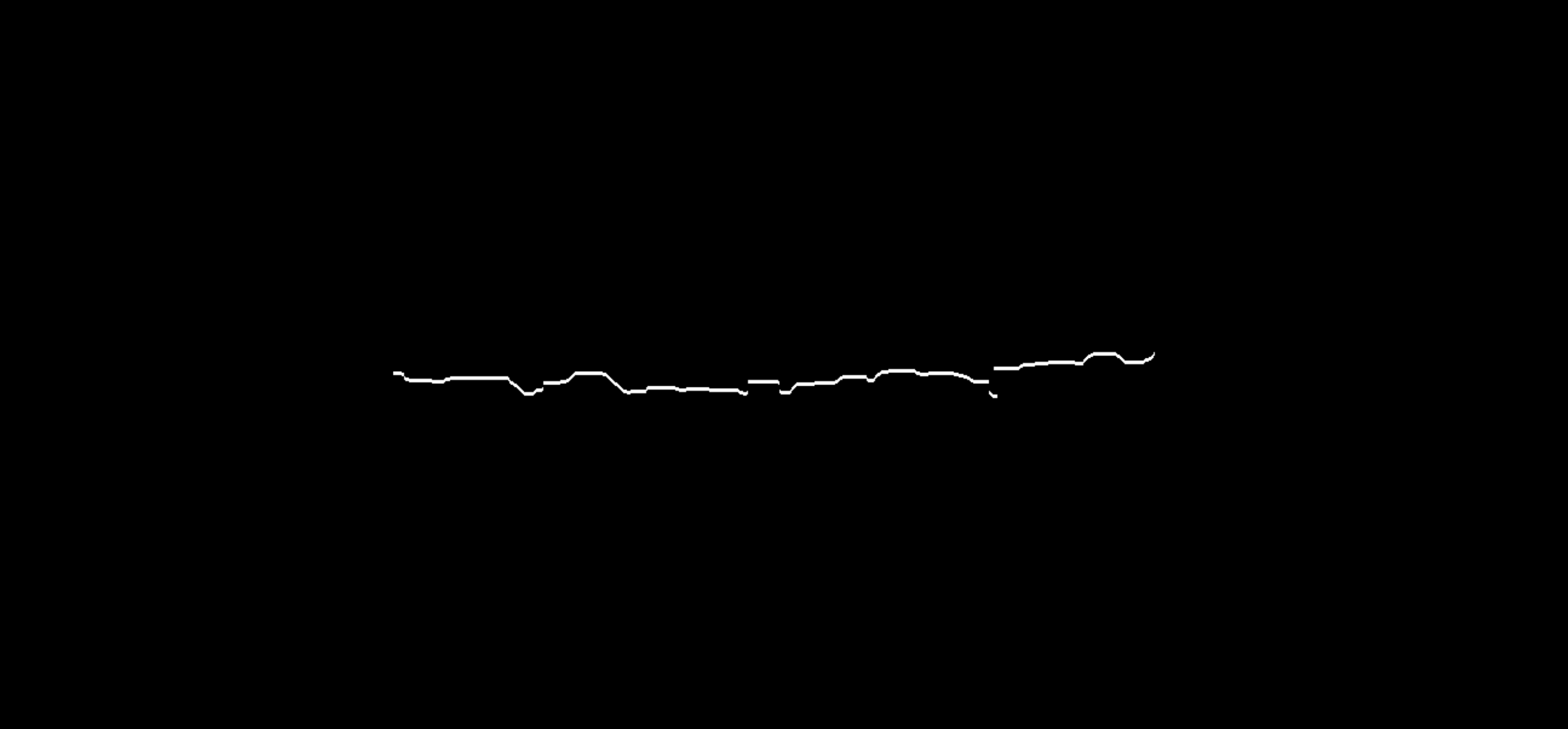}}

    \subfloat[]{\includegraphics[width=0.24\textwidth]{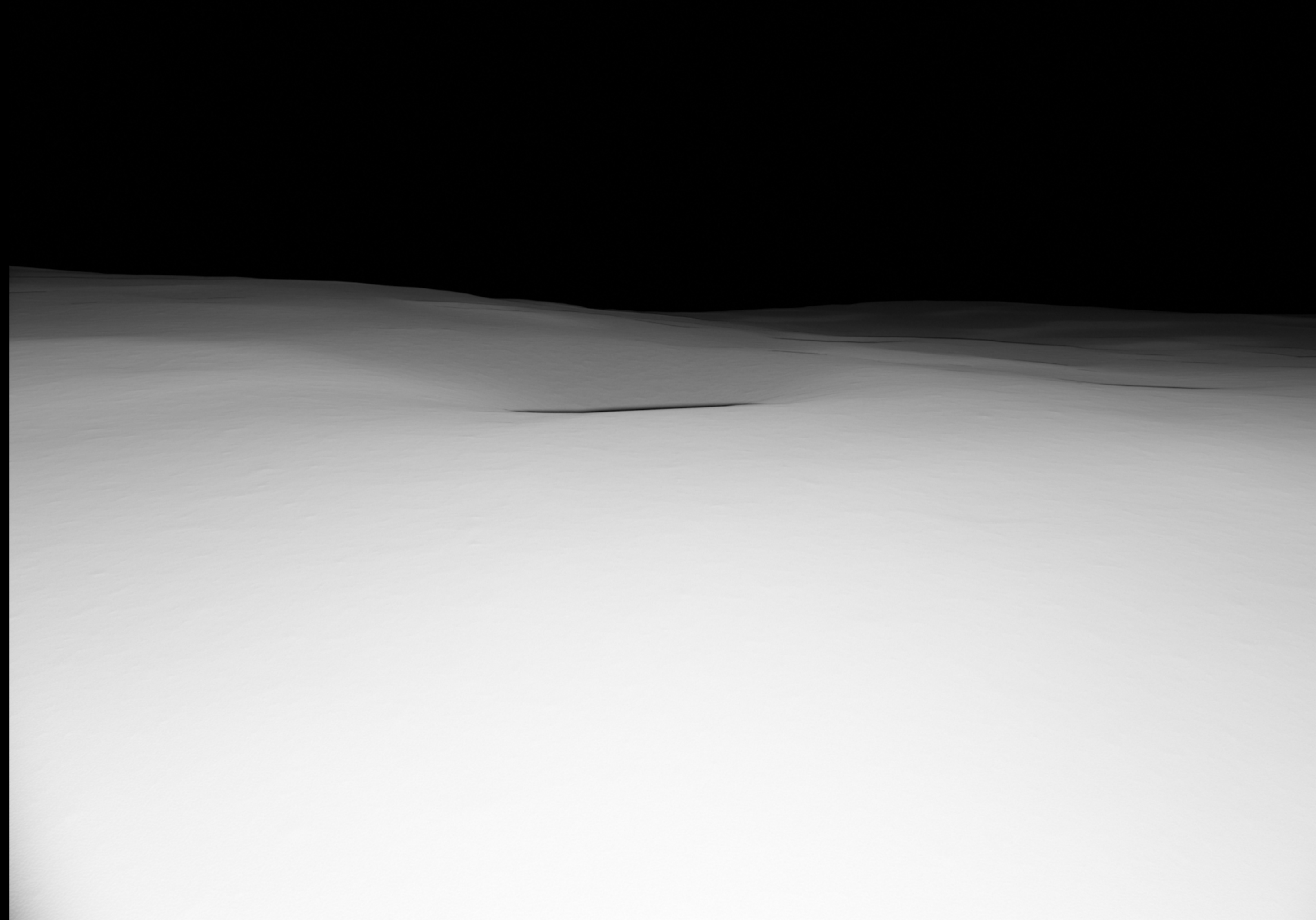}}
    \subfloat[]{\includegraphics[width=0.24\textwidth]{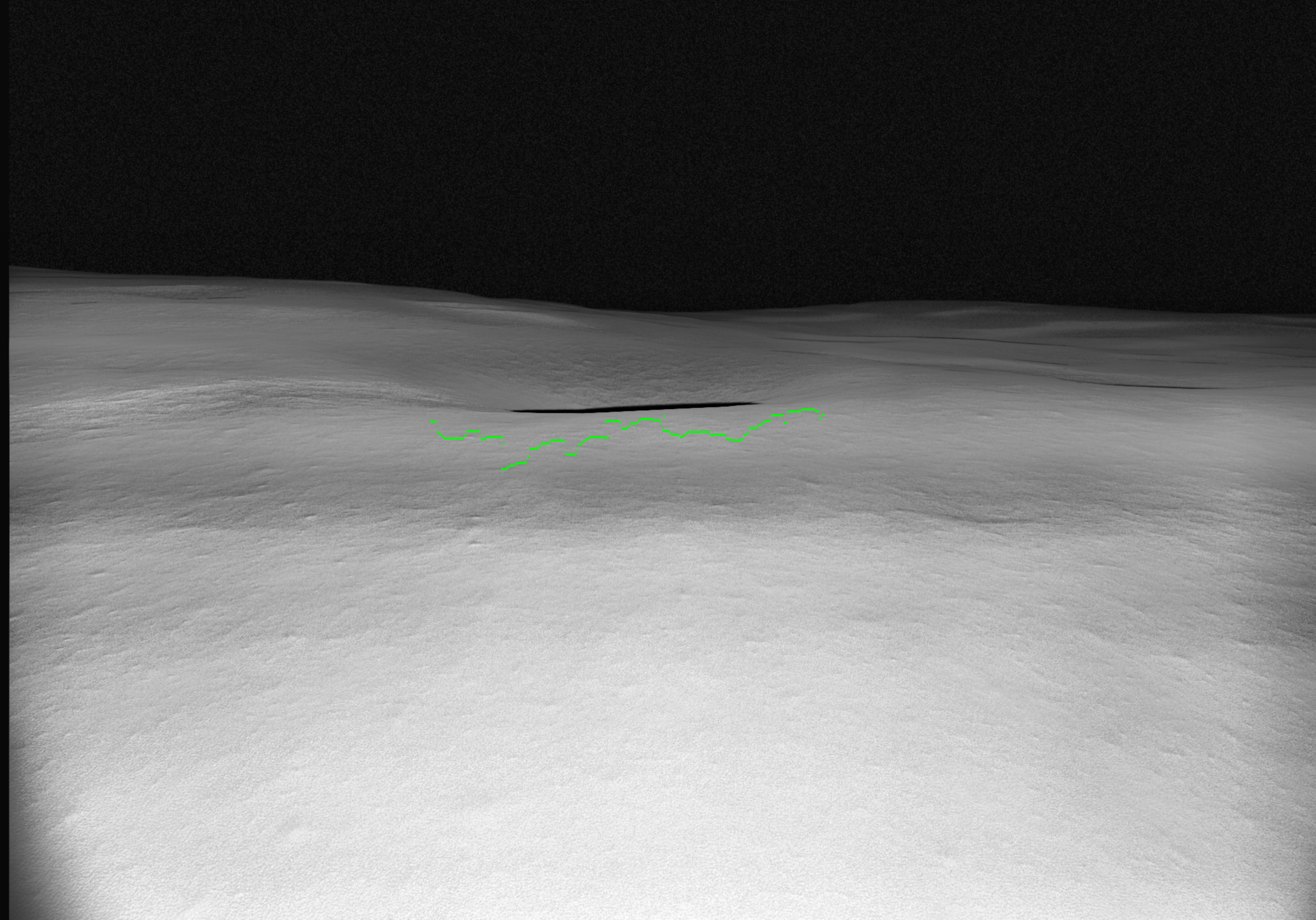}}
    \subfloat[]{\includegraphics[width=0.24\textwidth]{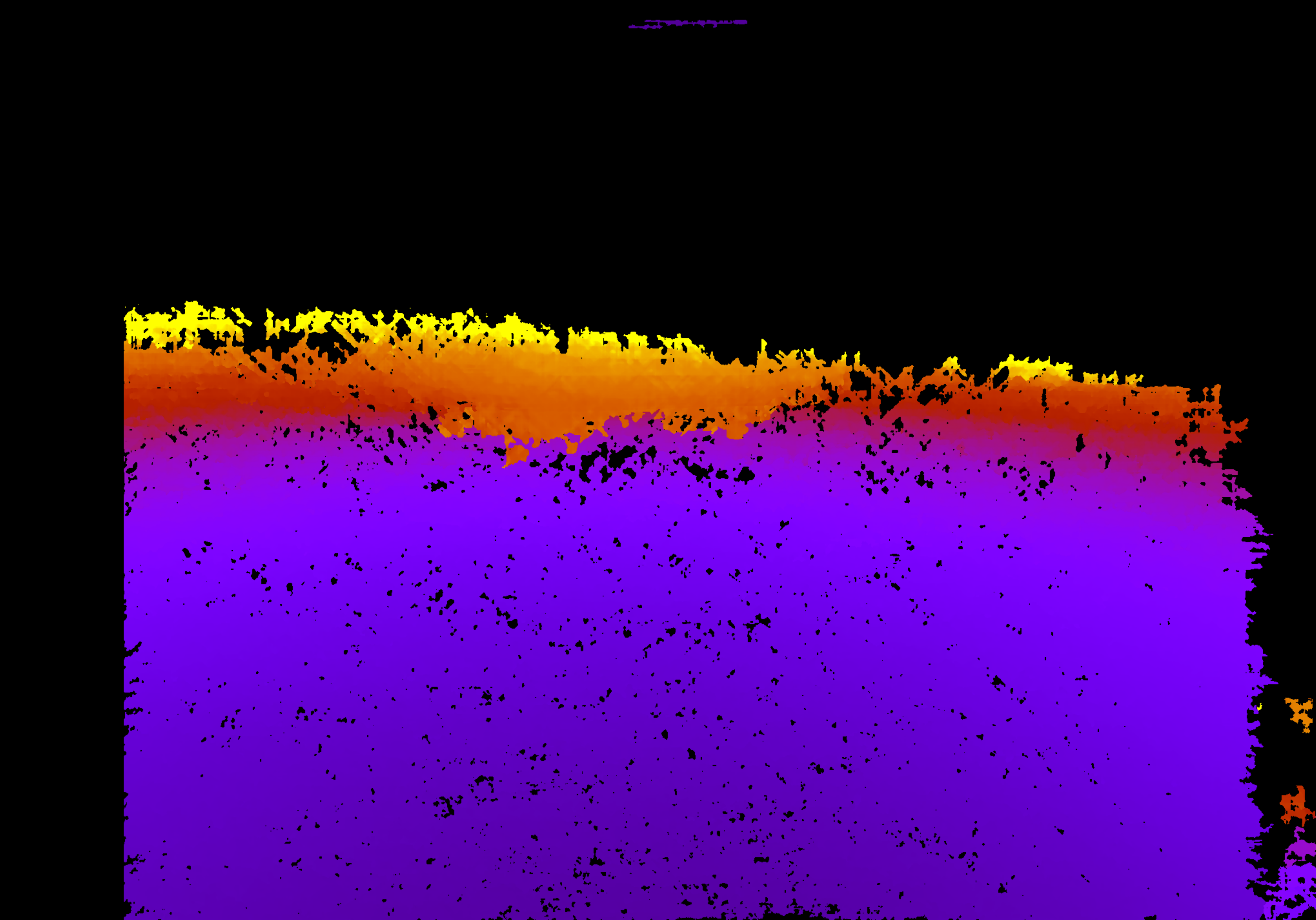}}
    \subfloat[]{\includegraphics[width=0.24\textwidth]{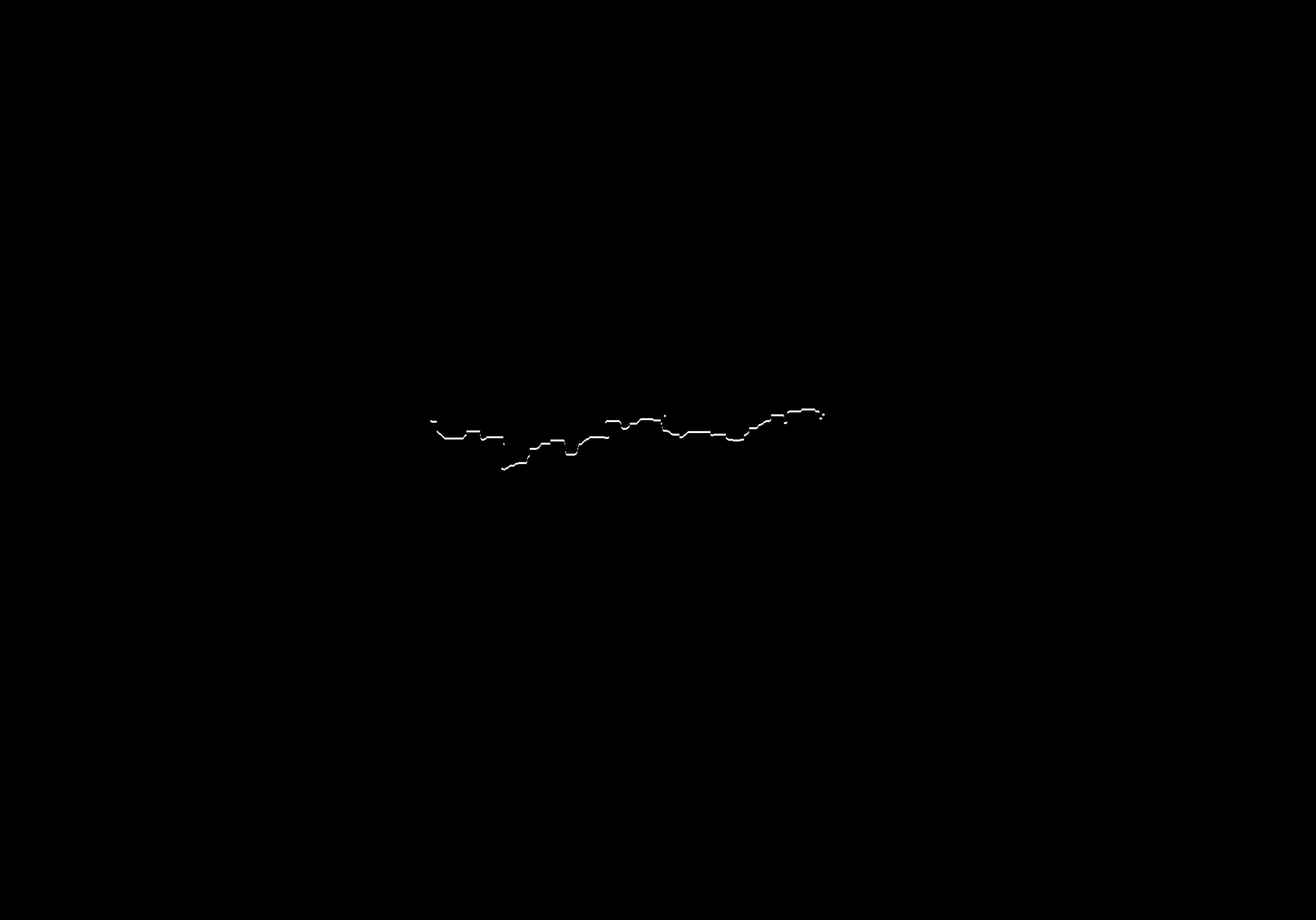}}

    \subfloat[]{\includegraphics[width=0.24\textwidth]{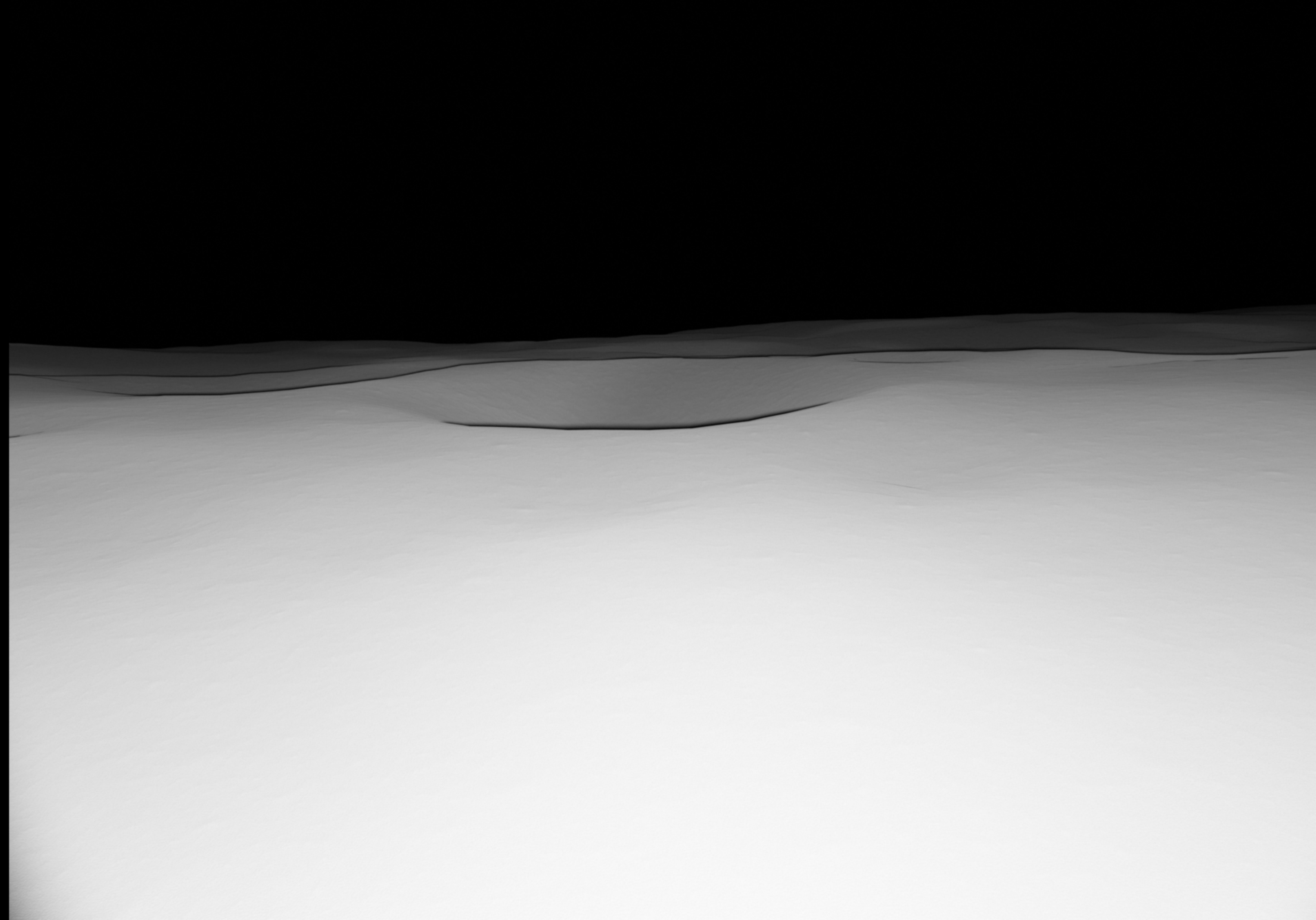}}
    \subfloat[]{\includegraphics[width=0.24\textwidth]{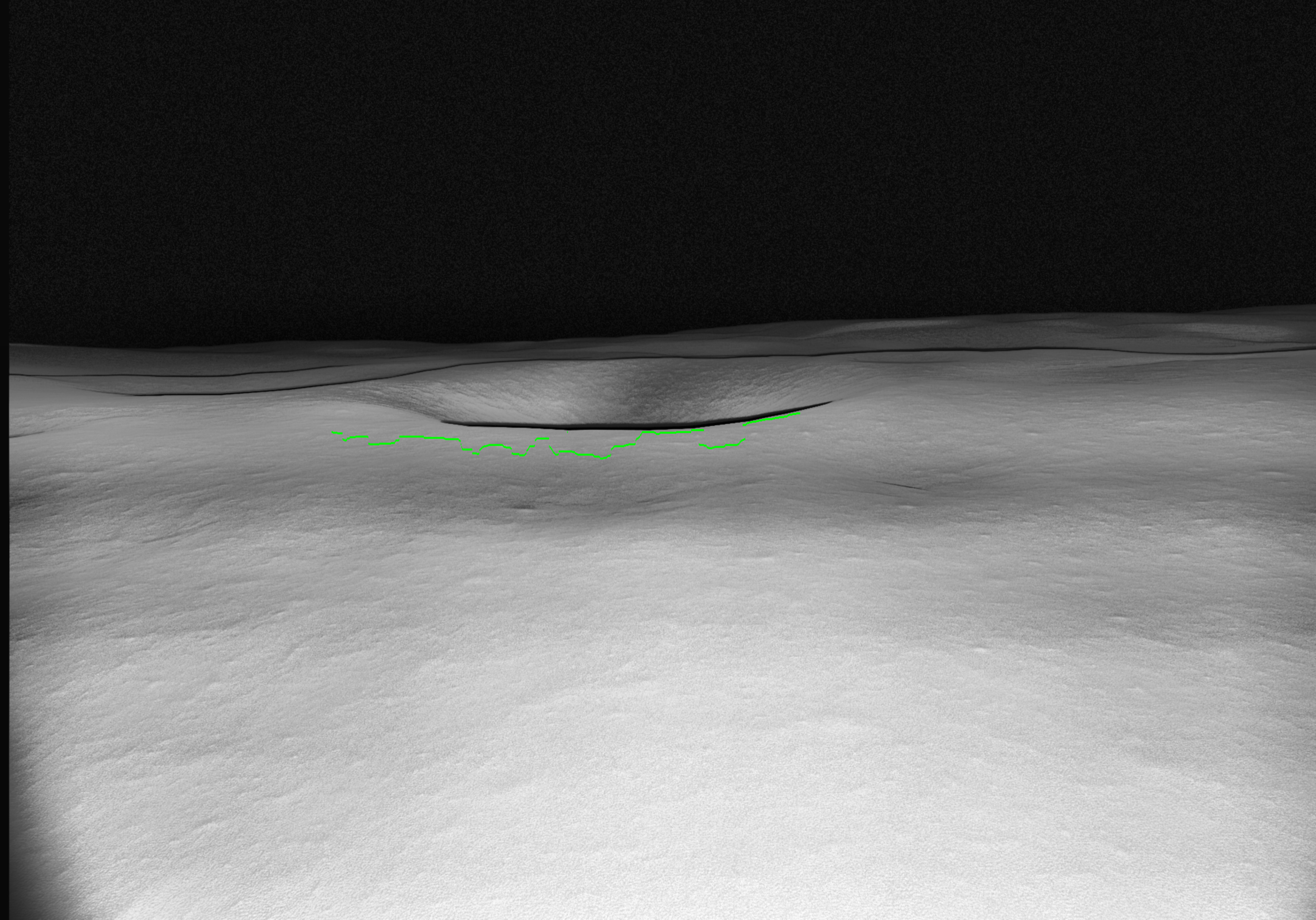}}
    \subfloat[]{\includegraphics[width=0.24\textwidth]{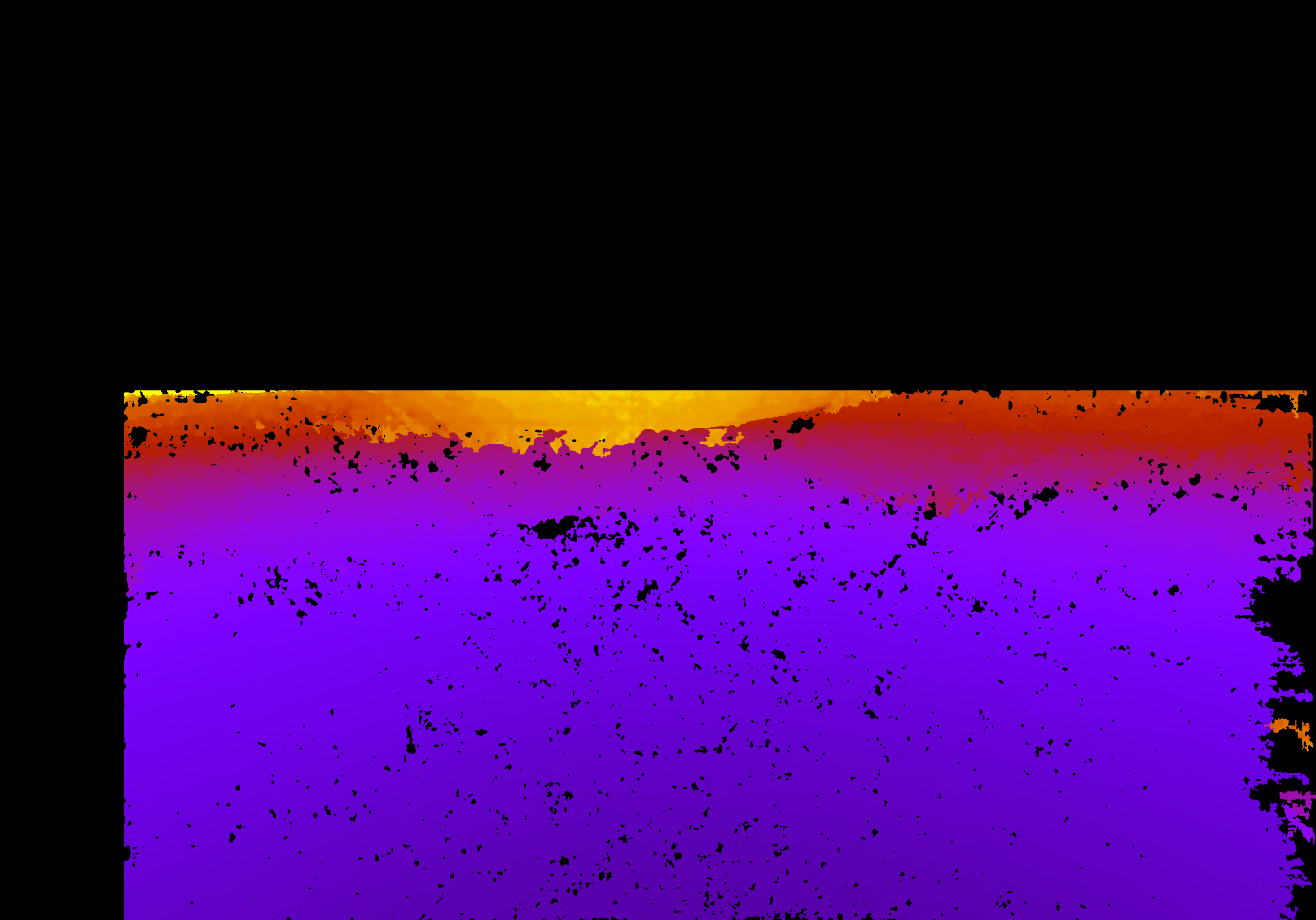}}
    \subfloat[]{\includegraphics[width=0.24\textwidth]{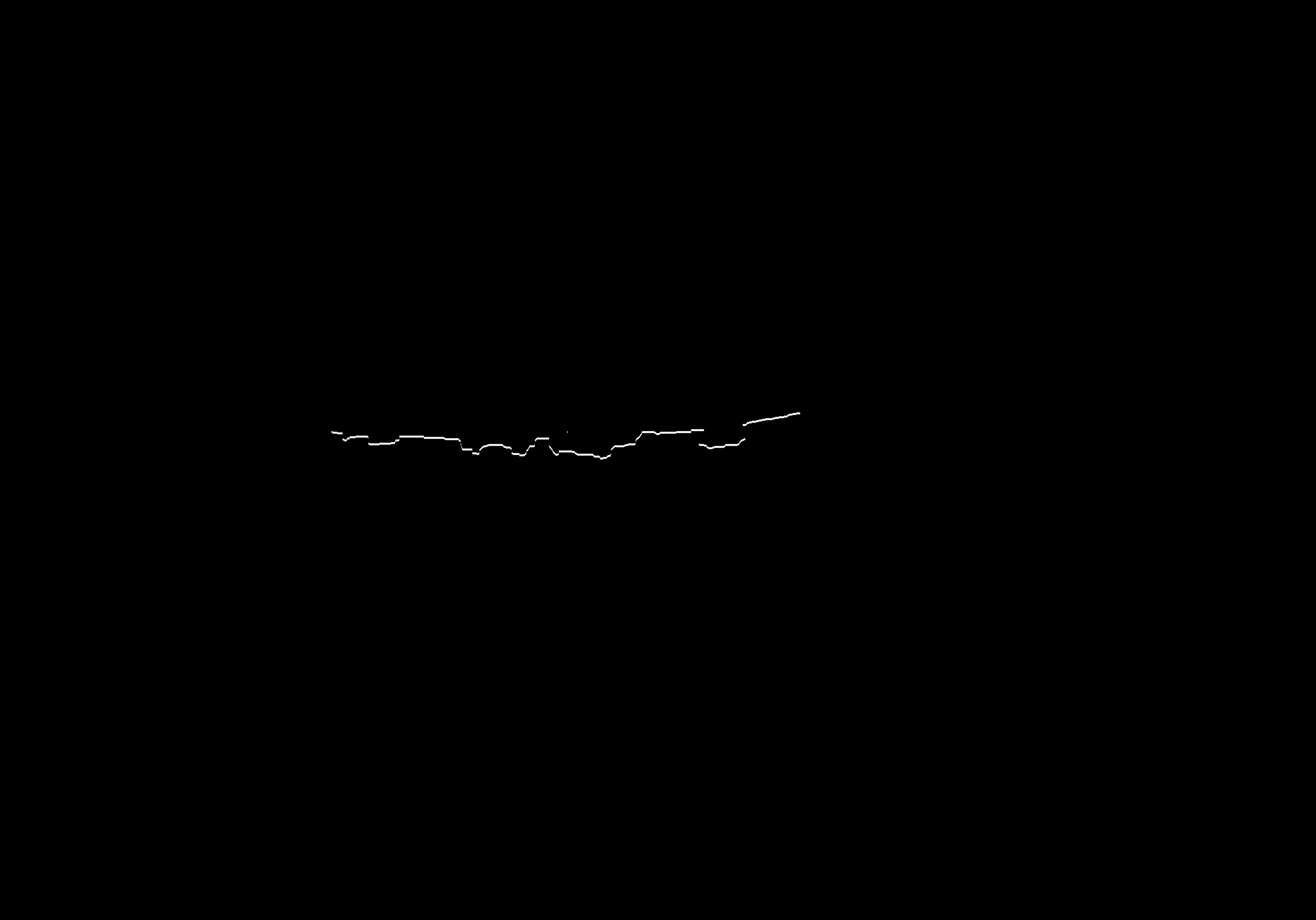}}

    \subfloat[]{\includegraphics[width=0.24\textwidth]{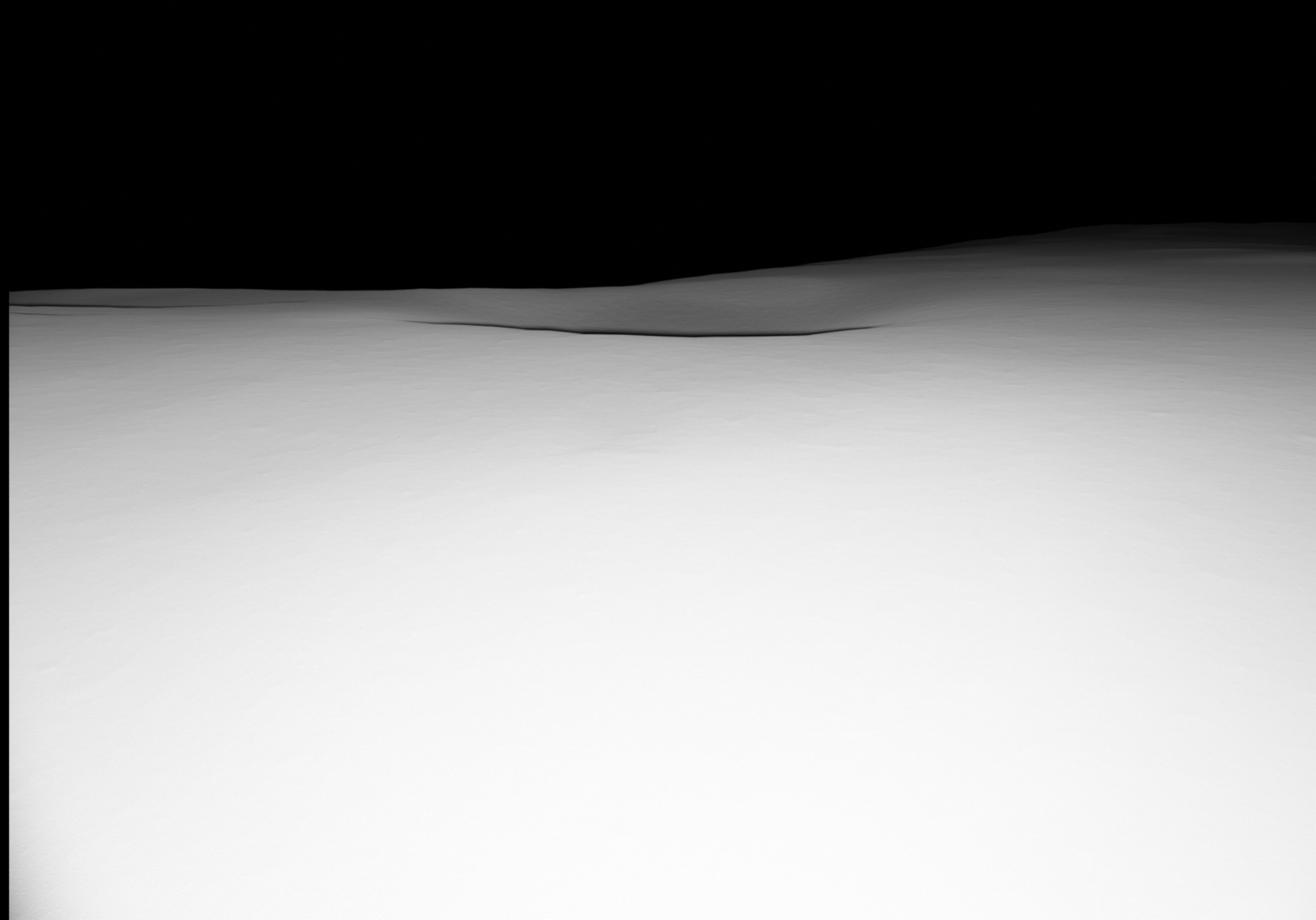}}
    \subfloat[]{\includegraphics[width=0.24\textwidth]{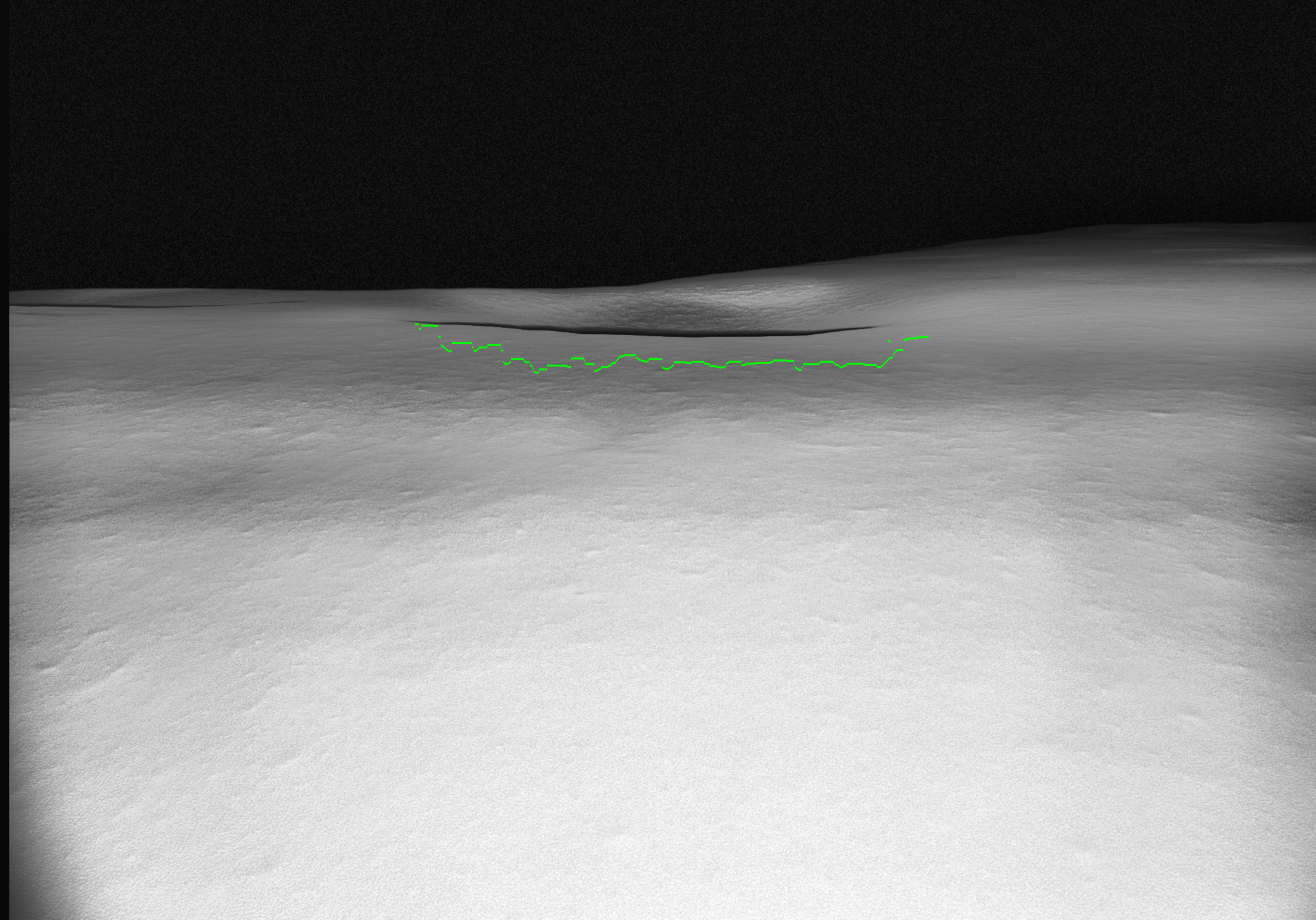}}
    \subfloat[]{\includegraphics[width=0.24\textwidth]{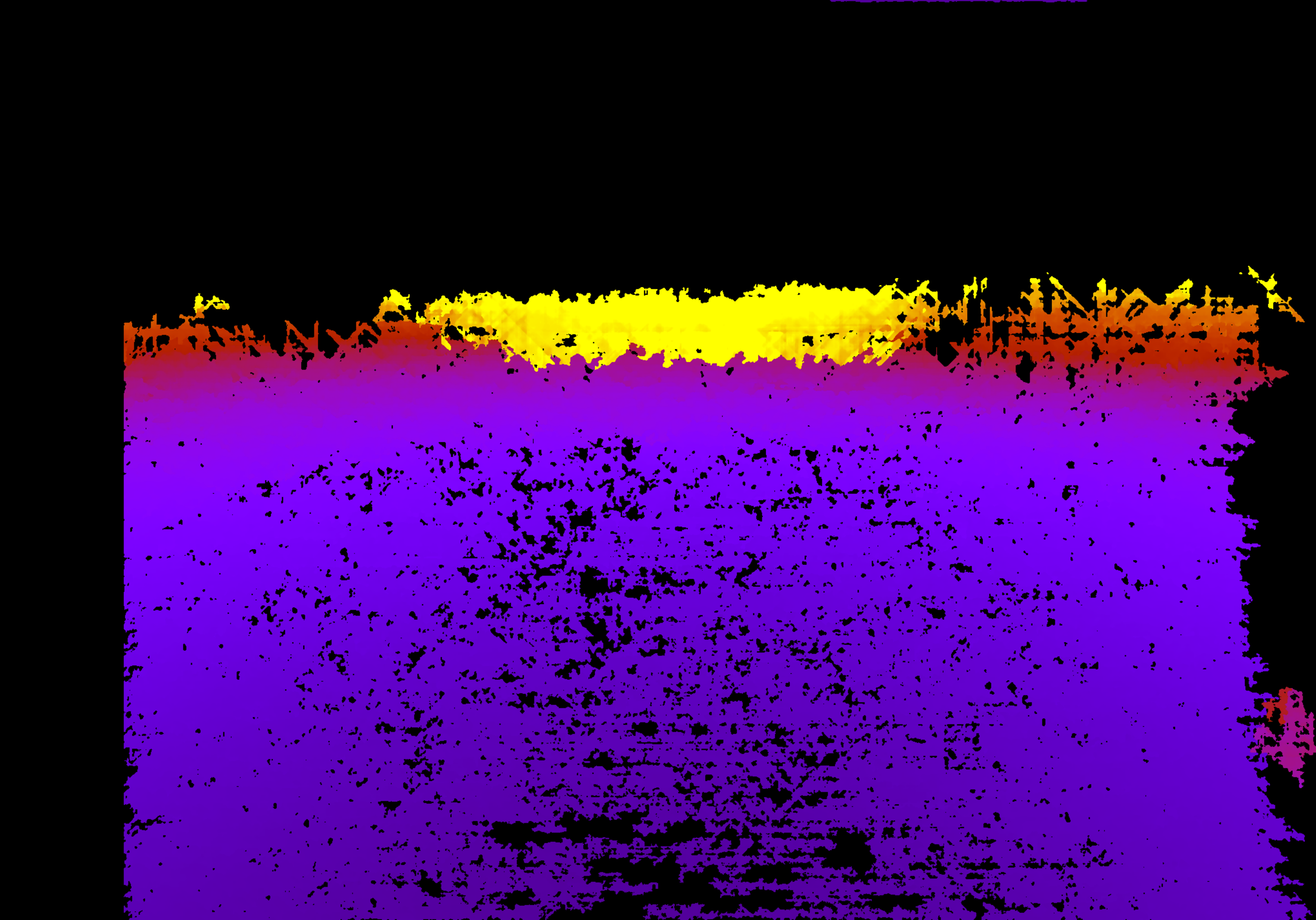}}
    \subfloat[]{\includegraphics[width=0.24\textwidth]{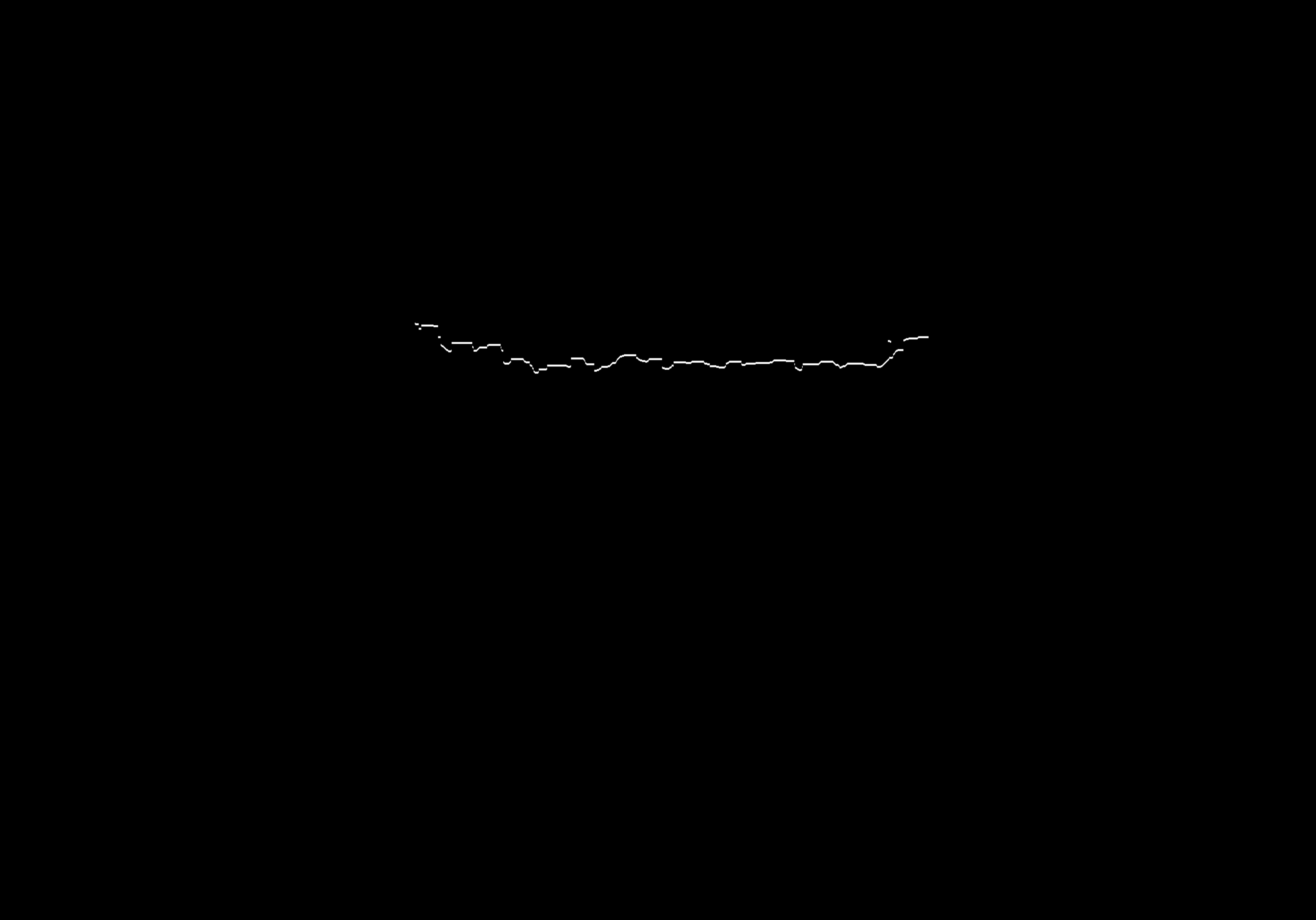}}
    
    \caption{Samples of three main perception components. In the left column is the raw image. The left-middle column is the enhanced image with a green line to mark rim detections. The right-middle column is a colorized stereo range image. The right column is the crater rim detection mask. The top three samples are from Cinder Lakes and the bottom three samples are from the Lunar simulator. \revised{In samples that include noisy stereo at range (o \& w), washed out near-range pixels (e), cluttered rims (b), and varying sizes and surface textures the approach is able to find the crater rim.}}
    \label{fig:percept-samples}
\end{figure*}

To further evaluate crater detection, simulated data is evaluated first by comparing crater rim detection and matching with and without Gaussian noise added.
Figure~\ref{fig:sim-noise-eval} contains the results of comparing noise free versus images with zero mean, 1.2 sigma pixel intensity Gaussian noise added.
From these results, it is observed that the addition of noise has minimal impact on the performance of crater detection.
Compared with the Cinder Lakes results, however, detection performance was a bit worse at both close to \SI{5}{\meter} to the crater and beyond \SI{15}{\meter} from crater rim though overall similar.
The larger difference in performance is observed in the Q-Score results where there is a more significant decrease in performance beyond \SI{15}{\meter}. 
However, for both real and simulated craters with diameters between \SI{7.5}-\SI{20}{\meter} Q-Score results are near or above 0.4 within \SI{15}{\meter}.
Potential sources of the performance difference between real and simulated data include the simulation containing images with a wider field-of-view, a \SI{2.5}{\meter} camera height compared to \SI{1.5}{\meter} camera height for Cinder Lakes data, and the inherent simulation versus real differences in lighting, camera exposure handling, and surface textures. 
Using the higher camera height within simulation could reduce the discontinuity distance and explain the performance difference within \SI{7}{\meter} to the front of the crater rim. 
The larger field-of-view camera within simulation would allow for more potential false positives to be captured as evidenced by strong crater rim percent detection but degraded Q-Scores at \SI{15}{\meter}-\SI{20}{\meter} ranges to crater rims.

Figure~\ref{fig:percept-samples} shows qualitative samples of both Cinder Lakes and simulated data images, stereo, and detections at \SI{12}{\meter} from crater rims.
We see from these samples that the simulated images have noisier stereo.
Further, there are holes in the stereo images and especially at ranges greater than~\SI{15}{\meter}.
Such stereo holes could lead to false positives and 
Such stereo holes could lead to false positives and noisy stereo could potentially yield noisier edge detections.
Finally, both Cinder Lakes and simulated imagery results in high-quality rim detections.

\begin{figure}[t]
    \centering
    \subfloat[]{\includegraphics[width=0.48\columnwidth]{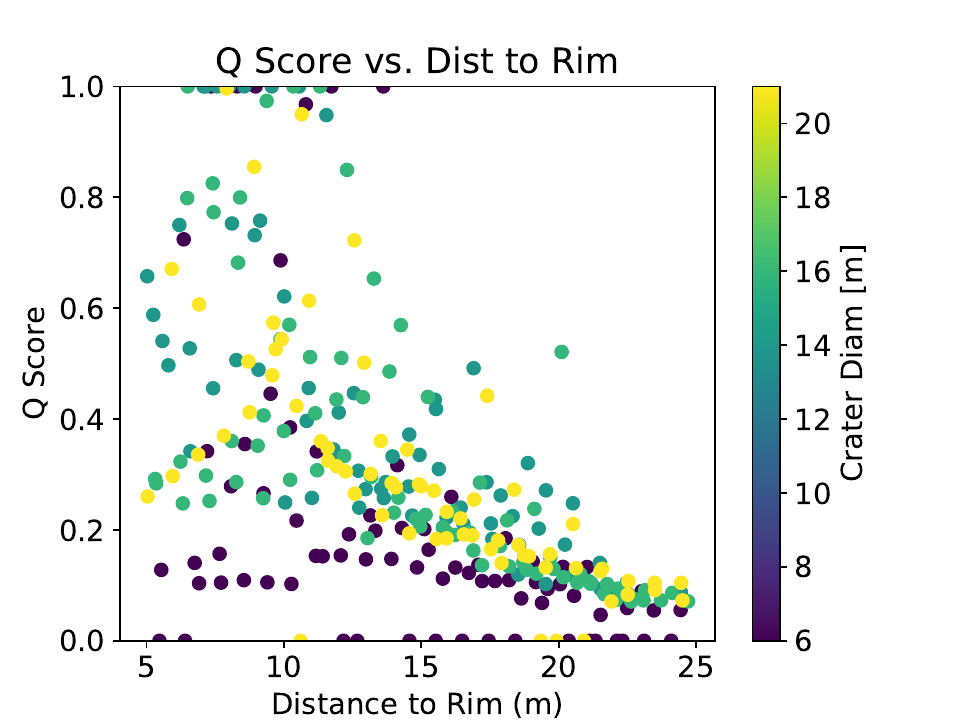}
    \label{fig:sim-off-q-5}}
    \subfloat[]{\includegraphics[width=0.48\columnwidth]{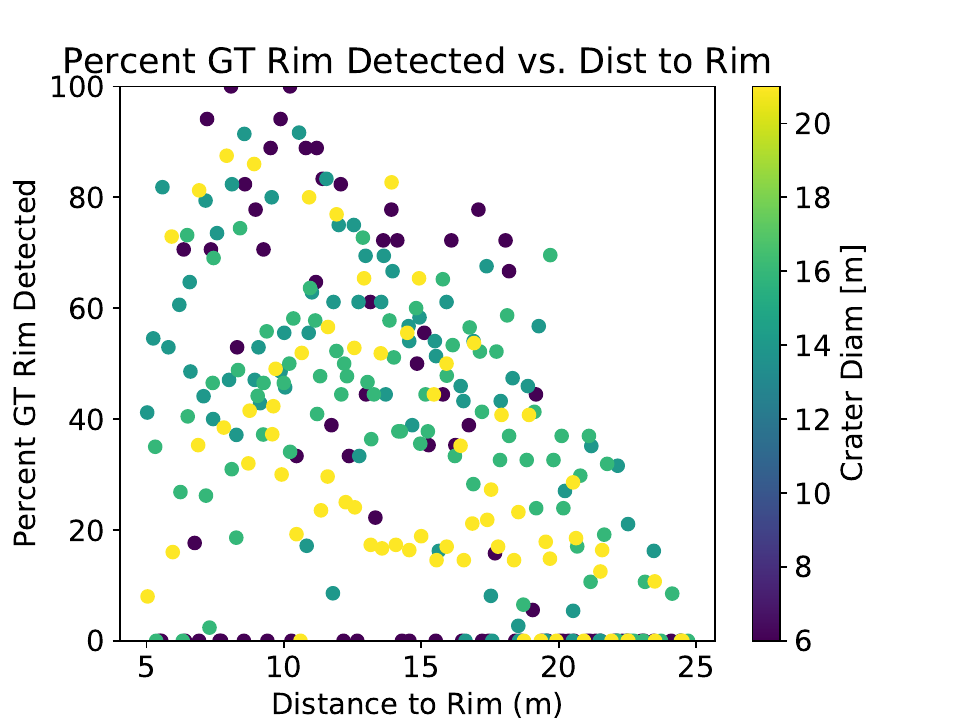}
    \label{fig:sim-off-q-10}} \\
    \subfloat[]{\includegraphics[width=0.48\columnwidth]{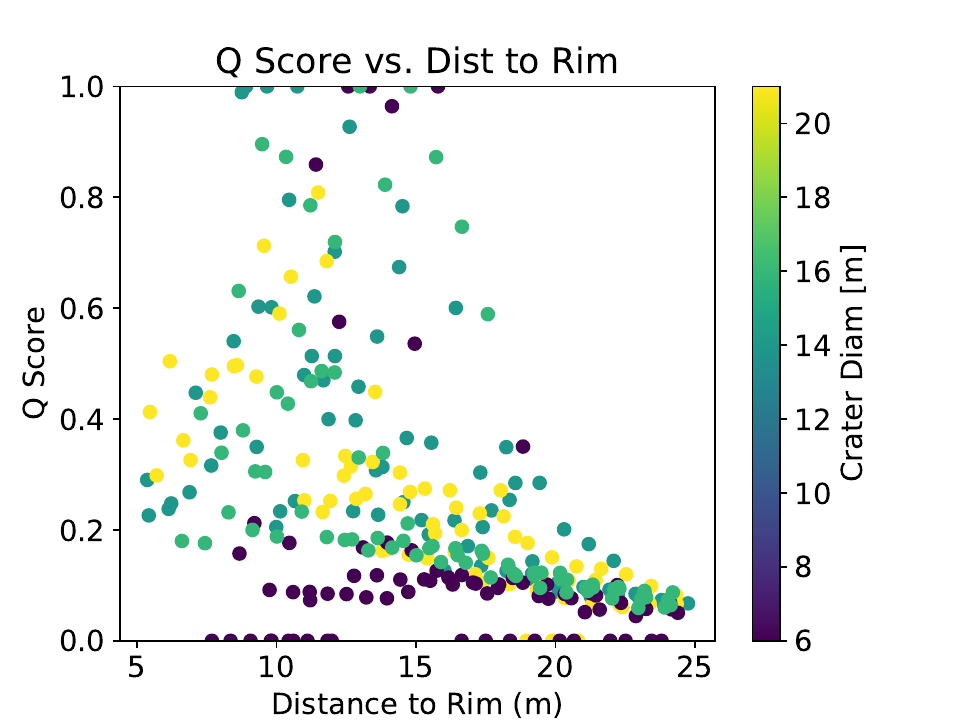}
    \label{fig:sim-off-percent-5}}
    \subfloat[]{\includegraphics[width=0.48\columnwidth]{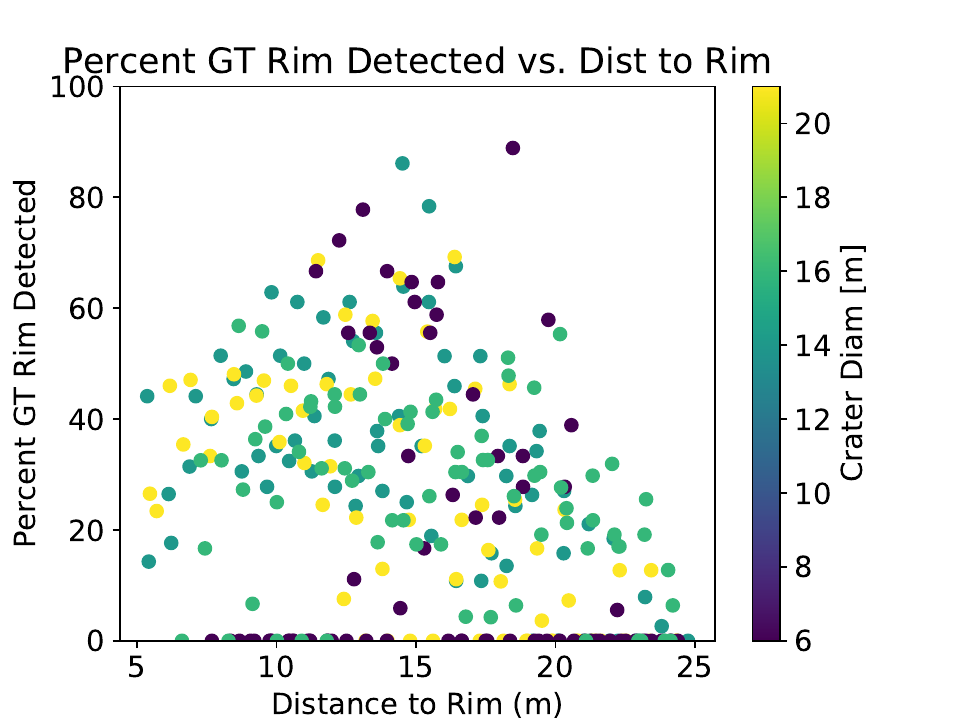}
    \label{fig:sim-off-percent-10}}
    \caption{Crater detection and matching results with positional error on simulated data with a camera height of 2.5m a light offset of 20cm below camera, high brightness, and 1.2 pixel intensity sigma Gaussian noise to image (a) Q-Score 5m error (b) Percent detected 5m error (c) Q-Score 10m error (d) Percent detected 10m error. \revised{Crater performance with \SI{5}{\meter} offset is similar to that demonstrated previously in Figure \ref{fig:cl-q-percent} and \ref{fig:sim-noise-eval}. With \SI{10}{\meter} offset though, the approach begins to degrade in the near-range. Fortunately, performance is maintained at ranges greater than \SI{10}{\meter} from the crater rim.}}
    \label{fig:sim-offcenter-q-percent}
\end{figure}

One of the limitations of the above analysis is that all of the craters were mostly centered within the image frame.
This centering will only be possible with both an accurate orientation and position knowledge,
However, upon approaching a crater, there is an expected error in absolute position due to the drift of relative localization.
To evaluate how these positional errors affect detection performance, data was rendered with a positional offset from the position from which the ideal heading was computed.
This data is described in more detail in Section \ref{sec:sim-data}.
Using this data, Figure \ref{fig:sim-offcenter-q-percent} shows the results of computing the Q-Scores and percents of crater rims detected with a \SI{5}{\meter} and \SI{10}{\meter} horizontal shift. 
Only horizontal errors are plotted because forward/backward errors are captured as a shift along the x-axis in Figure \ref{fig:sim-noise-eval}.
From Figure \ref{fig:sim-offcenter-q-percent}, we observe that a \SI{5}{\meter} horizontal offset contains some reduction in crater matching performance but percent of crater rim detected maintains similar performance without a position offset.
However, the performance of crater matching and detection degrades at less than~\SI{10}{\meter} range due to the crater not being captured within the field-of-view of the camera.
Crater matching performance still degrades around \SI{15}{\meter}, but the degradation is similar to the results without a positional offset.
Overall crater detection and matching demonstrates strong performance within the \SI{5}{\meter}-\SI{15}{\meter} range with potential for performance out to \SI{20}{\meter} detection range.


\revised{Through these experiments, reliable crater rim detection can be performed across a varying set of hardware specifications and positional uncertainty.
Using a camera \SI{1.5}{\meter} above the ground camera versus a camera \SI{2.5}{\meter} above ground has some impacts at the extremes of the range to craters \SI{5}{\meter} and \SI{20}{\meter}, but are both feasible especially while focusing on ranges to crater rims within the extremes. 
Within simulation, stereo and crater rim detection were robust to a couple of different lighting conditions and this is supported by additional experimentation on two different real locations, each containing different surface reflectance properties and therefore different pixel exposure values.
Existing stereo baselines used by Mars 2020 and proposed by the Endurance concept are shown to be feasible.
This baseline is not shown to be a requirement though.
Multiple camera field-of-views are used between simulation and real. While there is some risk for increased false positives with a wider field-of-view, it was not shown to be a limiting factor.
We determine a critical specification to place lighting source offset below a camera to minimize the impacts of the opposition effect.
}

\begin{table*}[!t]
\caption{Localization results for all simulated trajectories with 2\% added error using SGBM and discontinuity detection for perception. \revised{Localization accuracy is optimal when a half survey around the landmark crater occurs. A full survey can be performed but is likely unnecessary. Driving with a straight trajectory past the crater has significantly higher risk of failing to localize with \SI{5}{\meter} than a half or full survey.}}
\label{tab:blender-loc-results}
\centering
\begin{adjustbox}{width=1\textwidth}
\begin{tabular}{|c|c|c|c|c|c|c|}
\hline
Trajectory & Distance (m) & Num Landmarks & Avg error (m) & Stdev error (m) & Max error (m) & $>$ 5m error \\
\hline
straight crater15 & 558 & 2 & 6.36 & 2.16 & 11.63 & 0.77 \\
half survey crater15 & 617 & 2 & 3.75 & 1.28 & 5.40 & 0.13 \\
full survey crater15 & 759 & 2 & 5.18 & 0.84 & 6.78 & 0.67 \\
straight NtoS & 665 & 2 & 4.81 & 2.56 & 9.01 & 0.50 \\
half survey NtoS & 714 & 2 & 2.12 & 1.16 & 5.95 & 0.07 \\
full survey NtoS & 840 & 2 & 2.08 & 0.97 & 4.22 & 0.00 \\
straight StoN & 536 & 2 & 7.65 & 2.63 & 11.29 & 0.70 \\
half survey StoN & 591 & 2 & 4.16 & 2.06 & 9.07 & 0.30 \\
full survey StoN & 726 & 2 & 3.10 & 2.58 & 10.39 & 0.13 \\
traj 1km & 1058 & 3 & 2.84 & 1.28 & 5.59 & 0.10 \\
traj 1km rev & 1211 & 3 & 4.34 & 2.02 & 8.84 & 0.33 \\
\hline
\end{tabular}
\end{adjustbox}
\end{table*}

\subsection{Absolute Localization}

Instantaneous perception measurements demonstrated strong performance in the previous section, but also presented some issues.
Through the use of a particle filter approach outlined in Section \ref{sec:particle-filter}, multiple perception measurements can be fused overtime to correct absolute position after it has drifted. 
Both simulation and Cinder Lakes data is used to evaluate the performance of the absolute \revised{2D} localization.
\revised{For these experiments a star tracker is not available and we use an analogue means to find orientation.
For simulation, the ground truth orientation is used.
For the data collected at Cinder Lakes, the orientation is obtained from the \ac{ins} system mounted to the camera.}

\subsubsection{Simulation}

\begin{figure}[t]
    \centering
    \subfloat[]{\includegraphics[width=0.31\columnwidth]{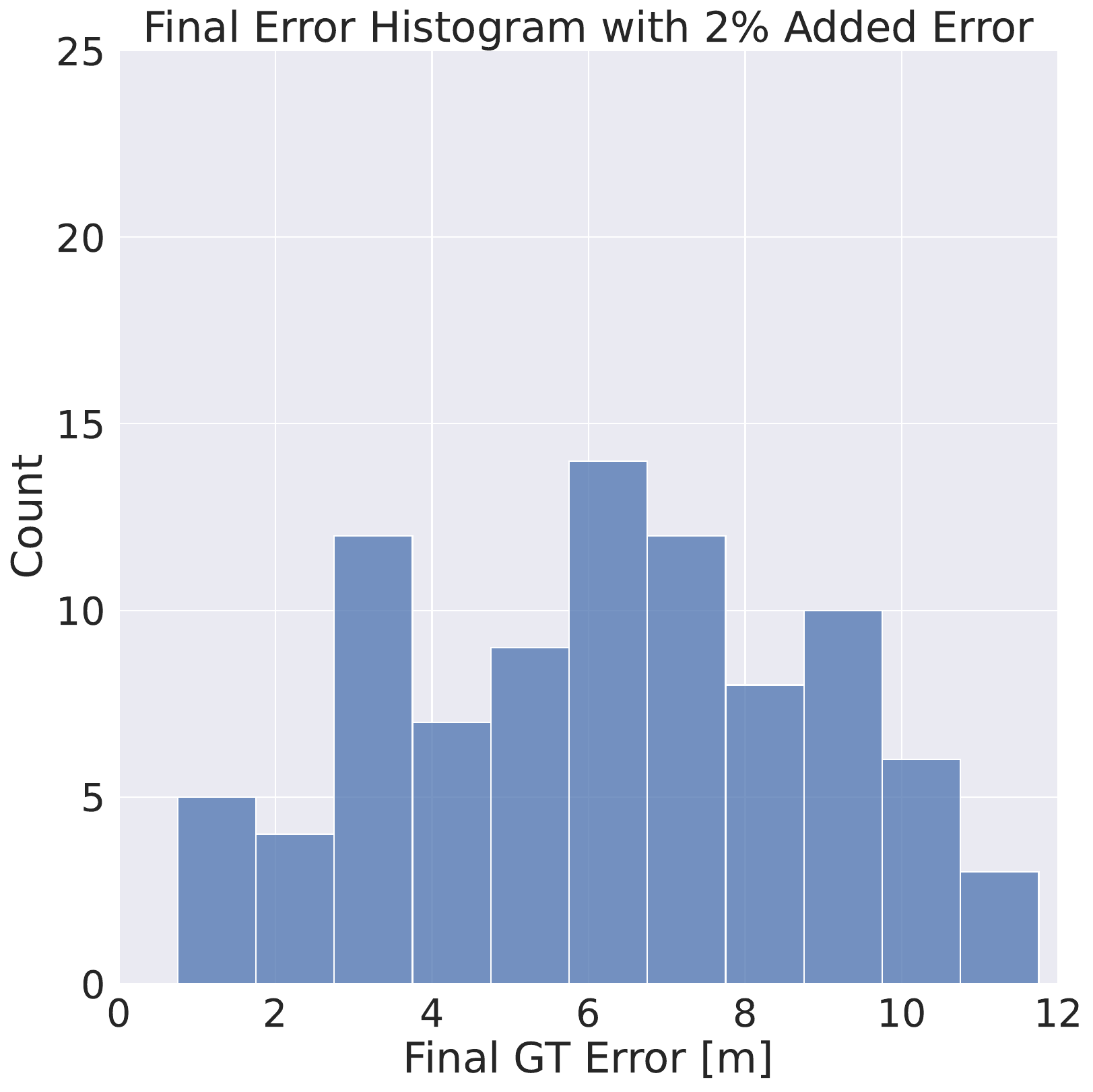}
    \label{fig:sim-hist-straight}}
    \hfil
    \subfloat[]{\includegraphics[width=0.31\columnwidth]{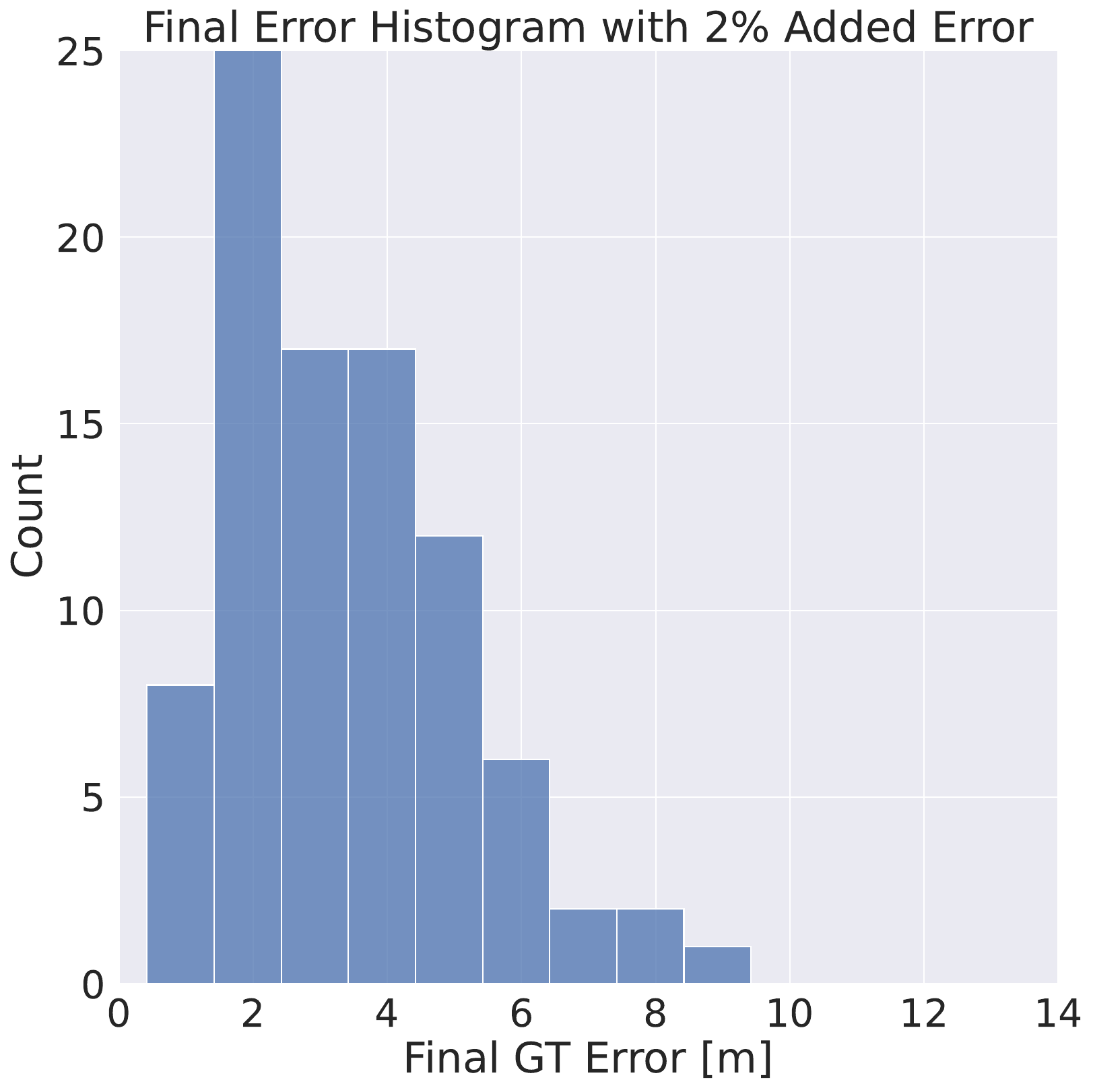}
    \label{fig:sim-hist-half}}
    \hfil
    \subfloat[]{\includegraphics[width=0.31\columnwidth]{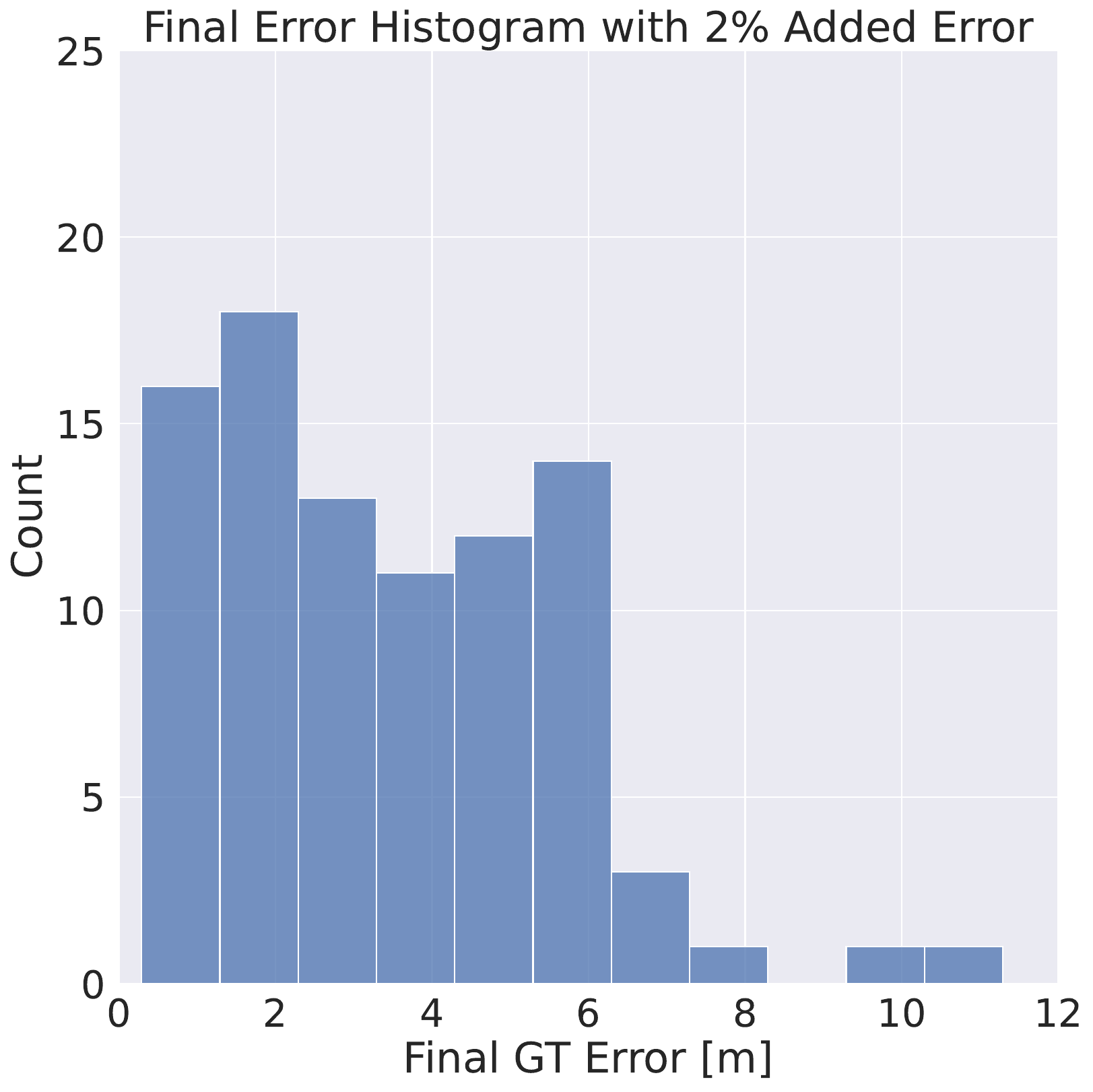}
    \label{fig:sim-hist-full}}
    \hfil
    \caption{Simulation absolute localization histogram results with 2\% error added for different trajectories near landmarks. (a) Straight trajectory (b) Survey half of crater rim (c) Survey full crater rim. \revised{Performing a half survey leads to final errors of less than \SI{5}{\meter} in most cases and is centered around \SI{2}-\SI{3}{\meter} error. Driving a straight trajectory past the crater demonstrates significant risk to fail to localize to less than \SI{5}{\meter} error.}}
    \label{fig:sim-hist-traj-type}
\end{figure}

\begin{figure}[t]
    \centering
    \subfloat[]{\includegraphics[width=0.31\columnwidth]{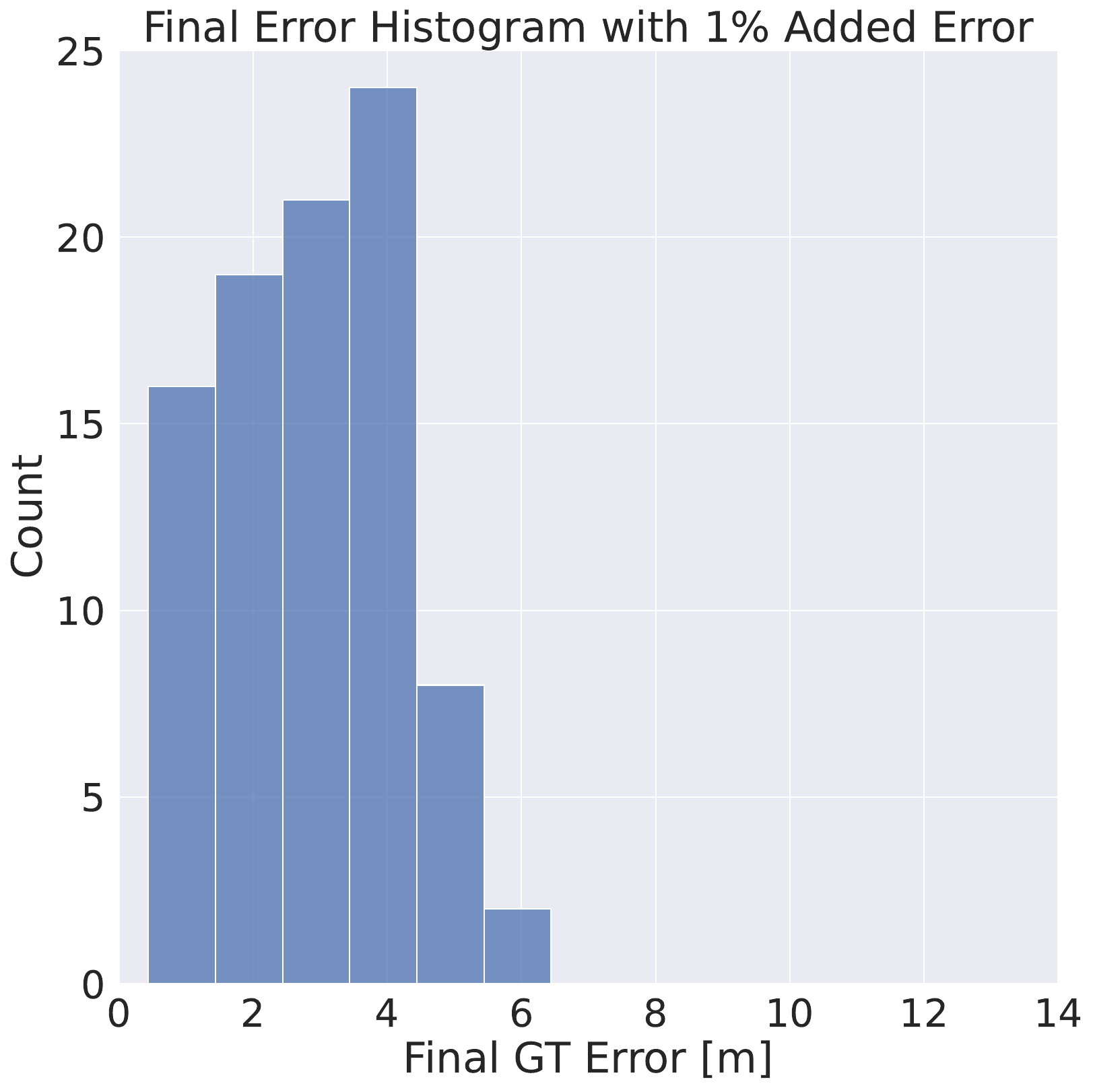}}
    \hfil
    \subfloat[]{\includegraphics[width=0.31\columnwidth]{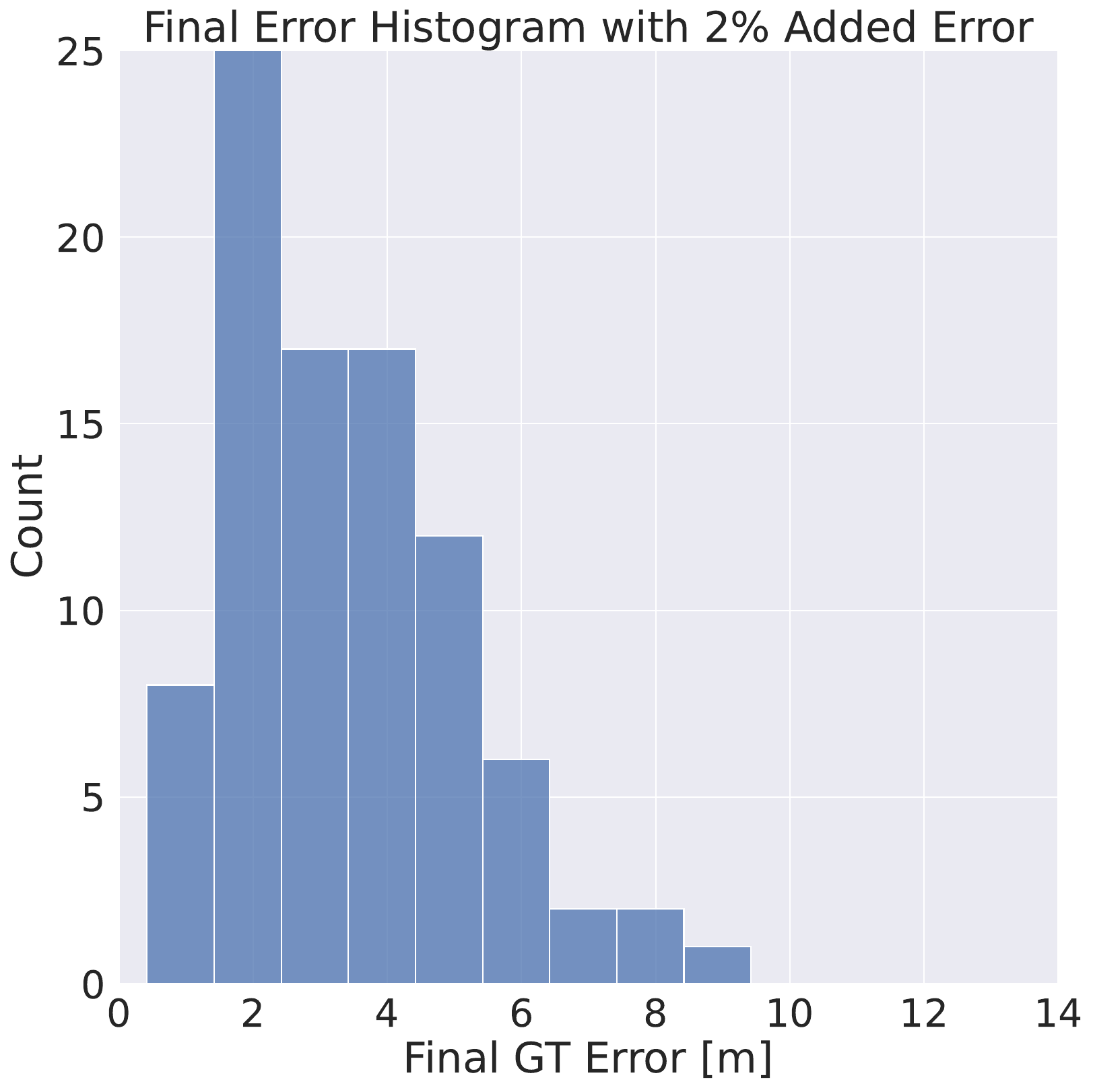}
    \label{fig:sim-hist-err2}}
    \hfil
    \subfloat[]{\includegraphics[width=0.31\columnwidth]{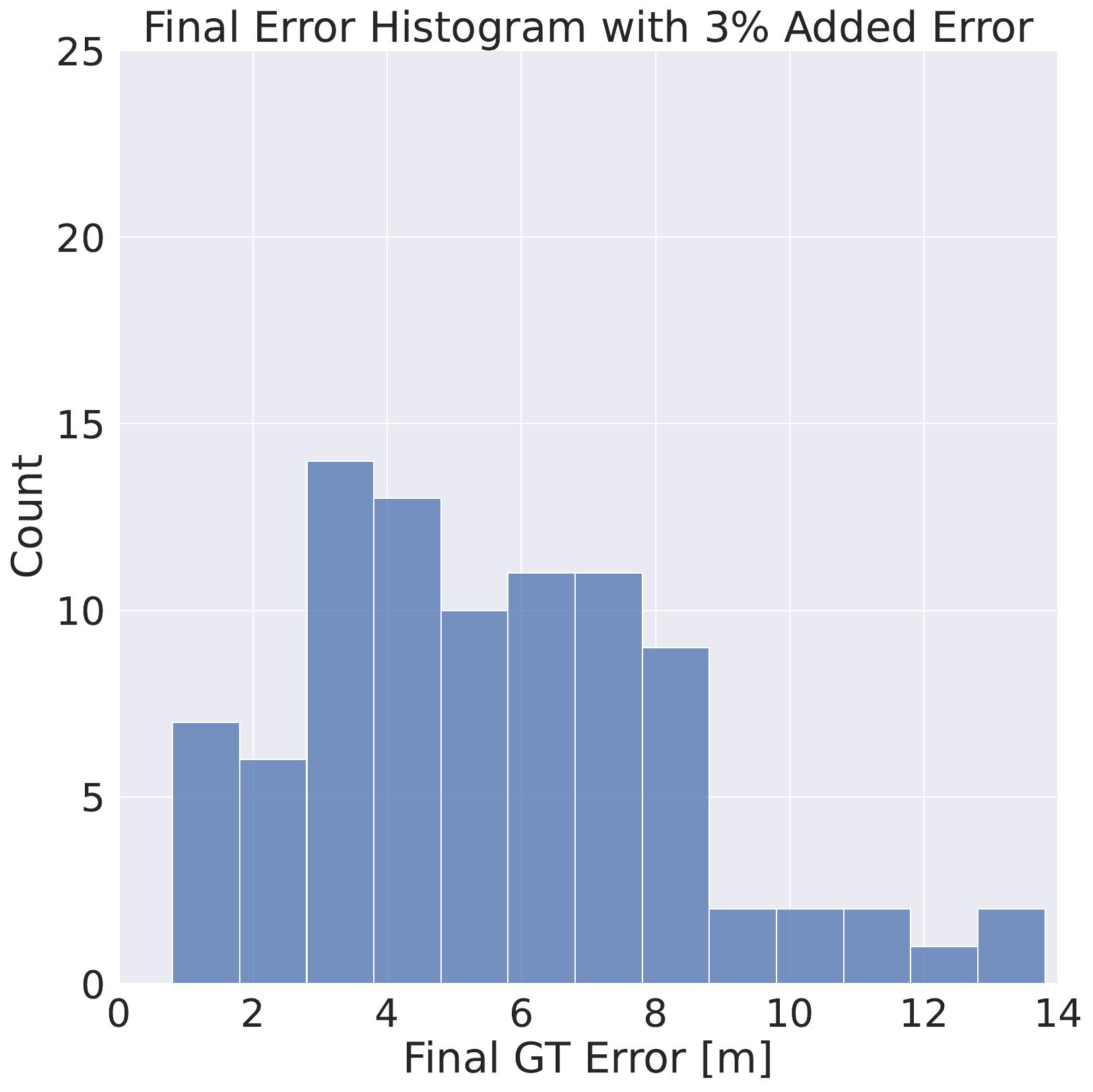}
    \label{fig:sim-hist-err3}}
    \hfil
    \caption{Simulation absolute localization histogram results for the half survey trajectories. (a) 1\% error added (b) 2\% error added (c) 3\% error added. \revised{With error rates of 2\% or less added to position along a trajectory, the approach is able to localize in nearly all samples to less than \SI{5}{\meter}. However, with more than 2\% error, the approach has risk to diverge.}}
    \label{fig:sim-hist}
\end{figure}

First, evaluation is performed using data from simulation to determine the best trajectory to observe a landmark crater.
This was accomplished by designing three types of trajectories around the same landmarks.
The first trajectory type was a straight trajectory which stops every \SI{10}{\meter} and travels directly past the crater with the nearest approach approximately \SI{10}{\meter} from the rim. 
The second trajectory type was a half survey, which was designed so that observation positions occur 180 degrees around the crater with each observation point approximately \SI{7}{\meter}-\SI{13}{\meter} from the crater rim. 
The third trajectory type was similar to half survey but had observation points a full 360 degrees around the crater. 
Table~\ref{tab:blender-loc-results} contains the results for all trajectory types with a 2\% error growth and Figure~\ref{fig:sim-hist-traj-type} is a histogram of final absolute errors of three different trajectories.
For each trajectory, we performed Monte Carlo runs of 30 different tests and each test having a 2\% relative localization error.
From these results, we observe that the half survey and the full survey approaches perform better than the straight trajectory, which is expected.
However, performing a full survey versus half survey does not provide additional benefit.
The full survey does have some trajectories that perform better but in other trajectories it actually performs worse. 
Therefore, we believe that performing a half survey provides enough positional variance to constrain the localization. 

\begin{table*}[t]
\caption{Localization results for all Cinder Lakes trajectories with 2\% added error using SGBM and discontinuity detection for perception. Error metrics are the absolute error at the final position of the trajectory. \revised{In all but one trajectory the rate of the final positional error averages less than \SI{5}{\meter} and a max error less than \SI{9}{\meter}. The longest trajectory does diverge in some cases which degrades the average error performance. However, the approach still localizes within \SI{5}{\meter} in 60\% of the samples of this trajectory.}}
\label{tab:cl-loc-results}
\centering
\begin{adjustbox}{width=1\textwidth}
\begin{tabular}{|c|c|c|c|c|c|c|}
\hline
Trajectory & Distance (m) & Num Landmarks & Avg error (m) & Stdev error (m) & Max error (m) & $>$ 5m error \\
\hline
south\_NtoS & 280 & 1 & 2.62 & 0.99 & 4.86 & 0.00 \\
south\_vA & 568 & 2 & 1.89 & 0.90 & 3.82 & 0.00 \\
south\_vA\_rev & 568 & 2 & 3.86 & 1.66 & 7.57 & 0.27 \\
south\_vA\_5loop & 4419 & 10 & 12.97 & 14.95 & 57.01 & 0.40 \\
north\_NA12\_NM12 & 424 & 2 & 2.71 & 1.23 & 5.41 & 0.03 \\
north\_ND1\_NA12\_NM12 & 424 & 3 & 3.02 & 1.63 & 6.43 & 0.10 \\
north\_ND1\_NA12\_NM12\_rev & 491 & 3 & 3.44 & 1.81 & 6.25 & 0.33 \\
north\_ND1\_NM12 & 424& 2 & 4.81 & 2.36 & 8.11 & 0.53 \\
north\_ND1\_NA12\_rev & 491 & 2 & 3.26 & 1.84 & 6.45 & 0.23 \\
north\_ND1\_NM12\_rev & 491 & 2 & 4.50 & 2.24 & 8.48 & 0.53 \\
\hline
\end{tabular}
\end{adjustbox}
\end{table*}

\begin{figure}[t]
    \centering
    \subfloat[]{\includegraphics[width=0.31\columnwidth]{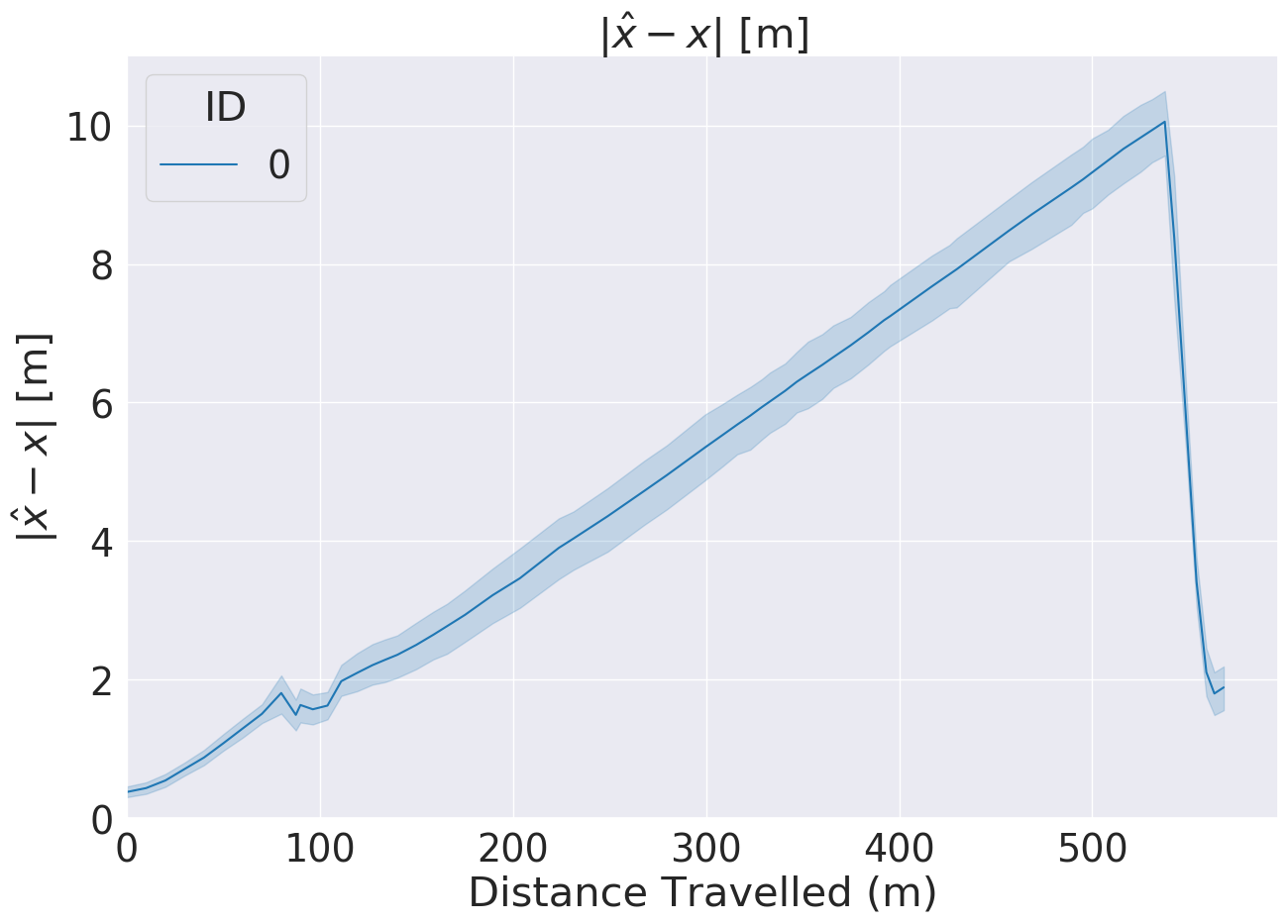}
    \label{fig:cl-err-traj}}
    \subfloat[]{\includegraphics[width=0.31\columnwidth]{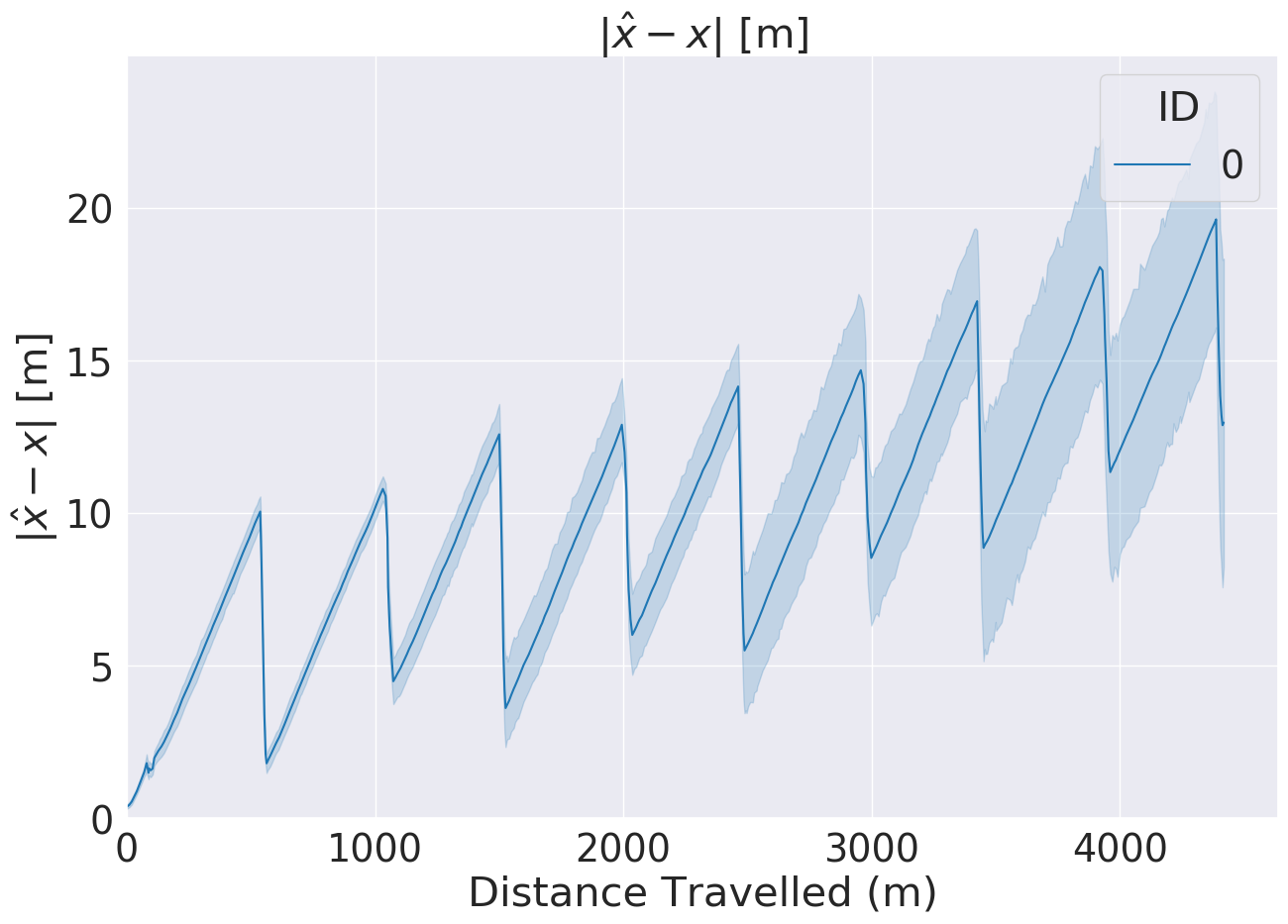}
    \label{fig:cl-err-traj-5loop}}
    \subfloat[]{\includegraphics[width=0.31\columnwidth]{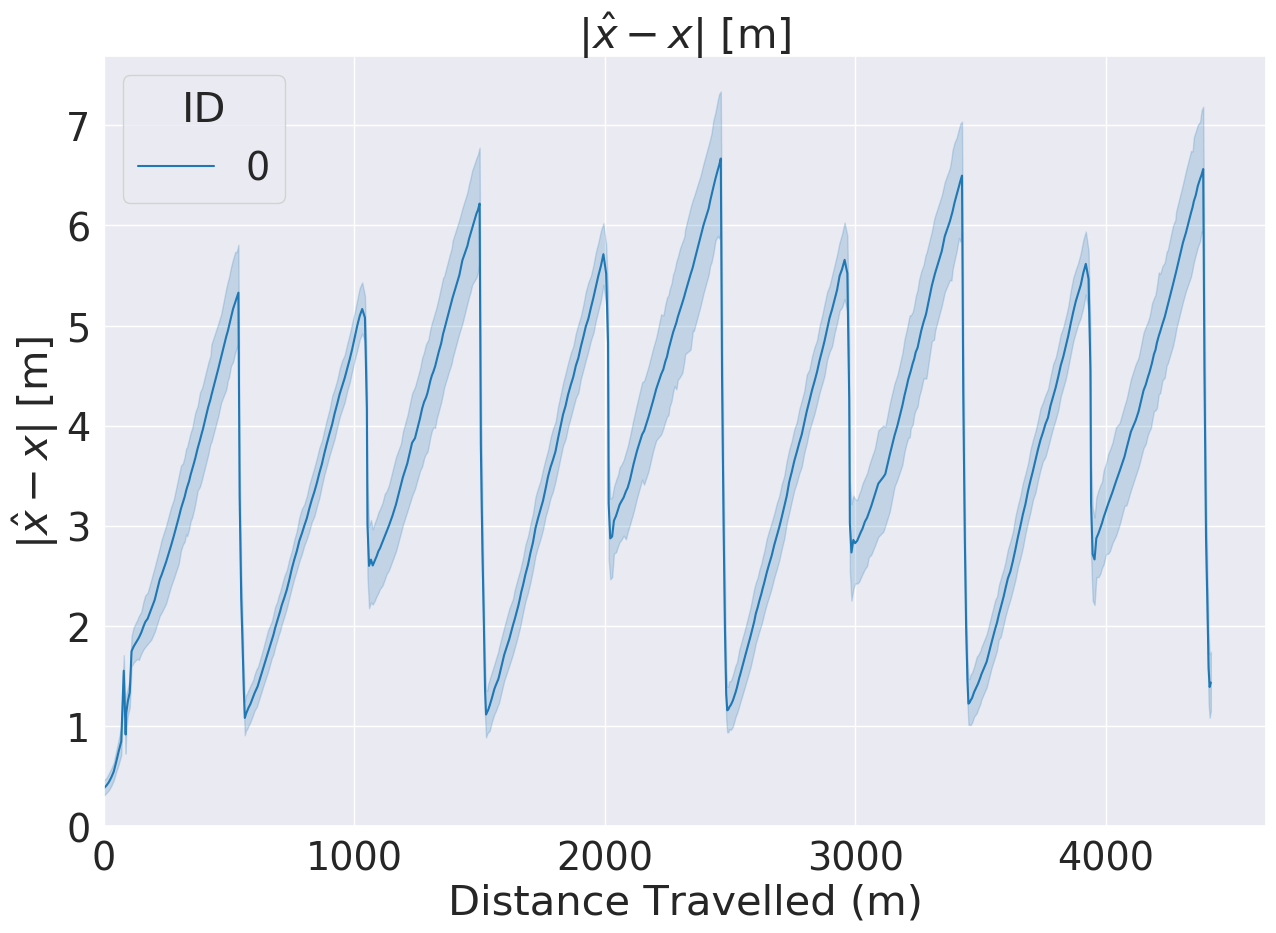}
    \label{fig:cl-err-traj-5loop-1}}
    \caption{Cinder Lakes crater absolute localization results for a single trajectory, south\_vA. (a) 1 time through trajectory with 2\% error added (b) 5 loops through trajectory with 2\% error added (c) 5 loops through trajectory with 1\% error added. \revised{(a) highlights the ability of the approach to localize to less than \SI{2}{\meter} from \SI{10}{\meter}, which is the max error target. (b) and (c) show multiple loops of the same trajectory with different error rates with the approach never diverging with 1\% error but some test cases diverging with 2\% error growth.}}
    \label{fig:cl-loop-sample}
\end{figure}

\begin{figure}[t]
    \centering
    \subfloat[]{\includegraphics[width=0.31\columnwidth]{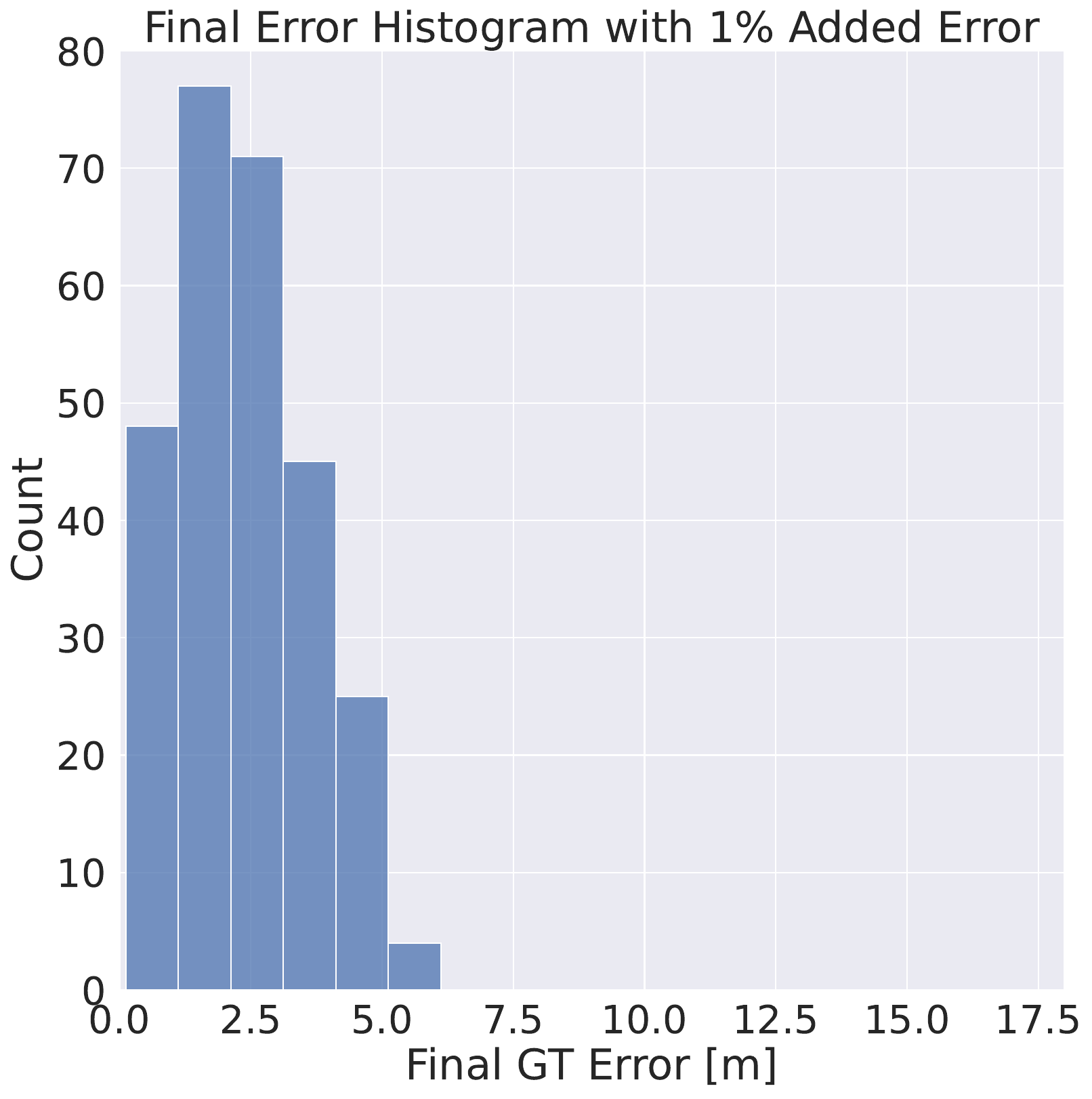}
    \label{fig:cl-hist-err1}}
    \hfil
    \subfloat[]{\includegraphics[width=0.31\columnwidth]{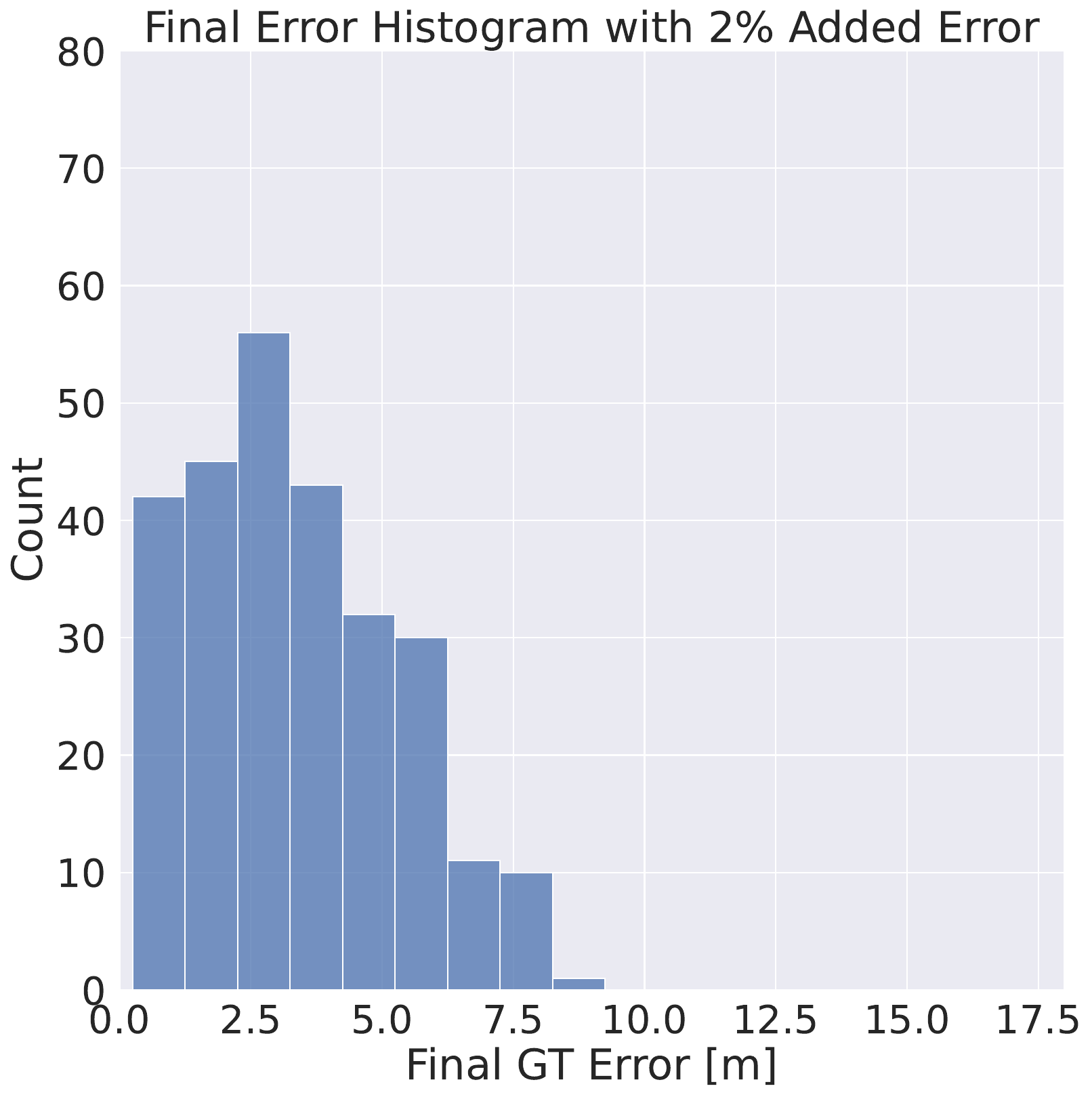}}
    \hfil
    \subfloat[]{\includegraphics[width=0.31\columnwidth]{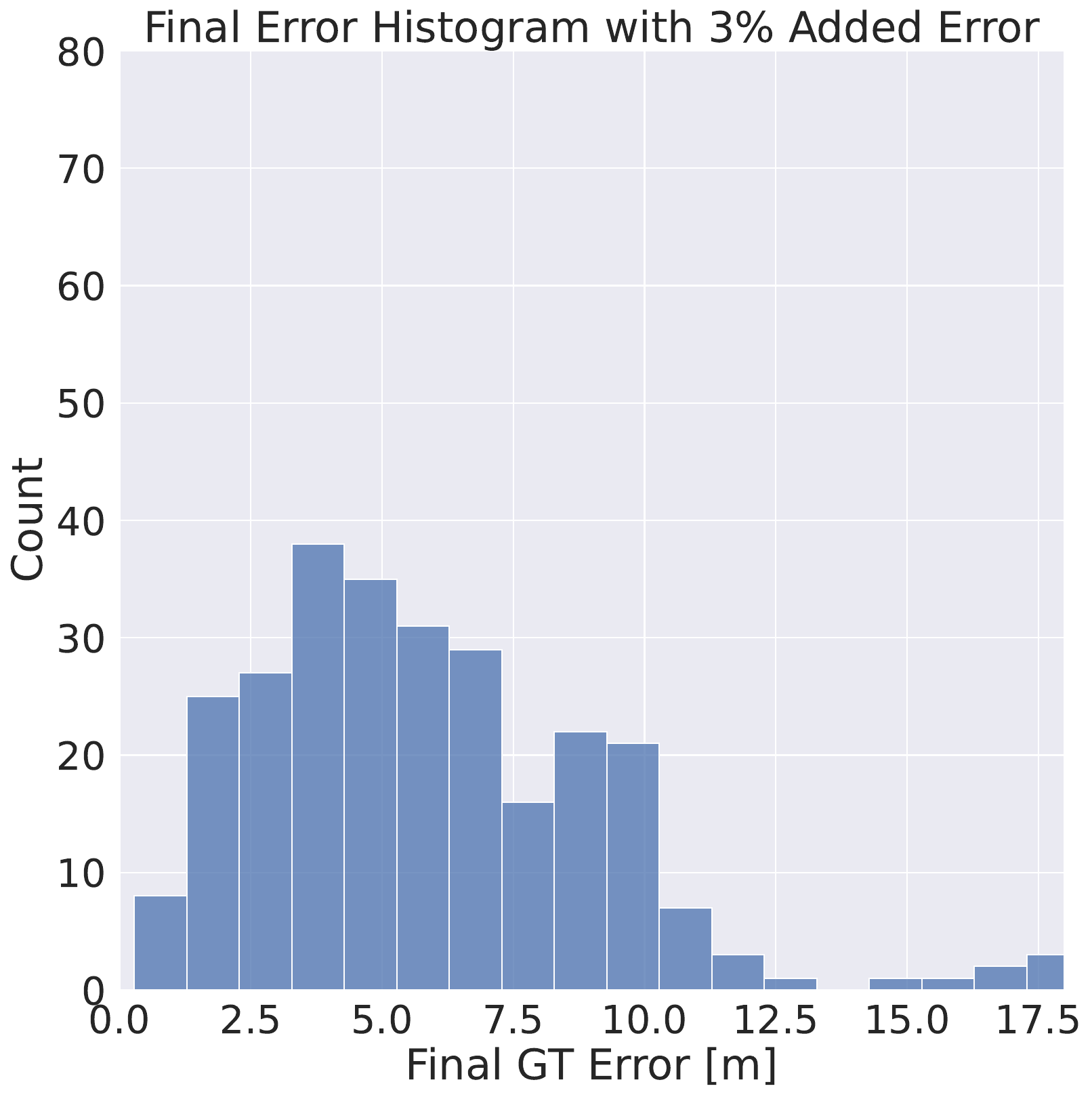}}
    \hfil
    \caption{Cinder Lakes crater absolute localization histogram results for single loop trajectories. (a) 1\% error added (b) 2\% error added (c) 3\% error added. \revised{With 2\% or less of positional error along a trajectory, the approach is able to localize within \SI{5}{\meter} most of the time and never diverges beyond \SI{10}{\meter}. With 3\% or more positional error, there is increased risk for divergence.}}
    \label{fig:cl-hist}
\end{figure}

Next, varying error rates were observed within the simulated half survey trajectories.
Figure~\ref{fig:sim-hist} shows the results of experiments for 1\%, 2\%, and 3\% added error. 
We observed that for 1\% and 2\% added error the localization is strong.
However, with 3\% added error, there are seven samples that violate the requirement of remaining below~\SI{10}{\meter} error.
This reduction in performance is expected though as the performance is designed to recover from up to \SI{10}{\meter} error.
At 3\% error over a \SI{300}{\meter} traverse, there will be about \SI{9}{\meter} of error added within our Monte Carlo runs.
The filter is only able to localize reliably within ~\SI{5}{\meter} and this potentially allows for the absolute error to increase beyond ~\SI{10}{\meter} at the second landmark.

Finally, a trajectory $\geq$\SI{1}{\km} trajectory was tested with three landmarks and involved running the trajectory forwards and backwards with half surveys of each landmark.
The results for this trajectory are also in Table \ref{tab:blender-loc-results}.
Overall, results from the \SI{1}{\km} trajectory demonstrate that the particle filter can maintain stability across more landmarks and longer trajectories.

\subsubsection{Cinder Lakes}

Next, we used the trajectories collected at Cinder Lakes to validate absolute localization in an analogue Lunar environment. 
For evaluation, we used a total of three trajectories with two from the South site and one from the North site. 
One of the south site trajectories, south\_vA was connected via artificial points to make it a loop.
There is no image data at the locations but the error grows with the distance. 
This loop utilizes the same image data repeatedly, but allows for a simulation of stability.
For the north site trajectory, different combinations of landmarks were utilized with two to three landmarks per trajectory.
The full results of these trajectories with 2\% added error are in Table \ref{tab:cl-loc-results}.
We observe that performance is under \SI{5}{\meter} average error and less than \SI{10}{\meter} max error with the exception of the looping trajectory.
Additionally, we observe that the filter experienced issues in attaining absolute errors less than \SI{5}{\meter} along the north site trajectories containing the ND1 and NM12 landmarks. 
More than 50\% of the final errors from these trajectories were greater than \SI{5}{\meter}.
Figure~\ref{fig:cl-loop-sample} presents a comparison of the actual trajectory with 2\% added error compared with the looped trajectory at 1\% and 2\% added error. 
It is shown that there is roughly \SI{400}{\meter} between craters.
With one traverse, the localization can recover from a \SI{10}{\meter} error.
However, we observed that the max error grows while running multiple loops with 2\% error and the filter diverges.
However, with multiple loops and 1\% error, the max error does not grow and the filter maintains stability.
Given these results, we can determine that error growth between landmarks should be no more than \SI{5}{\meter}-\SI{6}{\meter} error and the length of traverses between landmarks can be subsequently developed by assuming 2\% relative localization error.

Furthermore, the other non-looped trajectories were run for 1\%, 2\%, and 3\% added error rates.
The results are shown in Figure~\ref{fig:cl-hist} and, consistent with our findings in simulation, we observed filter divergence with increased relative error.
However, with 1\% and 2\% error rates, the absolute localization is able to consistently localize within \SI{5}{\meter} final error and contain no samples greater than \SI{10}{\meter} final error.

\subsection{Benchmark on Flight-Like Computer}

In this section, we demonstrate that the proposed ShadowNav approach is capable \revised{of running within the time constraints of the Endurance-A proposed \ac{conops}.}
The Endurance-A proposal ~\cite{KeaneTikooEtAl2022} calls for the rover to stop for approximately 10 minutes between drives in order to perform absolute localization.
Long exposure captures are proposed to be captured every~\SI{10}{\meter}. 
Therefore, for a~\SI{300}{\meter} traverse, there could be on the order of 30 iterations to process.
\revised{To process 30 iterations within~\SI{10}{\minute}, each iteration would need to occur in less than~\SI{20}{\second}.}
The Endurance-A concept also proposed~\SI{1}{\minute} stops every~\SI{10}{\meter} to capture long exposure images.
\revised{The last time constraint for consideration based on the proposed \ac{conops} is the time spent driving between long exposure captures in which one or more cores could potentially be allocated for perception or a localization step.
This will be at least~\SI{20}{\second} of driving based on the drive speed of the rover.
To evaluate whether the method in this paper can be run in the context of the \ac{conops}, a Snapdragon 8155 processor, which is an analogue of the computer specification (CPU cores, memory, power, etc.) for the mission, is utilized for the benchmark.
A Snapdragon 801 processor was successfully flown on the Mars Ingenuity helicopter~\cite{BalaramCanhamEtAl2018}, providing a successful technology demonstration of the Snapdragon architecture for future missions.}
First, the perception algorithms of \acp{sgbm} stereo and discontinuity detection were evaluated and the results are in Table \ref{tab:snapdragon-percept}.
\begin{table}[ht]
\caption{Perception computation results on a single core of a Snapdragon 8155}
\label{tab:snapdragon-percept}
\centering
\begin{adjustbox}{width=1\columnwidth}
\begin{tabular}{|c|c|c|c|}
\hline
Dataset & Resolution & SGBM max (sec) & Detect max (sec) \\
\hline
Cinder Lakes & 1295x602 & 0.48 & 0.25 \\
Simulation & 2039x1425 & 4.40 & 0.91 \\
\hline
\end{tabular}
\end{adjustbox}
\end{table}
\revised{For a high-resolution image, the worst case stereo time using a single CPU core is~\SI{4.4}{\second} and meets the performance constraint of being run during the long-exposure capture stop or while driving between exposure captures.}
\begin{table}[ht]
\caption{Localization step computation results on a single core of a Snapdragon 8155.}
\label{tab:snapdragon-loc}
\centering
\begin{adjustbox}{width=1\columnwidth}
\begin{tabular}{|c|c|c|c|}
\hline
Dataset & Num. Particles & PF step avg. [seconds] & Pf step max [seconds] \\
\hline
Cinder Lakes & 50 & 1.49 & 2.17 \\
Cinder Lakes & 100 & 2.96 & 4.29 \\
Cinder Lakes & 200 & 6.00 & 8.71 \\
Simulation & 50 & 1.71 & 2.78 \\
Simulation & 100 & 3.41 & 5.56 \\
Simulation & 200 & 6.96 & 11.16 \\
\hline
\end{tabular}
\end{adjustbox}
\end{table}
Table~\ref{tab:snapdragon-loc} shows the results of benchmarking these localization algorithms across a range of several different parameters. 
\revised{In the worst case using high resolution stereo and 200 particles could run in~\SI{16.47}{\second} for stereo, rim detection, and localization.
This satisfies the~\SI{20}{\second} per iteration requirement for both the time between long exposure stops and the batching all iterations for the~\SI{10}{\minute} absolute localization stop.
All of the results were computed using a single CPU core of the Snapdragon in this work.
Therefore, there exists potential to significantly increase the performance through optimizations and parallelization. The particle filter step especially lends itself to parallelization since each particle computation is independent.
Therefore, we believe that this approach fits within the computation time requirements of the mission conops and has potential to operate quicker than the requirements and is suitable continued consideration for onboard use.}

\section{CONCLUSION}~\label{sec:conclusion}
In this work, we presented an autonomous absolute localization framework for a Lunar rover mission driving at night or in dark regions of the Moon.
Our approach consists of using the leading edges of craters as landmarks and matching these detected craters with known Lunar craters from an offline map.
We validated our proposed approach in a simulation environment created using Blender and data collected from a field test conducted at Cinder Lakes.
Future contributions will focus on how the proposed ShadowNav approach would perform within a full system that is running relative localization on-board.
A further area of exploration is the problem of recovering from errors errors greater than \SI{10}{\meter} or to contain an understanding of faults to know if the system was unable to localize.
\revised{
We also plan to study the introduction of additional measurement models such as Lunar orbiter-based position updates and incorporate this into a sensor fusion framework that extends current crater detection and visual odometry measurements.
Inspired by the use of pose graph optimization techniques for tackling the localization problem for the upcoming CADRE Lunar rover mission~\cite{DeLaCroixRossiEtAl2024}, we plan to investigate the use of factor graph-techniques for handling the back end localization in lieu of the particle filter.
}
Finally, the system should be evaluated with absolute localization in the loop over long distances to ensure stability.

\section*{ACKNOWLEDGEMENTS}
The authors would like to thank John Elliott, Jeffery Hall, Satish Khanna, Hari Nayar, Issa Nesnas, Curtis Padgett, and Chris Yahnker for their discussions during the development of this work.


\bibliographystyle{unsrtnat}
\bibliography{project}

\begin{thebibliography}{49}
\providecommand{\natexlab}[1]{#1}
\providecommand{\url}[1]{\texttt{#1}}
\expandafter\ifx\csname urlstyle\endcsname\relax
  \providecommand{\doi}[1]{doi: #1}\else
  \providecommand{\doi}{doi: \begingroup \urlstyle{rm}\Url}\fi

\bibitem[NAS(2022)]{NASEM2022}
Origins, worlds, and life: A decadal strategy for planetary science and astrobiology 2023--2032.
\newblock Technical report, 2022.

\bibitem[Keane et~al.(2022)Keane, Tikoo, and Elliott]{KeaneTikooEtAl2022}
J.~T. Keane, S.~M. Tikoo, and J.~Elliott.
\newblock {Endurance}: {Lunar} {South} {Pole}-{Atken} {Basin} traverse and sample return rover.
\newblock Technical report, 2022.

\bibitem[Verma et~al.(2023)Verma, Maimone, Graser, Rankin, Kaplan, Myint, Huang, Chung, Davis, Tumbar, Tirona, and Lashore]{VermaMaimoneEtAl2023}
V.~Verma, M.~Maimone, E.~Graser, A.~Rankin, K.~Kaplan, S.~Myint, J.~Huang, A.~Chung, K.~Davis, A.~Tumbar, I.~Tirona, and M.~Lashore.
\newblock Results from the first year and a half of {Mars} 2020 robotic operations.
\newblock 2023.

\bibitem[Maimone et~al.(2007)Maimone, Cheng, and Matthies]{MaimoneChengEtAl2007}
M.~Maimone, Y.~Cheng, and L.~Matthies.
\newblock Two years of visual odometry on the {Mars} {Exploration} {Rovers}.
\newblock 24\penalty0 (3):\penalty0 169--186, 2007.

\bibitem[Maimone et~al.(2022)Maimone, Patel, Sabel, Holloway, and Rankin]{MaimonePatelEtAl2022}
M.~Maimone, N.~Patel, A.~Sabel, A.~Holloway, and A.~Rankin.
\newblock Visual odometry thinking while driving for the {Curiosity} {Mars} rover’s three-year test campaign: Impact of evolving constraints on verification and validation.
\newblock 2022.

\bibitem[Matthies et~al.(2022)Matthies, Daftry, Tepsuporn, Cheng, Atha, Swan, Ravichandar, and Ono]{MatthiesDaftryEtAl2022}
L.~Matthies, S.~Daftry, S.~Tepsuporn, Y.~Cheng, D.~Atha, R.~M. Swan, S.~Ravichandar, and M.~Ono.
\newblock Lunar rover localization using craters as landmarks.
\newblock 2022.

\bibitem[Daftry et~al.(2023)Daftry, Chen, Cheng, Tepsuporn, Khattak, Matthies, Coltin, Naam, Ma, and Deans]{DaftryEtAl2023}
S.~Daftry, Z.~Chen, Y.~Cheng, S.~Tepsuporn, S.~Khattak, L.~Matthies, B.~Coltin, U.~Naam, L.~M. Ma, and M.~Deans.
\newblock {LunarNav}: Crater-based localization for long-range autonomous rover navigation.
\newblock 2023.

\bibitem[Hiesinger et~al.(2012)Hiesinger, {van} {der}~Bogert, Pasckert, Funcke, Giacomini, Ostrach, and Robinson]{HiesingerVanDerBogertEtAl2012}
H.~Hiesinger, C.~H. {van} {der}~Bogert, J.~H. Pasckert, L.~Funcke, L.~Giacomini, L.~R. Ostrach, and M.~S. Robinson.
\newblock How old are young lunar craters?
\newblock 117\penalty0 (12):\penalty0 1--15, 2012.

\bibitem[Robinson et~al.(2010)Robinson, Brylow, Tschimmel, Humm, Lawrence, Thomas, Denevi, Bowman-Cisneros, Zerr, Ravine, Caplinger, Ghaemi, Schaffner, Malin, Mahanti, Bartels, Anderson, Tran, Eliason, {McEwen}, Turtle, Jolliff, and Hiesinger]{RobinsonBrylowEtAl2010}
M.~S. Robinson, S.~M. Brylow, M.~Tschimmel, D.~Humm, S.~J. Lawrence, P.~C. Thomas, B.~W. Denevi, E.~Bowman-Cisneros, J.~Zerr, M.~A. Ravine, M.~A. Caplinger, F.~T. Ghaemi, J.~A. Schaffner, M.~C. Malin, P.~Mahanti, A.~Bartels, J.~Anderson, T.~N. Tran, E.~M. Eliason, A.~S. {McEwen}, E.~Turtle, B.~L. Jolliff, and H.~Hiesinger.
\newblock {Lunar} {Reconnaissance} {Orbiter} ({LROC}) camera instrument overview.
\newblock 150, 2010.

\bibitem[Cisneros et~al.(2017)Cisneros, Awumah, Brown, Martin, Paris, Povilaitis, Boyd, Robinson, and {LROC Team}]{CisnerosAwumahEtAl2017}
E.~Cisneros, A.~Awumah, H.~M. Brown, A.~C. Martin, K.~N. Paris, R.~Z. Povilaitis, A.~K. Boyd, M.~S. Robinson, and {LROC Team}.
\newblock {Lunar} {Reconnaissance} {Orbiter} camera permanently shadowed region imaging -- atlas and controlled mosaics.
\newblock 2017.

\bibitem[Robinson and Elliott(2022)]{RobinsonElliott2020}
M.~Robinson and J.~Elliott.
\newblock {Intrepid} planetary mission concept study report.
\newblock Technical report, 2022.

\bibitem[INS(2022)]{INSPIRE2022}
{INSPIRE} ({IN} situ {Solar} system {Polar} {Ice} {Roving} {Explorer}): A mission concept study from the {Decadal} {Survey} for {Planetary} {Science} and {Astrobiology} 2022--2032.
\newblock Technical report, 2022.

\bibitem[Cauligi et~al.(2023)Cauligi, Swan, Ono, Daftry, Elliott, Matthies, and Atha]{CauligiSwanEtAl2023}
A.~Cauligi, R.~M. Swan, H.~Ono, S.~Daftry, J.~Elliott, L.~Matthies, and D.~Atha.
\newblock {ShadowNav}: Crater-based localization for nighttime and {Permanently} {Shadowed} {Region} {Lunar} navigation.
\newblock 2023.

\bibitem[Johnson et~al.(2008)Johnson, Goldberg, Cheng, and Matthies]{JohnsonGoldbergEtAl2008}
A.~E. Johnson, S.~B. Goldberg, Y.~Cheng, and L.~H. Matthies.
\newblock Robust and efficient stereo feature tracking for visual odometry.
\newblock 2008.

\bibitem[Nash et~al.(2024)Nash, Dwight, Saldyt, Wang, Myint, Ansar, and Verma]{NashDwightEtAl2024}
J.~Nash, Q.~Dwight, L.~Saldyt, H.~Wang, S.~Myint, A.~Ansar, and V.~Verma.
\newblock {Censible}: A robust and practical global localization framework for planetary surface missions.
\newblock 2024.

\bibitem[Verma et~al.(2024)Verma, Nash, Saldyt, Dwight, Wang, Myint, Biesiadecki, Maimone, Tumbar, Ansar, Kubiak, and Hogg]{VermaNashEtAl2024}
V.~Verma, J.~Nash, L.~Saldyt, Q.~Dwight, H.~Wang, S.~Myint, J.~Biesiadecki, M.~Maimone, A.~Tumbar, A.~Ansar, G.~Kubiak, and R.~Hogg.
\newblock Enabling long \& precise drives for the {Perseverance} {Mars} rover via onboard global localization.
\newblock 2024.

\bibitem[Enright et~al.(2012)Enright, Barfoot, and Soto]{EnrightBarfootEtAl2012}
J.~Enright, T.~Barfoot, and M.~Soto.
\newblock Star tracking for planetary rovers.
\newblock 2012.

\bibitem[Cozman and Krotkov(1997)]{CozmanKrotkov1997}
F.~Cozman and E.~Krotkov.
\newblock Automatic mountain detection and pose estimation for teleoperation of {Lunar} rovers.
\newblock 1997.

\bibitem[Ebadi et~al.(2022)Ebadi, Coble, Atha, Schwartz, Padgett, and Hook]{EbadiCobleEtAl2022}
K.~Ebadi, K.~Coble, D.~Atha, R.~Schwartz, C.~Padgett, and J.~V. Hook.
\newblock Semantic mapping in unstructured environments: Toward autonomous localization of planetary robotic explorers.
\newblock 2022.

\bibitem[Cozman et~al.(2000)Cozman, Krotkov, and Guestrin]{CozmanKrotkovEtAl2000}
F.~Cozman, E.~Krotkov, and C.~Guestrin.
\newblock Outdoor visual position estimation for planetary rovers.
\newblock 9:\penalty0 135--150, 2000.

\bibitem[Carle and Barfoot(2010)]{CarleBarfoot2010}
P.~J.~F. Carle and T.~D. Barfoot.
\newblock Global rover localization by matching lidar and orbital {3D} maps.
\newblock 2010.

\bibitem[Sun et~al.(2013)Sun, Abshire, {McGarry}, Neumann, Smith, Cavanaugh, Harding, Zwally, Smith, and Zuber]{SunAbshireEtAl2013}
X.~Sun, J.~B. Abshire, J.~F. {McGarry}, G.~A. Neumann, J.~C. Smith, J.~F. Cavanaugh, D.~J. Harding, H.~J. Zwally, D.~E. Smith, and M.~T. Zuber.
\newblock {Space} {Lidar} {Developed} at the {NASA} {Goddard} {Space} {Flight} {Center}-- {The} {First} 20 {Years}.
\newblock \emph{{IEEE Journal on Selected Topics in Applied Earth Observations and Remote Sensing}}, 6\penalty0 (3):\penalty0 1660--1675, 2013.

\bibitem[Lorenz et~al.(2018)Lorenz, Turtle, Barnes, Trainer, Adams, Hibbard, Sheldon, Zacny, Peplowski, Lawrence, Ravine, {McGee}, Sotzen, {MacKenzie}, Langelaan, Schmitz, Wolfarth, and Bedini]{LorenzTurtleEtAl2018}
R.~D. Lorenz, E.~P. Turtle, J.~W. Barnes, M.~G. Trainer, D.~S. Adams, K.~E. Hibbard, C.~Z. Sheldon, K.~Zacny, P.~N. Peplowski, D.~J. Lawrence, M.~A. Ravine, T.~G. {McGee}, K.~S. Sotzen, S.~M. {MacKenzie}, J.~W. Langelaan, S.~Schmitz, L.~S. Wolfarth, and P.~D. Bedini.
\newblock {Dragonfly}: A rotorcraft lander concept for scientific exploration at {Titan}.
\newblock \emph{{Johns Hopkins {APL} Technical Digest}}, 34\penalty0 (3):\penalty0 374--387, 2018.

\bibitem[Hirschmuller(2007)]{Hirschmuller2007}
H.~Hirschmuller.
\newblock Stereo processing by semiglobal matching and mutual information.
\newblock 30\penalty0 (2):\penalty0 328--341, 2007.

\bibitem[Zhang et~al.(2020)Zhang, Qi, Yang, Prisacariu, Wah, and Torr]{ZhangQiEtAl2020}
F.~Zhang, X.~Qi, R.~Yang, V.~Prisacariu, B.~Wah, and P.~Torr.
\newblock Domain-invariant stereo matching networks.
\newblock 2020.

\bibitem[Liounis et~al.(2019)Liounis, Swenson, Small, Lyzhoft, Ashman, Getzandanner, Highsmith, Moreau, Adam, Antreasian, and Lauretta]{LiounisSwensonEtAl2019}
A.~Liounis, J.~Swenson, J.~Small, J.~Lyzhoft, B.~Ashman, K.~Getzandanner, D.~Highsmith, M.~Moreau, C.~Adam, P.~Antreasian, and D.~S. Lauretta.
\newblock Independent optical navigation processing for the {OSIRIS-REx} mission using the {Goddard} {Image} {Analysis} and {Navigation} {Tool}.
\newblock In \emph{{RPI Space Imaging Workshop}}, 2019.

\bibitem[Woicke et~al.(2018)Woicke, Moreno~Gonzalez, {El}-{Hajj}, Mes, Henkel, Autar, and Klavers]{WoickeMorenoEtAl2018}
S.~Woicke, A.~S. Moreno~Gonzalez, I.~{El}-{Hajj}, J.~W.~F. Mes, M.~Henkel, R.~S.~D. Autar, and R.~A. Klavers.
\newblock Comparison of crater-detection algorithms for terrain-relative navigation.
\newblock 2018.

\bibitem[Silburt et~al.(2019)Silburt, {Ali}-{Dib}, Zhu, Jackson, Valencia, Kissin, Tamayo, and Menou]{SilburtAliDibEtAl2019}
A.~Silburt, M.~{Ali}-{Dib}, C.~Zhu, A.~Jackson, D.~Valencia, Y.~Kissin, D.~Tamayo, and K.~Menou.
\newblock Lunar crater identification via deep learning.
\newblock 317:\penalty0 27--38, 2019.

\bibitem[Klear(2018)]{Klear2018}
M.~R. Klear.
\newblock {PyCDA}: An open-source library for automated crater detection.
\newblock In \emph{{Planetary Crater Consortium}}, 2018.

\bibitem[Hwangbo et~al.(2009)Hwangbo, Di, and Li]{HwangboDiEtAl2009}
J.~W. Hwangbo, K.~Di, and R.~Li.
\newblock Integration of orbital and ground image networks for the automation of rover localization.
\newblock In \emph{American Society for Photogrammetry and Remote Sensing Annual Conference}, 2009.

\bibitem[Bhamidipati et~al.(2023)Bhamidipati, Mina, Sanchez, and Gao]{BhamidipatiMinaEtAl2023}
S.~Bhamidipati, T.~Mina, A.~Sanchez, and G.~Gao.
\newblock Satellite constellation design for a lunar navigation and communication system.
\newblock \emph{{NAVIGATION}}, 70\penalty0 (4), 2023.

\bibitem[Cortinovis et~al.(2024)Cortinovis, Mina, and Gao]{CortinovisMinaEtAl2024}
M.~Cortinovis, T.~Mina, and G.~Gao.
\newblock Assessment of single satellite-based lunar positioning for the {NASA} {Endurance} {Mission}.
\newblock 2024.

\bibitem[Audet et~al.(2024)Audet, Melman, Molli, Sesta, Plumaris, Psychas, Swinden, Giordano, and {Ventura}-{Traveset}]{AudetMelmanEtAl2024}
Y.~Audet, F.~T. Melman, S.~Molli, A.~Sesta, M.~Plumaris, D.~Psychas, R.~Swinden, P.~Giordano, and J.~{Ventura}-{Traveset}.
\newblock Positioning of a lunar surface rover on the south pole using {LCNS} and {DEMs}.
\newblock 74\penalty0 (6), 2024.

\bibitem[Wu et~al.(2019)Wu, Potter, Ludivig, Chung, and Seabrook]{WuPotterEtAl2019}
B.~Wu, R.~W.~K. Potter, P.~Ludivig, A.~S. Chung, and T.~Seabrook.
\newblock Absolute localization through orbital maps and surface perspective imagery: A synthetic lunar dataset and neural network approach.
\newblock 2019.

\bibitem[Franchi and Ntagiou(2022)]{FranchiNtagiou2022}
V.~Franchi and E.~Ntagiou.
\newblock Planetary rover localisation via surface and orbital image matching.
\newblock 2022.

\bibitem[Silvestrini et~al.(2022)Silvestrini, Piccinin, Zanotti, Brandonisio, Bloise, Feruglio, Lunghi, Lavagna, and Varile]{SilvestriniPiccininEtAl2022}
S.~Silvestrini, M.~Piccinin, G.~Zanotti, A.~Brandonisio, I.~Bloise, L.~Feruglio, P.~Lunghi, M.~Lavagna, and M.~Varile.
\newblock Optical navigation for lunar landing based on convolutional neural network crater detector.
\newblock \emph{{Aerospace Science and Technology}}, 123:\penalty0 107503, 2022.

\bibitem[Maki et~al.(2020)Maki, Gruel, {McKinney}, Ravine, Morales, Lee, Willson, {Copley}-{Woods}, Valvo, Goodsall, {McGuire}, Sellar, and {others}]{MakiGruelEtAl2020}
J.~N. Maki, D.~Gruel, C.~{McKinney}, M.~A. Ravine, M.~Morales, D.~Lee, R.~Willson, D.~{Copley}-{Woods}, M.~Valvo, T.~Goodsall, J.~{McGuire}, R.~G. Sellar, and {others}.
\newblock The {Mars} 2020 {Engineering} {Cameras} and microphone on the {Perseverance} {Rover}: A next-generation imaging system for {Mars} exploration.
\newblock 216\penalty0 (137):\penalty0 1--48, 2020.

\bibitem[Balaram et~al.(2018)Balaram, Canham, Duncan, Grip, Johnson, Maki, Quon, Stern, and Zhu]{BalaramCanhamEtAl2018}
B.~Balaram, T.~Canham, C.~Duncan, H.~F. Grip, W.~Johnson, J.~Maki, A.~Quon, R.~Stern, and D.~Zhu.
\newblock {Mars} {Helicopter} technology demonstrator.
\newblock 2018.

\bibitem[{de} {la}~{Croix} et~al.(2024){de} {la}~{Croix}, Rossi, Brockers, Aguilar, Albee, Boroson, Cauligi, Delaune, and {others}]{DeLaCroixRossiEtAl2024}
J.-P. {de} {la}~{Croix}, F.~Rossi, R.~Brockers, D.~Aguilar, K.~Albee, E.~Boroson, A.~Cauligi, J.~Delaune, and {others}.
\newblock Multi-agent autonomy for space exploration on the {CADRE} {Lunar} technology demonstration mission.
\newblock 2024.

\bibitem[Yadav et~al.(2014)Yadav, Maheshwari, and Agarwal]{YadavMaheshwariEtAl2014}
G.~Yadav, S.~Maheshwari, and A.~Agarwal.
\newblock Contrast limited adaptive histogram equalization based enhancement for real time video system.
\newblock In \emph{{Proc.\ IEEE Int.\ Conf.\ on Advances in Computing, Communications and Informatics }}, 2014.

\bibitem[Kogan(2024)]{mrcal}
D.~Kogan.
\newblock mrcal.
\newblock \url{http://mrcal.secretsauce.net}, 2024.

\bibitem[Fox et~al.(2001)Fox, Thrun, Burgard, and Dellaert]{FoxThrunEtAl2001}
D.~Fox, S.~Thrun, W.~Burgard, and F.~Dellaert.
\newblock \emph{Sequential {Monte} {Carlo} Methods in Practice}, chapter Particle Filters for Mobile Robot Localization, pages 401--428.
\newblock 2001.

\bibitem[Arulampalam et~al.(2002)Arulampalam, Maskell, Gordon, and Clapp]{ArulampalamMaskellEtAl2002}
M.~S. Arulampalam, S.~Maskell, N.~Gordon, and T.~Clapp.
\newblock A tutorial on particle filters for online nonlinear/non-{Gaussian} {Bayesian} tracking.
\newblock 50\penalty0 (2):\penalty0 174--188, 2002.

\bibitem[Crues et~al.(2023)Crues, Bielski, Paddock, Foreman, Bell, Raymond, Hunt, and Bulikhov]{CruesBielskiEtAl2023}
E.~Z. Crues, P.~Bielski, E.~Paddock, C.~Foreman, B.~Bell, C.~Raymond, T.~Hunt, and D.~Bulikhov.
\newblock Approaches for validation of lighting environments in realtime {Lunar} {South} {Pole} simulations.
\newblock 2023.

\bibitem[Hapke(2002)]{Hapke2002}
B.~Hapke.
\newblock Bidirectional reflectance spectroscopy: 5. the coherent backscatter opposition effect and anisotropic scattering.
\newblock \emph{Icarus}, 157\penalty0 (2):\penalty0 523--534, 2002.

\bibitem[Hapke(2012)]{Hapke2012}
B.~Hapke.
\newblock \emph{Theory of reflectance and emittance spectroscopy}.
\newblock 2012.

\bibitem[Schmidt and Bourguignon(2019)]{SchmidtBourguignon2019}
F.~Schmidt and S.~Bourguignon.
\newblock Efficiency of {BRDF} sampling and bias on the average photometric behavior.
\newblock 317:\penalty0 10--26, 2019.

\bibitem[Xu et~al.(2020)Xu, Liu, Liu, Liu, and Shu]{XuLiuEtAl2020}
X.~Xu, J.~Liu, D.~Liu, B.~Liu, and R.~Shu.
\newblock Photometric correction of {Chang’E}-1 interference imaging spectrometer’s ({IIM}) limited observing geometries data with {Hapke} model.
\newblock \emph{Remote Sensing}, 12\penalty0 (22):\penalty0 3676, 2020.

\bibitem[Scharstein et~al.(2001)Scharstein, Szeliski, and Zabih]{ScharsteinSzeliskiEtAl2001}
D.~Scharstein, R.~Szeliski, and R.~Zabih.
\newblock A taxonomy and evaluation of dense two-frame stereo correspondence algorithms.
\newblock 2001.

\end{thebibliography}


\begin{IEEEbiography}[{\includegraphics[width=1in,height=1.25in,clip,keepaspectratio]{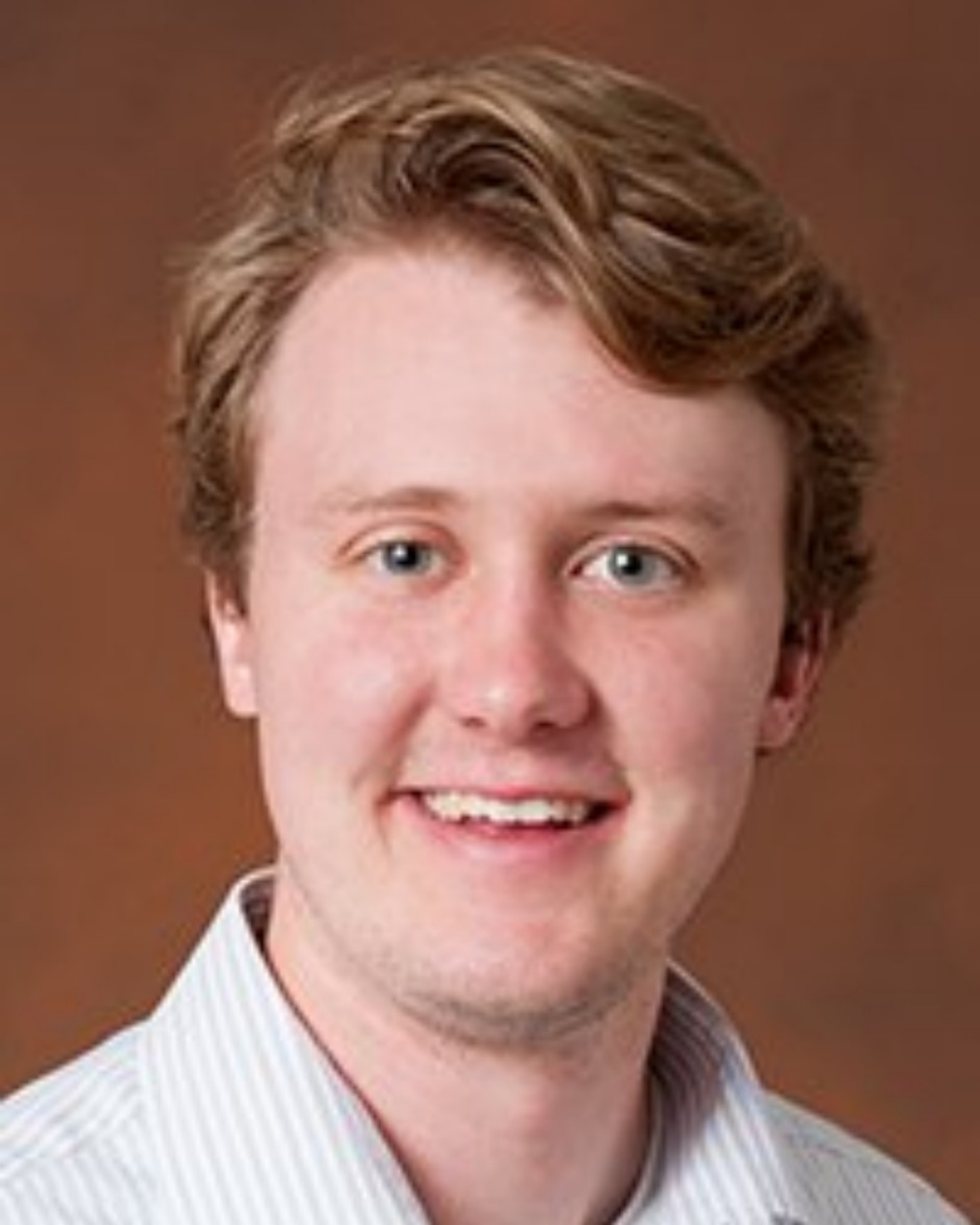}}]
{Deegan Atha}~is a Robotics Technologist within the Perception Systems Group of the Mobility and Robotic Systems Section at NASA's Jet Propulsion Laboratory. He received his B.S. degree from Purdue University in electrical engineering and his M.S. in computer science from the Georgia Institute of Technology. His research focuses on the infusion of robotic perception and learning into autonomous systems operating in unstructured environments. He is the Principal Investigator for the ShadowNav task and the Perception Lead for JPL's team in the DARPA RACER program to develop high-speed, resilient off-road autonomy.
\end{IEEEbiography} 
\begin{IEEEbiography}
[{\includegraphics[width=1in,height=1.25in,clip,keepaspectratio]{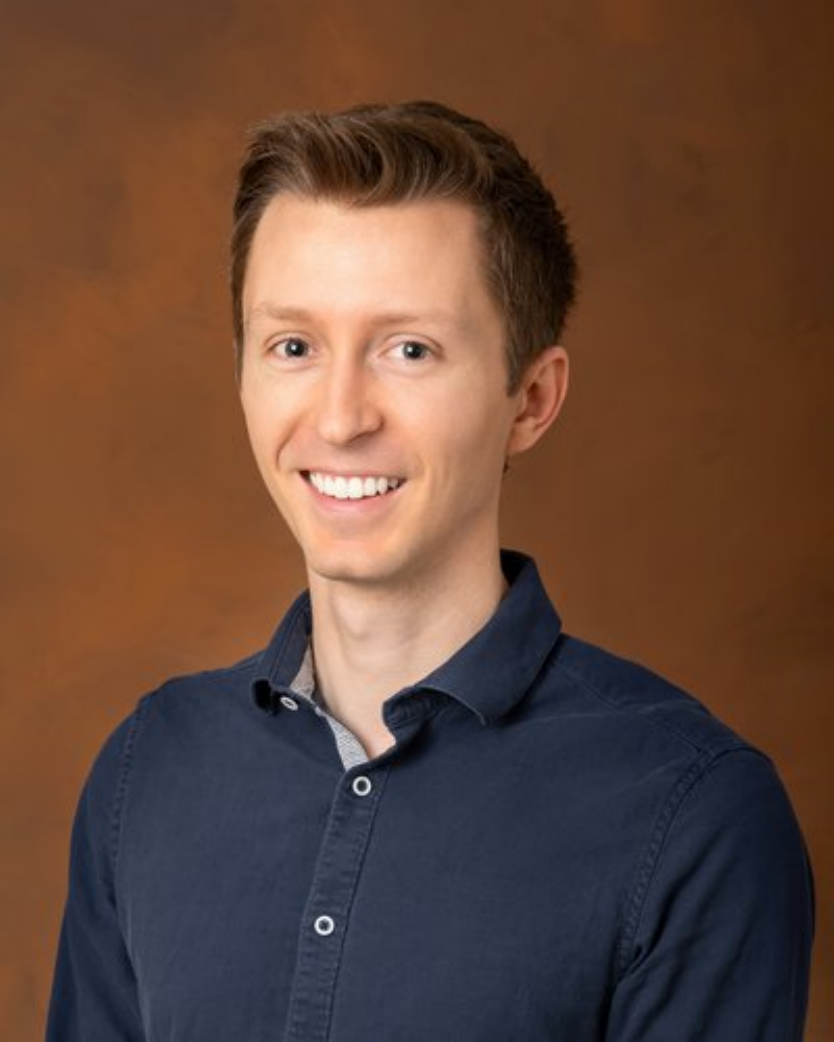}}]
{R. Michael Swan}~is a Robotics Systems Engineer in the Robotic Surface Mobility Group at NASA Jet Propulsion Laboratory. He received his B.S in Computer Engineering from Walla Walla University and his M.S. in Computer Science from the University of Southern California. His work is focused on building reliable robotic software systems. He is the integration lead for multiple projects at JPL, recently architecting the NEO robotics stack for snake robots with the Extant Exobiology Life Surveyor (EELS) project team. He has broad research interests in robotics and supporting technologies such as machine vision, global localization, simulation, extended reality, human robot interaction, and more. 
\end{IEEEbiography} 
\begin{IEEEbiography}[{\includegraphics[width=1in,height=1.25in,clip,keepaspectratio]{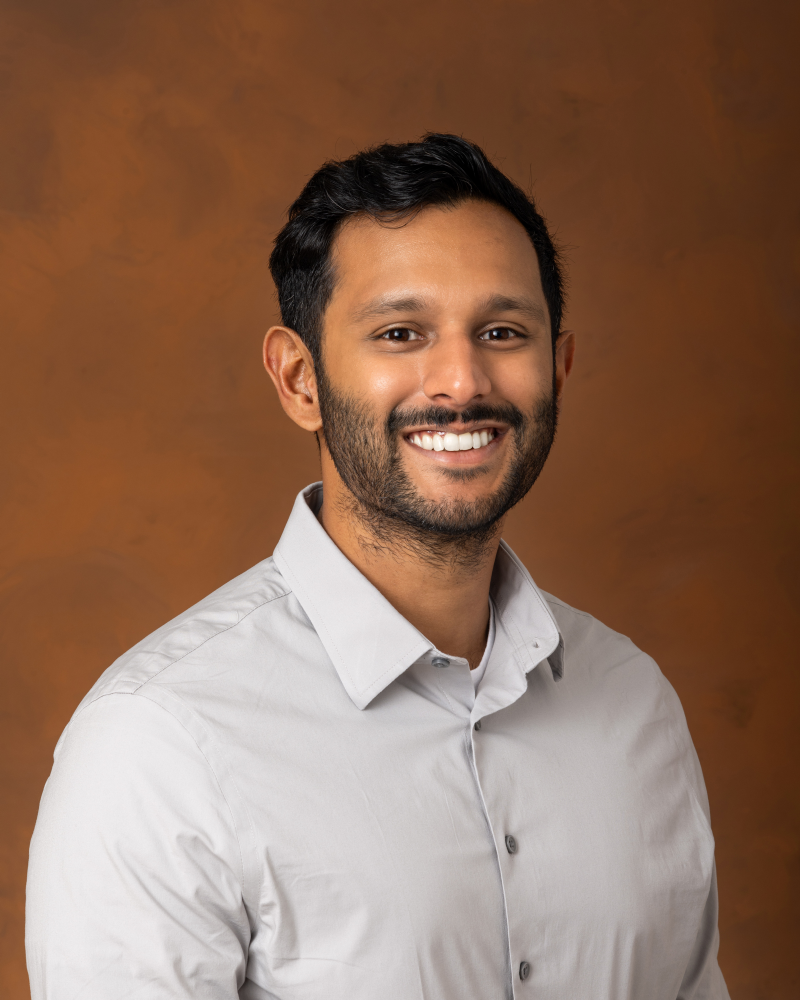}}]
{Abhishek Cauligi}~is a Robotics Technologist in the Robotic Surface Mobility Group at  NASA  Jet  Propulsion  Laboratory, California Institute of Technology.
Abhishek received his PhD. in Aeronautics and Astronautics from Stanford University and has served as the Cognizant Engineer for the motion planner on the CADRE Lunar rover mission.
His research interests lie in leveraging recent advances in nonlinear optimization, machine learning, and control theory towards planning and control for complex spacecraft robotic systems.
\end{IEEEbiography}
\begin{IEEEbiography}[{\includegraphics[width=1in,height=1.25in,clip,keepaspectratio]{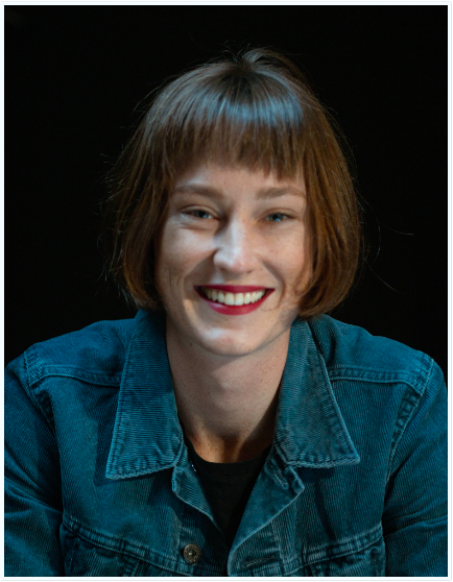}}]
{Anne Bettens}~is an Associate Lecturer in the School of Aerospace, Mechanical, and Mechatronic Engineering at the University of Sydney. She earned her PhD in Aerospace Engineering from the same institution. During her doctoral studies, Anne interned at NASA's Jet Propulsion Laboratory, where she contributed to the ShadowNav project. Her present research focus is on Vision-based Autonomous Navigation of Robotic Craft for Space Exploration.
\end{IEEEbiography}
\begin{IEEEbiography}[{\includegraphics[width=1in,height=1.25in,clip,keepaspectratio]{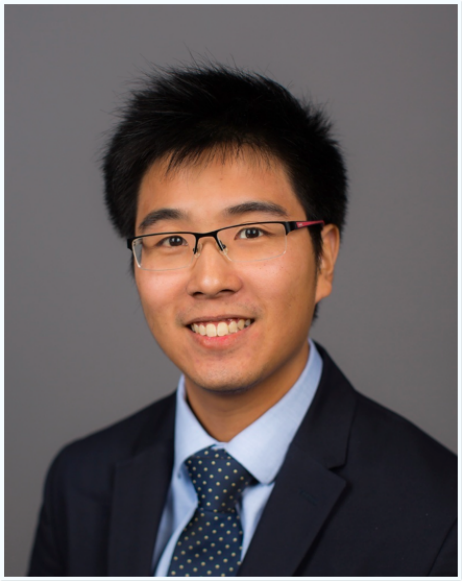}}]
{Edwin Goh}~is a data scientist at the Jet Propulsion Laboratory, California Institute of Technology. He received his B.S., M.S., and Ph.D. degrees in Aerospace Engineering from the Georgia Institute of Technology. His research focuses on the application of machine learning to enable data-driven design, optimization and operation of aerospace systems. His work includes DSN scheduling using reinforcement learning, automated machine learning systems, and self-supervised computer vision for planetary and earth science.
\end{IEEEbiography} 
\begin{IEEEbiography}[{\includegraphics[width=1in,height=1.25in,clip,keepaspectratio]{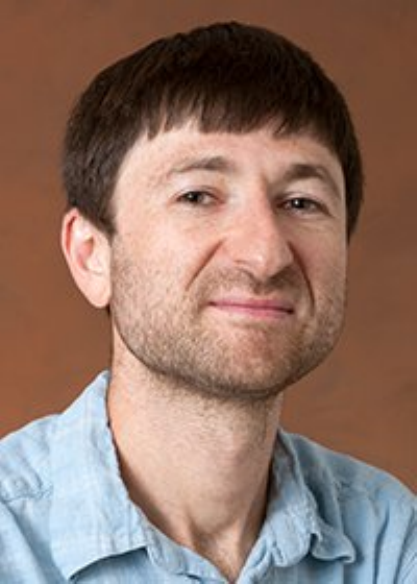}}]
{Dima Kogan}~is a Robotics Technologist at NASA/JPL. Dima received his BS in Engineering and Mathematics at Harvey Mudd College and an MS in Control and Dynamical Systems at Caltech. He is a long time programmer, GNU/Linux user and a Debian Developer. He's interested in estimation and optimization algorithms, particularly as applied to principled sensor calibration.
\end{IEEEbiography}
\begin{IEEEbiography}[{\includegraphics[width=1in,height=1.25in,clip,keepaspectratio]{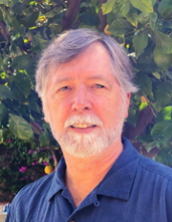}}]
{Larry Matthies}~received B.S., M. Math, and PhD degrees in Computer Science from the University of Regina (1979), University of Waterloo (1981), and Carnegie Mellon University (1989). He has been with JPL for more than 34 years. He has conducted technology development in perception systems for autonomous navigation of robotic vehicles for land, sea, air, and space. He supervised the JPL Computer Vision group for 21 years. He led development of computer vision algorithms for Mars rovers, landers, and helicopters. He is a Fellow of the IEEE and a member of the editorial boards of Autonomous Robots and the Transactions on Field Robotics.
\end{IEEEbiography}
\begin{IEEEbiography}[{\includegraphics[width=1in,height=1.25in,clip,keepaspectratio]{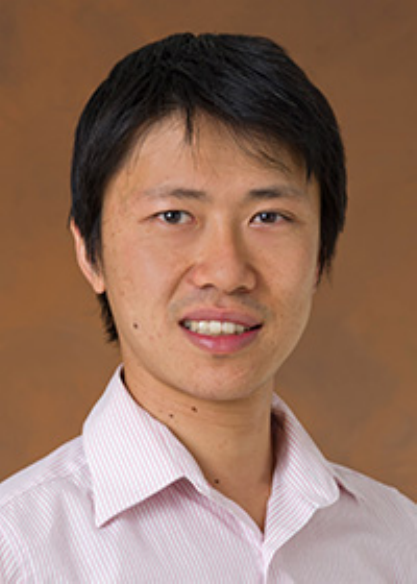}}]
{Masahiro Ono}~is the Group Supervisor of the Robotic Surface Mobility Group. Since he joined JPL in 2013, he has led a number of robotic autonomy research. Recently, he was the PI of the EELS (Exobiology Extant Life Surveyor) project where he led the development of an intelligent snake robot that demonstrated vertical mobility in a glacial shaft. Hiro is a rover operator of the Perseverance rover, and previously he was a flight software developer of the rover’s autonomous driving capability. He also led the development of a machine learning-based Martian terrain classifier, SPOC (Soil Property and Object Classification), which won JPL’s Software of the Year Award in 2020. 
\end{IEEEbiography} 


\end{document}